\documentclass[10pt,journal,compsoc,twoside]{IEEEtran}
\usepackage{amsmath,amssymb,amsthm}
\usepackage{array,arydshln}
\usepackage{graphicx,subfigure}
\usepackage{algorithm}
\usepackage{algorithmic}
\usepackage{color}
\usepackage{multirow}
\usepackage{url}
\usepackage{bm} 
\usepackage{bbm}
\usepackage[hidelinks]{hyperref} 

\input{Definitions}

\newcommand{\norm}[1]{\left\lVert#1\right\rVert}

\newcommand {\bs}[1] {\boldsymbol{#1}}
\newcommand{\bbR}{\mathbb{R}}

\newtheorem{assumption}{Assumption}

\begin{document}

\title{Multivariate Regression with Gross Errors \\ on Manifold-valued Data}

\author{Xiaowei~Zhang,~\IEEEmembership{Member, IEEE},
        Xudong~Shi,
        Yu~Sun,~\IEEEmembership{Member, IEEE},
        Li~Cheng,~\IEEEmembership{Senior~Member, IEEE}
\IEEEcompsocitemizethanks{\IEEEcompsocthanksitem Xiaowei~Zhang is with Bioinformatics Institute, A*STAR, Singapore. ~\protect 
 E-mail: zxwtroy87@gmail.com%
\IEEEcompsocthanksitem Xudong Shi is with School of Computing, National University of Singapore, Singapore. The research is carried out when he is an intern in A*STAR. ~\protect 
\IEEEcompsocthanksitem Yu Sun is with the Singapore Institute for Neurotechnology, National University of Singapore, Singapore 117456. ~\protect 
\IEEEcompsocthanksitem Li~Cheng is with Bioinformatics Institute, A*STAR, Singapore (Corresponding author) ~\protect 
E-mail: chengli@bii.a-star.edu.sg%
}
\thanks{}%
}

%
%



\IEEEcompsoctitleabstractindextext{%
\begin{abstract}
We consider the topic of multivariate regression on manifold-valued output, that is, for a multivariate observation, its output response lies on a manifold.
Moreover, we propose a new regression model to deal with the presence of grossly corrupted manifold-valued responses, a bottleneck issue commonly encountered in practical scenarios. Our model first takes a correction step on the grossly corrupted responses via geodesic curves on the manifold, then performs multivariate linear regression on the corrected data. This results in a nonconvex and nonsmooth optimization problem on Riemannian manifolds.
To this end, we propose a dedicated approach named PALMR, by utilizing and extending the proximal alternating linearized minimization techniques for optimization problems on Euclidean spaces. Theoretically, we investigate its convergence property, where it is shown to converge to a critical point under mild conditions. Empirically, we test our model on both synthetic and real diffusion tensor imaging data, and show that our model outperforms other multivariate regression models when manifold-valued responses contain gross errors, and is effective in identifying gross errors.
\end{abstract}

\begin{IEEEkeywords}
Manifold-valued data, multivariate linear regression, gross error, nonsmooth optimization on manifolds, diffusion tensor imaging.
\end{IEEEkeywords}
}

\maketitle

\IEEEraisesectionheading{\section{Introduction}\label{sec:intro}}
\IEEEPARstart{T}{his} paper focuses on multivariate regression on manifolds~\cite{DavisFBJ:ijcv10,KimEtAl:cvpr14,Cornea:RSSB17,Muralidharan:CVPR12}, where given a multivariate observation $\bs{x}\in\bbR^{d}$, the output response $\bs{y}$ lies on a Riemannian manifold $\mathcal{M}$. This line of work has many applications. For example, research evidence in diffusion tensor imaging (DTI) (e.g.~\cite{Hsu:2008}) indicates that the shape and orientation of diffusion tensors are profoundly affected by age, gender and handedness (i.e. left- or right-handed).
{In particular, we consider noisy manifold-valued output scenarios where data are subject to sporadic contamination by gross errors of large or even unbounded magnitude. Such grossly corrupted data are often encountered in practice due to unreliable data collection or data with missing values: For example, errors in DTI data can be introduced by Echo-Planar Imaging (EPI) distortion~\cite{WuCWLBMP:miccai08} or inter-subject registration~\cite{Zalesky:11}, where practical measurement errors such as Rician noise or other sensor noise have a significant impact on the shape and orientation of tensors \cite{Bastin:98,BasuFW:miccai06}.}
Although the problem of learning from data with possible gross error in Euclidean spaces
has gained increasing interest~\cite{Wright:TIT10,Candes:RPCA11,ChenJSC:TIT13,NguTra:tit13,XuLen:aistat12,Bhatia15}, to our best knowledge, there exists no prior work in dealing with manifold-valued response with gross errors.

\begin{figure}[!t]
\begin{center}
\includegraphics[width=0.9\columnwidth]{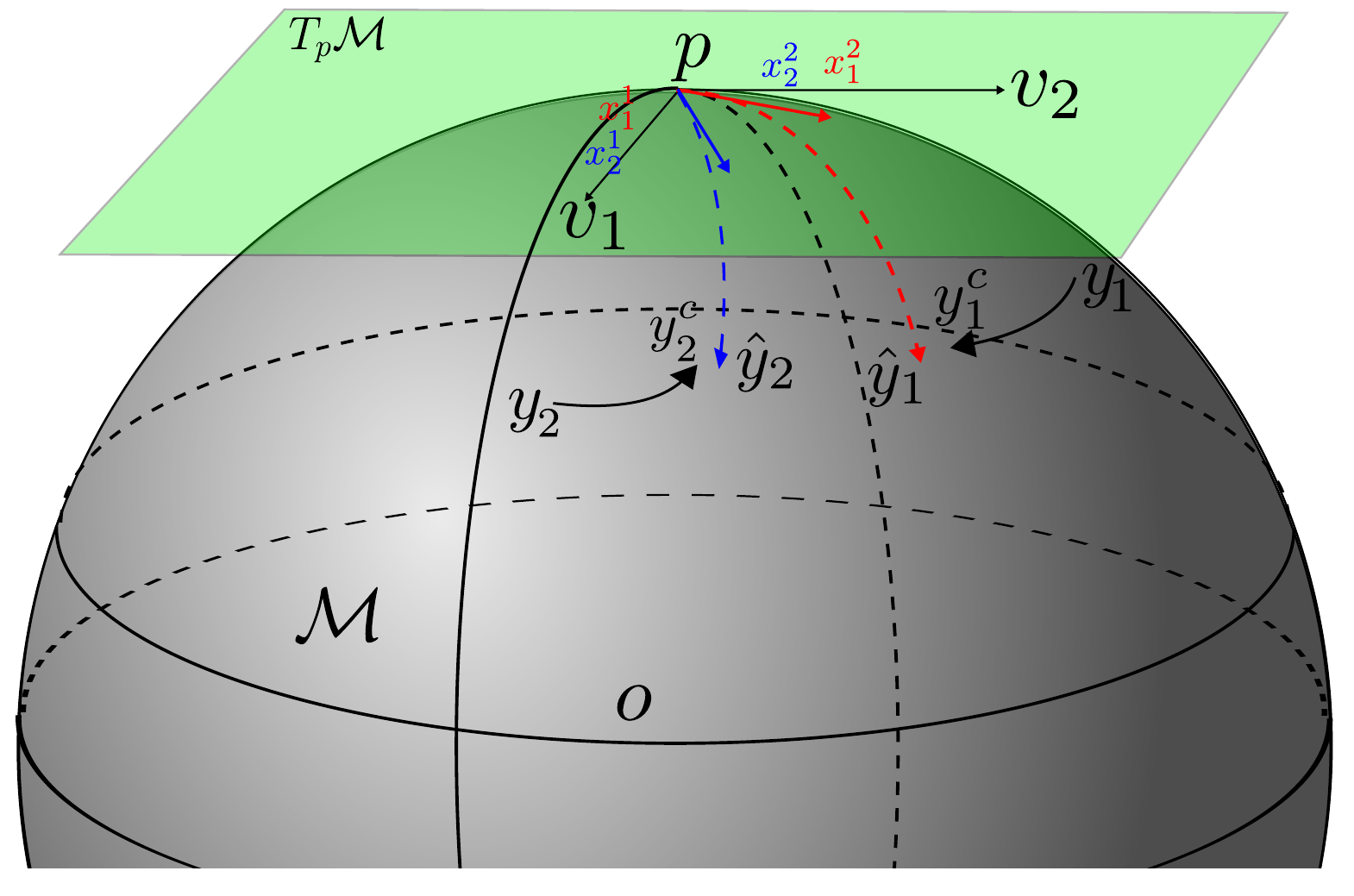}
\end{center}
\vspace{-17pt}
\caption{An illustration of the proposed approach for multivariate regression on grossly corrupted manifold-valued data, which contains two main ingredients:
The first is to obtain the corrected response $\bs{y}_i^c$ by removing its possible gross error, as illustrated by the directed curves on the manifold; The second one involves the manifold-valued regression process using $\{\bs{x}_i, \bs{y}_i^c\}$. Here $x_1^1$ and $x_1^2$ are the components of input $\bs{x}_1$ along tangent vectors $\bs{v}_1$ and $\bs{v}_2$ of point $\bs{p}\in\mathcal{M}$. {Red solid arrow denotes a tangent vector $x_1^1 \bs{v}_1 + x_1^2 \bs{v}_2$ at $\bs{p}$, and red dash arrow is its corresponding geodesic path. $x_2^1$ and $x_2^2$ are also defined similarly using blue color.} See section~\ref{ssec:Euclidean2Manifolds} for details.}\label{fig:sphere}
\vspace{-10pt}
\end{figure}

Our main idea can be summarized as follows: For each manifold-valued response $\bs{y} \in \mathcal{M}$, we explicitly model its possible gross error (in $\bs{y}$). This gives rise to a \emph{corrected} manifold-valued data $\bs{y}^c$ by removing the identified gross error component from $\bs{y}$, which is realized via geodesic curves on $\mathcal{M}$. Note that $\bs{y}^c$ could be the same as $\bs{y}$, corresponding to no gross error in $\bs{y}$. Then the corrected manifold-valued data can be utilized as the responses in multivariate geodesic regression, which boils down to a known problem~\cite{KimEtAl:cvpr14}. More details are illustrated in \figurename~\ref{fig:sphere} and are fully described in Section~\ref{sec:ourAppro}.
Unfortunately, the induced optimization problem becomes rather challenging as it contains nonconvex and nonsmooth functions on manifolds.
Inspired by the recent development of proximal alternating linearized minimization (PALM) methods in Euclidean spaces, in this paper we propose to generalize this technique onto Riemannian manifolds~\cite{BolSabTeb:mp14}, which we have named as \emph{PALMR}. 

The main contributions of this paper are three-fold.
First, we propose to address a novel problem of multivariate regression on manifolds where the manifold-valued responses are subject to possible contamination of gross errors. Second, a new algorithm named PALMR is proposed to tackle the induced nonconvex and nonsmooth optimization on manifolds, for which we also provide the convergence analysis.
The algorithm and analysis is applicable to a class of nonconvex and nonsmooth optimization problem on manifolds. Empirically our algorithm has been evaluated on both synthetic and real DTI data, where results suggest the algorithm is effective in identifying gross errors and recovering corrupted data, and it produces better predictive results than regression models that do not consider gross errors. Third, our approach makes connections to two established research areas, namely learning from grossly corrupted data and multivariate regression on manifolds:
When we restrict ourselves to the special case of Euclidean space, our approach reduces to robust regression considered in e.g.~\cite{NguTra:tit13,XuLen:aistat12};
On the other hand, when there is no gross error, the problem boils down to that of multivariate regression on manifolds as considered in~\cite{KimEtAl:cvpr14}, where the method of~\cite{KimEtAl:cvpr14} can be regarded as a special case of our approach.
Our code is also made publicly available.~\footnote{Our implementation is available at the project website \url{http://web.bii.a-star.edu.sg/~zhangxw/palmr-SPD/}.}

\subsection{Related work}

Manifold-valued data arise from a wide range of application domains including neural imaging~\cite{Basser02}, shape modelling~\cite{Srivastava:PAMI2005, FletcherLPJ04, HinkleFJ:jmiv14}, robotics~\cite{SaxDriNg:icra09}, graphics~\cite{WanPulPop:atg07}, and symmetric positive matrices~\cite{PorikliTM06, Pennec:ijcv06, Carreira:PAMI2014, Cherian:TNNLS17}.
One prominent example is DTI~\cite{Pennec:ijcv06} where data lie in the Riemannian manifold of $3\times 3$ symmetric positive definite (SPD) matrices. In this work, we use $\mathcal{S}(n)$ and $\mathcal{S}_{++}(n)$ to denote the set of $n\times n$ symmetric matrices and $n\times n$ SPD matrices, respectively. Other examples include higher angular resolution diffusion imaging where data can be modelled as the square root of orientation distribution functions lying on the unit sphere~\cite{DuGKQ14,KimEtAl:cvpr14}, as well as group-valued data such as $SO(3)$ and $SE(3)$ in shape analysis~\cite{FletcherLPJ04} and robotics~\cite{SaxDriNg:icra09}.
It is well known that for such scenarios, it is in general much better to conduct statistical analysis directly on the manifold (i.e. curved space) instead of in the ambient Euclidean space (i.e. flat space), which we also verify empirically.

Unsurprisingly, there exists plenty of prior work studying statistics on manifolds~\cite{Fle:ijcv13,FletcherLPJ04,FletcherJ07,Pennec:ijcv06,Pennec:2006}. This is to be distinguished from the well-known topic of \textit{manifold learning}~\cite{Belkin:jmlr06}, where the data are assumed to be sampled from certain manifold embedded in a usually much higher dimensional Euclidean space and one is supposed to extract intrinsic geometric properties of the manifold from observations. Instead here the manifold is usually known in priori, and the task is to engage appropriate statistical models in the analysis of the manifold-valued data.

In the area of regression on manifolds, Fletcher~\cite{Fle:ijcv13} proposes \textit{geodesic regression} that generalizes univariate linear regression on flat spaces to manifolds by regressing a manifold-valued response from a real-valued independent data with a geodesic curve. \cite{DuGKQ14} adapts the idea of geodesic regression for regressing sphere-valued data against real scalar. \cite{HinkleFJ:jmiv14} investigates parametric polynomial regression on Riemannian manifolds, while \cite{DavisFBJ:ijcv10} studies regression on the group of diffeomorphisms for detecting longitudinal anatomical shape changes. Banerjee et al.~\cite{Banerjee_2016_CVPR} propose a nonlinear kernel-based regression method for manifold-valued data. Hong et al.~\cite{HongKSVN:16} propose a shooting spline-based regression technique specifically designed for the Grassmannian manifold. \cite{SteinkeM:NIPS2008, Hein:NIPS2009} investigate a
family of nonparametric regression models for data on manifolds. The closest work might be~\cite{KimEtAl:cvpr14}, which extends the idea of geodesic regression~\cite{Fle:ijcv13} to multivariate regression on manifolds, and applies it to analyze diffusion weighted imaging data. {In~\cite{Cornea:RSSB17}, the authors investigate multivariate regression models on Riemannian symmetric spaces from a statistical perspective and develop several test statistics for evaluating linear hypotheses of the regression coefficients.} In the area of learning with grossly corrupted data, there have been various methods~\cite{Wright:TIT10,NguTra:tit13,XuLen:aistat12,Bhatia15,LiCS2012} proposed for linear regression with gross errors in the Euclidean space, among which robust lasso in~\cite{NguTra:tit13} and robust multi-task regression in~\cite{XuLen:aistat12} can be considered as special cases of our approach when restricted to Euclidean spaces.

{A recent trend in manifold data analysis is kernel methods on manifolds which aim at embedding the manifold to a reproducing kernel Hilbert space (RKHS). In~\cite{Harandi:ECCV2012} and~\cite{HarandiHSLS15}, kernel methods are developed for sparse coding and dictionary learning on SPD and Grassmann manifolds, respectively. In~\cite{Vemulapalli:CVPR2013}, kernels on SPD and Grassmann manifolds are considered for classification. 
As it is important for such kernels on manifolds to satisfy the positive definite constraint,
significant efforts~\cite{JayasumanaHSLH14, HarandiSJHL14, JayasumanaHSLH15, Harandi:CVPR15, FeragenLH15} have been made in this regard.
Meanwhile, as shown in \cite{FeragenLH15}, these kernels tend to either disregard the original Riemannian structure due to linearization requirement, or violates the positive definiteness constraint. In particular, a geodesic Gaussian kernel is positive definite only if the underlying manifold is Euclidean. Moreover, a geodesic Laplacian kernel is positive definite if and only if conditionally negative definite conditions are satisfied, which is in general not true for curved Riemannian manifolds.
These results suggest that the application of kernel methods in curved manifolds has its limitation.
On the other hand, it is also of interest for the community to investigate on approaches other than kernel based methods.
This motivates us to consider in this work a manifold-valued geodesic regression approach by directly considering the intrinsic Riemannian metric.
}

\section{Background}\label{sec:bg}
We first briefly review some concepts in Riemannian manifolds in subsection \ref{subsec:concept}, nonsmooth analysis and Kurdyka--{\L}ojasiewicz property on Riemannian manifolds in subsections \ref{subsec:nonsmooth} and \ref{subsec:kl}, respectively, which are necessary for the derivation of our algorithm and the proof of convergence. We then review some models regarding multivariate linear regression with gross errors in Euclidean space, whose ideas are utilized to design our new model.

\subsection{Riemannian manifolds}\label{subsec:concept}
Let $(\mathcal{M},\varrho)$ denote a smooth manifold $\mathcal{M}$ endowed with a Riemannian metric $\varrho$. {Moreover, $T_{\bs{p}}\mathcal{M}$ denotes the tangent space at point $\bs{p}$ and $T\mathcal{M}:= \cup_{\bs{p}\in\mathcal{M}}T_{\bs{p}}\mathcal{M}$ denotes the tangent bundle.} Notation $(\bs{p}, \bs{v})\in \mathcal{M}\times T\mathcal{M}$ refers to $\bs{p}$ being a point of $\mathcal{M}$ and $\bs{v}$ being a tangent vector at $\bs{p}$. $\left\langle\bs{u}, ~\bs{v}\right\rangle_{\bs{p}}:=\varrho_{\bs{p}}(\bs{u},\bs{v})$ is the inner product between two vectors $\bs{u}$ and $\bs{v}$ in $T_{\bs{p}}\mathcal{M}$, with $\varrho_{\bs{p}}$ being the metric at $\bs{p}$. The induced norm thus becomes $\|\bs{u}\|_{\bs{p}}:=\left\langle\bs{u}, ~\bs{u}\right\rangle_{\bs{p}}^{1/2}$. Let $\gamma:[a,b]\to \mathcal{M}$ be a piecewise smooth curve such that $\gamma(a)=\bs{p}$ and $\gamma(b)=\bs{q}$, with the curve length as $\int_{a}^b \|\gamma'(t)\|_{\gamma(t)}dt$ where $\gamma'(t)$ denotes derivative. The Riemannian distance $d_{\mathcal{M}}(\bs{p},\bs{q})$ between $\bs{p}$ and $\bs{q}$ is defined as the infimum of the length over all piecewise smooth curves joining these two points.
Let $\nabla$ be the Levi-Civita connection~\footnote{{Roughly speaking, a connection acts as a generalization of directional derivative that connects tangent spaces of nearby points and provides a consistent manner of transporting tangent vectors from one point to another along geodesic curves. A manifold may have many connections. Levi-Civita connection, also called Riemannian connection, is a unique connection that is symmetric and compatible with the Riemannian metric.}} associated with $(\mathcal{M},\varrho)$. Curve $\gamma$ is called a \textit{geodesic} if $\nabla_{\gamma'}\gamma' = 0$. A Riemannian manifold is \textit{complete} if its geodesics $\gamma(t)$ are defined for any value of $t\in\mathbb{R}$. {The \textit{parallel transport} along $\gamma$ from $\bs{p} = \gamma(a)$ to $\bs{q} = \gamma(b)$ is a mapping $P_{\gamma(a)\gamma(b)}:T_{\bs{p}}\mathcal{M}\to T_{\bs{q}}\mathcal{M}$ defined by $P_{\gamma(a)\gamma(b)}(\bs{v}) = V(b)$, where $V$ is the unique vector field satisfying $\nabla_{\gamma'}V = 0$ and $V(a) = \bs{v}$.} The exponential map at point $\bs{p}$ is a mapping $\text{Exp}_{\bs{p}}:T_{\bs{p}}\mathcal{M} \to \mathcal{M}$ defined as $\text{Exp}_{\bs{p}}(\bs{v})=\gamma(1)$, where $\gamma:[0,1]\to\mathcal{M}$ is the geodesic such that $\gamma(0)=\bs{p}$ and $\gamma'(0)=\bs{v}$. The inverse of the exponential map, if exists, is denoted by $\text{Exp}_{\bs{p}}^{-1}$. To simplify notations, we also use $\left\langle,~\right\rangle$, $\|\cdot\|$,  $d(\cdot,\cdot)$, and $\text{Exp}(\bs{p},\bs{v})$ to denote inner product, norm, Riemannian distance, and exponential map respectively, when there is no confusion.
We focus on \textit{Hadamard manifold} $\mathcal{M}$, which is a complete and simply connected finite dimensional Riemannian manifold with nonpositive sectional curvature. The class of Hadamard manifolds possesses many nice properties: For example, any two points in $\mathcal{M}$ can be joined by a \textit{unique} geodesic. In this case, the exponential map is a global diffeomorphism and $d(\bs{p},\bs{q})=\|\text{Exp}_{\bs{p}}^{-1} \bs{q}\|_{\bs{p}}$. One example of Hadamard manifold is the manifold of symmetric positive definite matrices. Motivated readers can consult~\cite{DoCarmo:book92} for further details of manifolds and differential geometry.

\subsection{Nonsmooth analysis on Riemannian manifolds}\label{subsec:nonsmooth}
Given an extended real-valued function $\sigma :\mathcal{M} \to\bbR\cup\{+\infty\}$ we define its domain by $\mbox{dom}~\sigma:=\{\bs{p}\in\mathcal{M} : \sigma(\bs{p}) < +\infty\}$ and its epigraph by $\mbox{epi}\sigma:=\{(\bs{p},\beta)\in\mathcal{M}\times \bbR: \sigma(\bs{p})\leq \beta\}$. {We say that $\sigma$ is a lower semicontinuous function if $\mbox{epi}\sigma$ is closed, and is  proper if $\mbox{dom}~\sigma \neq \emptyset$ and $\sigma(\bs{p}) > -\infty$ for all $\bs{p}\in\mbox{dom}~\sigma$. Proper and lower semicontinuous (PLS) functions play important roles in optimization, since it guarantees the well-definedness of the proximal operator. In particular, given $\bs{p}$ and $\lambda>0$, the proximal map defined as
\[
\mathrm{prox}^{\sigma}_{\lambda}(\bs{p}) := \argmin\limits_{\bs{z}}\left\{\sigma(\bs{z}) + \frac{\lambda}{2}\|\bs{z} - \bs{p}\|\right\}
\]
is well-defined when $\sigma$ is PLS and $\text{inf}~\sigma(\bs{z}) > 0$. In Section~\ref{sec:ourAppro}, we will see that the objective function in our approach is a PLS. Moreover, we have the following definition of (sub)differential of PLS functions on manifolds.}

\begin{definition}[\cite{Quiroz13}] Let $\sigma$ be a PLS function, then
\begin{itemize}
\item the Fr$\acute{e}$chet subdifferential of $\sigma$ at any $\bs{p}\in\mbox{dom}~\sigma$, denoted as $\hat{\partial}\sigma(\bs{p})$, is defined as the set of all $\bs{v}\in T_{\bs{p}}\mathcal{M}$ which satisfies
\[
\lim_{\bs{q}\neq \bs{p}}\inf_{\bs{q}\to \bs{p}} \frac{\sigma(\bs{q}) - \sigma(\bs{p}) - \left\langle \bs{v},~\gamma'(0)\right\rangle}{d(\bs{p},\bs{q})} \geq 0,
\]
for geodesic $\gamma$ joining $\gamma(0) = \bs{p}$ and $\gamma(1) = \bs{q}$. When $\bs{p}\notin\mbox{dom}~\sigma$, we set $\hat{\partial}\sigma(\bs{p}) = \emptyset$.
\item the (limiting) subdifferential of $\sigma$ at any $\bs{p}\in\mathcal{M}$, denoted as $\partial \sigma(\bs{p})$, is defined as
\begin{align*}
\partial \sigma(\bs{p}) = \{ & \bs{v}\in T_{\bs{p}}\mathcal{M}: \exists (\bs{p}^k, \sigma(\bs{p}^k))\to (\bs{p}, \sigma(\bs{p})), \\
 & \exists \bs{v}^k \in \hat{\partial}\sigma(\bs{p}^k) \mbox{~s.t.~} P_{\gamma^k(0)\gamma^k(1)}(\bs{v}^k) \to \bs{v}\},
\end{align*}
where $\gamma^k$ is the geodesic joining $\bs{p}^k$ and $\bs{p}$.
\item $\bs{p}\in\mathcal{M}$ is a critical point of $\sigma$ if $0 \in \partial \sigma(\bs{p})$. We denote the set of critical points of $\sigma$ by $\mbox{crit}~\sigma$. That is,
\[
\mbox{crit}~\sigma = \{\bs{x}\in\mathcal{M} :~ 0 \in \partial \sigma(\bs{x})\}.
\]
\end{itemize}

\end{definition}

If $\bs{p}$ is a local minimizer of $\sigma$ then by the Fermat's rule $0 \in \partial \sigma(\bs{p})$. {If $\sigma$ is differentiable, then its subdifferential reduces to a unique gradient, denoted as $\mbox{grad}\sigma$, which is a vector field satisfying $\left\langle \mbox{grad}\sigma(\bs{p}),~\bs{v}\right\rangle_{\bs{p}} = \bs{v}(\sigma)$ for all $\bs{v}\in T_{\bs{p}}\mathcal{M}$ and $\bs{p}\in\mathcal{M}$. Here $\bs{v}(\sigma)$ denotes the directional derivative of $\sigma$ in the direction $\bs{v}$.} In this case $\partial \sigma(\bs{p}) = \{\mbox{grad}\sigma(\bs{p})\}$. Moreover, we have the following definition of Lipschitz gradients for smooth functions on manifolds:
\begin{definition}[\cite{NetoLimaO:98}]
Let $\sigma:\mathcal{M} \to\bbR$ be a continuously differentiable function and $L > 0$. $\sigma$ is said to have $L$-Lipschitz gradient if, for any $\bs{p},~\bs{q}\in\mathcal{M}$ and any geodesic segment $\gamma:[0, r]\to\mathcal{M}$ joining $\bs{p}$ and $\bs{q}$, then
\[\|\partial \sigma(\gamma(t)) - P_{\gamma(0)\gamma(t)}\partial \sigma(\bs{p})\|_{\gamma(t)} \leq L l(t), ~\forall t\in [0,r],\]
where $\gamma(0)=\bs{p}$, $P_{\gamma(0)\gamma(t)}$ is the parallel transport along $\gamma$ from $\bs{p}$ to $\gamma(t)$, and $l(t)$ denotes the length of the segment between $\bs{p}$ and $\gamma(t)$. In addition, if $\mathcal{M}$ is a Hadamard manifold, then the last inequality becomes
\[\|\partial \sigma(\gamma(t)) - P_{\gamma(0)\gamma(t)}\partial \sigma(\bs{p})\|_{\gamma(t)} \leq L d(\bs{p}, \gamma(t)).\]
\end{definition}

{Since $\sigma$ is continuously differentiable, $\partial \sigma(\gamma(t))$ and $\partial \sigma(\bs{p})$ are the unique tangent vectors at $\gamma(t)$ and $\bs{p}$, respectively. Parallel transport thus becomes necessary to move them onto the same tangent space. Note in general, the right hand sides of the two inequalities above are different. This is due to the fact that for non-Hadamard manifolds, geodesic between two points is usually not unique. Since $d(\bs{p}, \gamma(t))$ is defined as the infimum length of geodesic segments between $\bs{p}$ and $\gamma(t)$, it could be smaller than $l(t)$, which is the length of the segment between $\bs{p}$ and $\gamma(t)$ along a given geodesic $\gamma$. For Hadamard manifolds on the other hand, there exists a unique geodesic between any two points, hence $d(\bs{p}, \gamma(t)) = l(t)$ always holds.}

\subsection{Kurdyka--{\L}ojasiewicz (K-L) property on Riemannian manifolds}\label{subsec:kl}
The Kurdyka--{\L}ojasiewicz (K-L) property plays a crucial role in nonsmooth analysis~\cite{AttouchBRS:10,NetoOSS:13}. In this subsection we extend the K-L property from Euclidean spaces to Riemannian manifolds. To do this, we need to introduce some basic notations. If $A$ is a subset of $\mathcal{M}$, then the distance between $\bs{x}\in\mathcal{M}$ and $A$ is defined by
\[
\mbox{dist}(\bs{x},A):=\inf\{d(\bs{x},\bs{y}): \bs{y}\in A\},
\]
where $A$ is nonempty, and $\mbox{dist}(\bs{x},A) = +\infty$ for all $\bs{x}\in\mathcal{M}$ when $A$ is empty. For a fixed $\bs{x}\in\mathcal{M}$, the open ball neighborhood of $\bs{x}$ with radius $\eta$ is defined as $B(\bs{x},\eta):=\{\bs{y} \in \mathcal{M}: d(\bs{x},\bs{y}) < \eta\}$.

{
\begin{definition}
Given real scalars $\alpha$, $\beta$, and PLS function $\sigma$, we define
\[
[\alpha \leq \sigma \leq \beta] := \{\bs{x} \in \mathcal{M}: \alpha \leq \sigma(\bs{x}) \leq \beta\}.
\]
We define similarly $[\alpha < \sigma < \beta]$.
\end{definition}
}

Now, we define the K-L property.
\begin{definition}[\cite{NetoOSS:13}]\label{KL-def}
Let $\sigma:\mathcal{M}\to \bbR\cup\{+\infty\}$ be a PLS function. The function $\sigma$ is said to have K-L property at $\bar{\bs{x}} \in \mbox{dom}~\sigma$ if there exists $\eta\in (0,\infty]$, a neighborhood $U$ of $\bar{\bs{x}}$ and a continuous concave function $\phi:[0,\eta)\to\bbR_{+}$ such that
\begin{itemize}
\item[(i)] $\phi(0)=0$, $\phi$ is continuously differentiable on $(0,\eta)$ and $\phi'(s)>0$ for all $s\in(0,\eta)$;
\item[(ii)] the following K-L inequality holds
\[
\phi'(\sigma(\bs{x}) - \sigma(\bar{\bs{x}})) \mbox{dist}(0,\partial \sigma(\bs{x})) \geq 1,
\]
$\forall \bs{x} \in U\cap [\sigma(\bar{\bs{x}}) < \sigma < \sigma(\bar{\bs{x}}) + \eta]$.
\end{itemize}
We call $\sigma$ a K-L function if it has K-L property at each point of $\mbox{dom}~\sigma$.
\end{definition}

{The K-L property basically asserts that function $\sigma$ can be made sharp by a reparameterization of its values using $\phi$. In particular, when $\sigma$ is differentiable and $\bar{\bs{x}}$ is critical, i.e., $\partial \sigma(\bar{\bs{x}})=0$,  we can define reparameterization $f(\bs{x}):=\phi(\sigma(\bs{x}) - \sigma(\bar{\bs{x}}))$, then the K-L inequality becomes $\|\partial f(\bs{x}) \|\geq 1$, which avoids flatness around $\bar{\bs{x}}$.  This geometrical feature plays a critical role in proving that the sequence generated by our algorithm converges to a critical point. In Proposition 4 of the supplementary, we also establish K-L property in the neighborhood of non-critical points. K-L functions are ubiquitous in a wide range of applications, including for example semi-algebraic, subanalytic, semiconvex, uniformly convex, and log-exp functions~\cite{AttouchBRS:10,NetoOSS:13}.}

\subsection{Multivariate linear regression with gross errors}
Given a matrix representation of $N$ observations $X \in \bbR^{N \times d}$, and the corresponding $m$-dimensional response matrix $Y\in\bbR^{N\times m}$, one of the central problems in linear regression is to accurately estimate the regression matrix $V\in\bbR^{d\times m}$ from
\begin{equation}\label{LR_Mdl}
Y = X V^* + Z,
\end{equation}
with $Z\in\bbR^{N\times m}$ being the stochastic noise. In most of existing work regarding linear regression, $Z$ is assumed to be composed of entries following normal distribution with zero mean. However, when the response $Y$ is subject to possible gross error, the estimated regression matrix deviates significantly from the true value and becomes unreliable.
To deal with this problem, several recent works~\cite{ChenJSC:TIT13,NguTra:tit13,XuLen:aistat12} suggest to consider model
\begin{equation}\label{LR_Gross_Mdl}
Y = X V^* + G^* + Z,
\end{equation}
where $G^* \in \mathbb{R}^{N \times m}$ is used to explicitly characterize the gross error component.
As in practice only a subset of responses are corrupted by gross error, $G^*$ is a sparse matrix whose nonzero entries are unknown and magnitudes can be arbitrarily large. Moreover, this model can as well be applied to deal with the case where some entries of $Y$ are missing. A commonly used paradigm of estimating $(V^*,G^*)$ is by solving convex optimization problem
\begin{equation}\label{LR_Gross_OP}
\min_{V, G}  \frac{1}{2} \| Y - XV - G \|_F^2 + \lambda R_v(V) + \rho R_g(G),
\end{equation}
where $\lambda>0$ and $\rho>0$ are tuning parameters, and $R_v$ and $R_g$ are regularization terms of $V$ and $G$, respectively. Some frequently used regularization norms include $\ell_1$ norm $\|\cdot\|_1$ which is the summation of the absolute value of all entries, and $\ell_{1,2}$ norm which is the summation of $\ell_2$ norm of rows of a matrix. For example, in \cite{XuLen:aistat12} the authors propose to use $R_v(V)=\|V\|_{1,2}$ and $R_g(G)=\|G\|_1$.

\section{Our Approach}\label{sec:ourAppro}
Consider a set of training examples $\{(\bs{x}_i, \bs{y}_i)\}_{i=1}^N$, where $\bs{y}_i$ lies on Riemannian manifold $\mathcal{M}$ and $\bs{x}_i\in\bbR^{d}$ is the associated independent variable. We propose a novel extension of the modeling approach of Eq.~\eqref{LR_Gross_Mdl} for Euclidean spaces to deal with the more general curved spaces, as follows. 

\subsection{From Euclidean spaces to manifolds}
\label{ssec:Euclidean2Manifolds}
The Model of Eq.~\eqref{LR_Gross_Mdl} can be reformulated as $Y - G^* = X V^* + Z$. Denote $Y^{c} := Y-G^*$, which can be interpreted as \emph{corrected} response after removing the gross error. Now the model of Eq.~\eqref{LR_Gross_Mdl} can be reformulated as standard linear regression in Eq.~\eqref{LR_Mdl} with response $Y^c$. With this in mind, we proceed to extend the aforementioned idea to regression on manifolds. For each manifold-valued response $\bs{y}_i$, denote as $\bs{y}_i^c$ its corrected version. Different from the Euclidean space setting where $Y^c$ can be obtained from $Y$ simply by a translation, we need to ensure that $\bs{y}_i^c$ remains on the manifold. This is accomplished by the exponential map $\bs{y}_i^c=\text{Exp}_{\bs{y}_i}(\bs{g}_i)$ with the gross error $\bs{g}_i\in T_{\bs{y}_i}\mathcal{M}$ over each of the training examples, $i \in \{1,\cdots,N\}$. Note that when $\mathcal{M}$ is an Euclidean space, the exponential map reduces to addition, as $\text{Exp}_{\bs{y}_i}(\bs{g}_i)=\bs{y}_i+\bs{g}_i$. In other words, translation in the affine space is a special case of exponential map in the more general curved space.

As illustrated in \figurename~\ref{fig:sphere}, we first obtain the corrected manifold-valued response $\bs{y}_i^c = \text{Exp}_{\bs{y}_i}(\bs{g}_i)$. Then the relationship between $\bs{x}_i$ and $\bs{y}_i^c$ can be modeled as
\begin{equation}
\label{GR_Gross_Mdl}
\text{Exp}_{\bs{y}_i}(\bs{g}_i) = \text{Exp}\Big(\text{Exp}\big(\bs{p},\sum_{j=1}^d x_i^j\bs{v}_j\big), ~\bs{z}_i\Big),
\end{equation}
where $\bs{p}\in\mathcal{M}$ and $\{\bs{v}_j\}_{j=1}^d \in T_p \mathcal{M}$ is a set of tangent vectors at $\bs{p}$, $x_i^j$ is the $j^{th}$ component of $\bs{x}_i$, and $\bs{z}_i$ is a tangent vector at $\text{Exp}\left(\bs{p},\sum_{j=1}^d x_i^j\bs{v}_j\right)$. Our model can be viewed as a generalization of linear regression model of Eq.~\eqref{LR_Mdl} from flat spaces to manifolds, where $\bs{p}$ denotes the intercept that is in analogy to the origin $0$ in the flat space as in Eq.~\eqref{LR_Mdl}, and exponential map corresponds to the addition operator in Eq.~\eqref{LR_Mdl}.

To measure the training loss, we use
\[
E \left( \bs{p},\{\bs{v}_j\},\{\bs{g}_i\} \right) := \frac{1}{2}\sum_i d^2 \Big( \text{Exp}_{\bs{y}_i} (\bs{g}_i), \text{Exp}_{\bs{p}}(\sum_j x_i^{j} \bs{v}_j ) \Big)
\]
to denote the sum-of-squared Riemannian distance between the corrected data $\bs{y}_i^c = \text{Exp}_{\bs{y}_i} (\bs{g}_i)$ and the prediction $\hat{\bs{y}}_i = \text{Exp}_{\bs{p}}\left(\sum_j x_i^{j} \bs{v}_j \right)$, and let $R_v$ and $R_g$ denote two regularization terms controlling the magnitude of $\{\bs{v}_j\}$ and $\{\bs{g}_i\}$, respectively.
The problem considered in our paper can now be formulated as the following optimization problem
{\small
\begin{align}\label{GR_Gross_Mdl_objFunc}
(\tilde{\bs{p}}, \{\tilde{\bs{v}}_j\},\{\tilde{\bs{g}}\}_i)=
& \argmin_{\tiny \begin{array}{c}(\bs{p},\{\bs{v}_j\}) \in \mathcal{M} \times T\mathcal{M}\\ \{\bs{g}_i\} \in T_{\bs{y}_i}\mathcal{M} \end{array}}
E \left(\bs{p},\{\bs{v}_j\},\{\bs{g}_i\} \right)  \nonumber \\
& \qquad\qquad\quad + \lambda R_v \left(\{\bs{v}_j\} \right) + \rho R_g \left( \{\bs{g}_i\} \right),
\end{align}
}
where $\lambda \geq 0$ and $\rho \geq 0$ are regularization parameters.  {Without loss of generality, we consider regularization terms $R_v \left(\{\bs{v}_j\} \right):=\sum_{j=1}^d \norm{ \bs{v}_j }_{\bs{p}}$ and $R_g \left(\{\bs{g}_i\} \right):=\sum_{i=1}^N \norm{ \bs{g}_i }_{\bs{y}_i}$, with $\norm{ \cdot }_{\bs{p}}$ and $\norm{ \cdot }_{\bs{y}_i}$ being the norm of tangent vectors at $\bs{p}$ and $\bs{y}_i$, respectively. There are two reasons for the choice of $R_v $: First, it enables problem of Eq.~\eqref{GR_Gross_Mdl_objFunc} to contain the multivariate linear regression problems with feature selection in Euclidean spaces as special cases, as shown in Example 1 and Example 2 below; Second, in many applications one may collect a large set of possible variables $\{\bs{x}^j\}$ for each response, and want to find a compact subset of base tangent vectors from $\{\bs{v}_j\}$ and the corresponding $\{\bs{x}^j\}$ that are significant to the manifold-valued output $\bs{y}$. The choice of $R_g$ is based on the assumption that gross errors are usually sporadically spread among data. Now, the optimization problem becomes
{\footnotesize
\begin{align}\label{GR_Gross_OP}
\min\limits_{\tiny \begin{array}{c}(\bs{p},\{\bs{v}_j\})\in \mathcal{M} \times T\mathcal{M} \\ \bs{g}_i\in T_{\bs{y}_i}\mathcal{M} \end{array}}  E \left(\bs{p},\{\bs{v}_j\},\{\bs{g}_i\} \right) + \lambda \sum_{j=1}^d \norm{ \bs{v}_j }_{\bs{p}} + \rho \sum_{i=1}^N \norm{ \bs{g}_i }_{\bs{y}_i}.
\end{align}
}
}


\subsection{Connections to existing works}
We would like to point out that model in Eq.~\eqref{GR_Gross_OP} includes as special cases a number of related research works on gross error or on manifold-valued regression. In this subsection, we provide three such examples.

\begin{example}\label{ex1}
When $\mathcal{M} = \bbR^{m}$, we can establish a connection between the model of Eq.~\eqref{GR_Gross_OP} and the robust multi-task regression studied in \cite{XuLen:aistat12}. Specifically, instead of optimizing Eq.~\eqref{GR_Gross_OP} over $\bs{p}\in\bbR^{m}$ we select $\bs{p}=0$, resulting in
{\small
\begin{align*}
\min\limits_{\bs{v}_j,\bs{g}_i\in\bbR^m} ~
\frac{1}{2}\sum_{i=1}^{N} \| \bs{y}_i-\sum_{j=1}^{d} \bs{x}_i^{j} \bs{v}_j+\bs{g}_i \|^2 + \lambda \sum_{j=1}^{d} \norm{ \bs{v}_j }  + \rho \sum_{i=1}^{N} \norm{ \bs{g}_i },
\end{align*}
}
which can be rewritten as
\begin{align}\label{GR_Gross_RMTL}
\min\limits_{V,G} ~ \frac{1}{2} \norm{ Y-XV-G }_F^2 + \lambda \norm{ V }_{1,2} +\rho \norm{ G }_{1,2},
\end{align}
where $\|\cdot\|$ becomes the usual Euclidean norm, $Y=[\bs{y}_1, \cdots, \bs{y}_N]^\top\in\bbR^{N\times m}$, $X=[\bs{x}_1, \cdots, \bs{x}_N]^\top\in\bbR^{N\times d}$, $V=[\bs{v}_1, \cdots, \bs{v}_d]^\top\in\bbR^{d\times m}$ and $G=[\bs{g}_1, \cdots, \bs{g}_N]^\top\in\bbR^{N\times m}$. The resulting model of Eq.~\eqref{GR_Gross_RMTL} is exactly the one considered in \cite{XuLen:aistat12} except that regularization term $\norm{ G }_1$ in \cite{XuLen:aistat12} is replaced by $\norm{ G }_{1,2}$ here.
\end{example}

\begin{example}\label{ex2}
If $\mathcal{M} = \bbR^{m}$, we can show by Fermat's rule that the optimal solution $\tilde{\bs{p}}$ is given by $\tilde{\bs{p}}=\frac{1}{N}\sum\limits_{i=1}^N(\bs{y}_i+\bs{g}_i-\sum_j \bs{x}_i^{j} \bs{v}_j)$. By substituting $\tilde{\bs{p}}$ into problem of Eq.~\eqref{GR_Gross_OP} and assuming $\{(\bs{x}_i, \bs{y}_i)\}$ has empirical mean $0$, that is, $\sum_{i=1}^{N} \bs{x}_i=0$ and $\sum_{i=1}^{N} \bs{y}_i=0$, the optimization problem of Eq.~\eqref{GR_Gross_OP} reduces to
\begin{align*}
\min\limits_{\bs{v}_j,\bs{g}_i\in\bbR^m} & ~
\frac{1}{2}\sum_{i=1}^{N} \| \bs{y}_i-\sum_{j=1}^{d} \bs{x}_i^{j} \bs{v}_j+\bs{g}_i- \frac{1}{N}\sum_{i=1}^{N}\bs{g}_i \|^2  \\
 & ~ + \lambda \sum_{j=1}^{d} \norm{ \bs{v}_j }  + \rho \sum_{i=1}^{N} \norm{ \bs{g}_i },
\end{align*}
which can be reformulated as
{\small
\begin{align}\label{GR_Gross_Exap}
\min\limits_{V,G} & ~ \frac{1}{2} \|Y-XV-\bar{G}\|_F^2 + \lambda \|V\|_{1,2} +\rho \|G\|_{1,2} \nonumber \\
 s.t. & ~ \bar{G}=\left(I-\frac{1}{N}\mathbbm{1}_N\mathbbm{1}_N^\top\right)G,
\end{align}
}
where $\mathbbm{1}_N\in\mathbb{R}^N$ is a column vector with all entries being 1.
\end{example}

The difference between Example~\ref{ex1} and Example~\ref{ex2} lies in that the former is obtained from selecting $\bs{p} = 0$ while the latter is from optimizing $\bs{p}$ which exactly follows model of Eq.~\eqref{GR_Gross_OP}. The resulting models are quite similar except that model of Eq.~\eqref{GR_Gross_Exap} needs to center variable $G$.

\begin{example}
If we let $\lambda = 0$ and $\rho = +\infty$, then optimization problem of Eq.~\eqref{GR_Gross_OP} reduces to
\begin{align*}
\min\limits_{(\bs{p},\{\bs{v}_j\})\in \mathcal{M}\times T\mathcal{M}} ~ \frac{1}{2}\sum_{i=1}^{N} d^2 \bigg( \bs{y}_i, \text{Exp}_{\bs{p}}\Big(\sum_{j=1}^{d} x_i^{j} \bs{v}_j \Big) \bigg),
\end{align*}
which recovers exactly the model considered in~\cite{KimEtAl:cvpr14}. In this regard, the MGLM model in \cite{KimEtAl:cvpr14} is a special case of our model.
\end{example}

\subsection{PALM for optimization on Hadamard manifolds}
{In this subsection, we propose a new algorithm to solve optimization problem of Eq.~\eqref{GR_Gross_OP}, which is actually a nonsmooth optimization problem on Hadamard manifolds. As explained in details in subsection \ref{subsec:gross},  problem of Eq.~\eqref{GR_Gross_OP} admits the form
\begin{align}\label{Opt_generic}
\min_{\bs{x}\in\mathcal{M}_1, \bs{y}\in\mathcal{M}_2} \Psi(\bs{x},\bs{y}) := f(\bs{x}) + g(\bs{y}) + h(\bs{x},\bs{y}),
\end{align}
where $\mathcal{M}_1$ and $\mathcal{M}_2$ are Hadamard manifolds, $f:\mathcal{M}_1\rightarrow\bbR\cup\{+\infty\}$ and $g:\mathcal{M}_2\rightarrow\bbR\cup\{+\infty\}$ are PLS functions, and $h:\mathcal{M}_1\times\mathcal{M}_2\rightarrow\bbR$ is a smooth function.}

{Many existing optimization techniques are developed to work with Euclidean spaces, thus not directly applicable to curved manifolds.
Meanwhile, an increasing amount of attention has been drawn to the field of optimization on manifolds~\cite{Absil:book07}. For smooth optimization, classical optimization techniques, such as gradient, conjugate gradients, and trust-region methods, have been generalized to the manifold setting~\cite{Absil:book07,BoumalMAS:13,EdelmanAS:1999,HuaGalAbs2015}, which are however not suitable for the nonconvex and nonsmooth optimization manifold-based problem of Eq.~\eqref{GR_Gross_OP}. For nonsmooth optimization, there exist many prior works~\cite{Ferreira:1998, BBSW:SISC16,KovnatskyGB16,Hosseini:SOPT17}. Unfortunately they either cannot exploit the composition structure in Eq.~\eqref{Opt_generic} (e.g., \cite{Ferreira:1998, BBSW:SISC16,Hosseini:SOPT17}), or fail to guarantee convergence (e.g., \cite{KovnatskyGB16}).}

{Recently, a proximal alternating linearized minimization (PALM) algorithm has been proposed in~\cite{BolSabTeb:mp14} for optimization problem of Eq.~\eqref{Opt_generic} with $\mathcal{M}_1 = \mathbb{R}^{n}$ and $\mathcal{M}_2 = \mathbb{R}^{m}$. Inspired by the success of PALM in the Euclidean setting, in what follows we propose PALMR, an inexact proximal alternating minimization algorithm for problem of Eq.~\eqref{Opt_generic}.}

We alternately solve the following two proximally linearized subproblems
\begin{align}
\bs{x}^{k+1}\in\argmin_{\bs{x}\in\mathcal{M}_1} & ~ f(\bs{x}) + \left\langle \mbox{Exp}_{\bs{x}^k}^{-1}\bs{x}, ~\partial_{\bs{x}}h(\bs{x}^k,\bs{y}^k)\right\rangle \nonumber \\
& ~ + \frac{c_k}{2}d_{\mathcal{M}_1}^2(\bs{x}^k,\bs{x}), \label{Subprob_x} \\
\bs{y}^{k+1}\in\argmin_{\bs{y}\in\mathcal{M}_2} & ~ g(\bs{y}) + \left\langle \mbox{Exp}_{\bs{y}^k}^{-1}\bs{y}, ~\partial_{\bs{y}}h(\bs{x}^{k+1},\bs{y}^k)\right\rangle \nonumber \\
& ~ + \frac{d_k}{2}d_{\mathcal{M}_2}^2(\bs{y}^k,\bs{y}), \label{Subprob_y}
\end{align}
where $c_k = \mu_1 L_1(\bs{y}^k)$ and $d_k = \mu_2 L_2(\bs{x}^{k+1})$ with $\mu_1>1$, $\mu_2 > 1$ and $L_1(\bs{y}^k)$, $L_2(\bs{x}^{k+1})$ being the Lipschitz constants of $\partial_{\bs{x}}h$ and $\partial_{\bs{y}}h$, respectively, as to be explained in Assumption \ref{Assump1}. {In particular, by exploiting the fact that $\mathcal{M}_1$ is a Hadamard manifold on which any two points can be joined by a unique geodesic, we have a one-to-one mapping between $\bs{v}\in T_{\bs{x}^k}\mathcal{M}_{1}$ and $\bs{x}\in\mathcal{M}_1$ such that $\bs{x}=\mbox{Exp}_{\bs{x}^k} (\bs{v})$, $\bs{v} = \mbox{Exp}_{\bs{x}^k}^{-1}\bs{x}$ and $d_{\mathcal{M}_1}(\bs{x}^k,\bs{x}) = \|\bs{v}\|$. Thus, a simple substitution reformulates Eq.~\eqref{Subprob_x} as
\[
\bs{v}^{k}\in\argmin_{\bs{v}\in T_{\bs{x}^k}\mathcal{M}_1} \left(f \circ \mbox{Exp}_{\bs{x}^k}\right) (\bs{v}) + \left\langle \bs{v},~\partial_{\bs{x}}h(\bs{x}^k,\bs{y}^k)\right\rangle + \frac{c_k}{2}\|\bs{v}\|^2,
\]
or equivalently,
\[
\bs{v}^{k}\in\argmin_{\bs{v}\in T_{\bs{x}^k}\mathcal{M}_1} \left(f \circ \mbox{Exp}_{\bs{x}^k}\right) (\bs{v}) + \frac{c_k}{2}\|\bs{v} + \frac{1}{c_k}\partial_{\bs{x}}h(\bs{x}^k,\bs{y}^k)\|^2,
\]
which becomes an optimization problem in linear space $T_{\bs{x}^k}\mathcal{M}_1$, and as a result, we have $\bs{x}^{k+1}=\mbox{Exp}_{\bs{x}^k}(\bs{v}^{k})$.} Since $f$ is PLS satisfying $\inf_{\bs{x}\in\mathcal{M}} f(\bs{x}) > -\infty$ and $\mbox{Exp}_{\bs{x}^k}$ is smooth, it follows that the composite function $f\circ\mbox{Exp}_{\bs{x}^k}$ is PLS and $\inf_{\bs{v}\in T_{\bs{x}^k}\mathcal{M}} f\circ\mbox{Exp}_{\bs{x}^k}(\bs{v}) > -\infty$ which, together with Theorem 1.25 of~\cite{RockWets:98}, implies that $\bs{v}^k$ is well-defined. Moreover, the above optimization problem for $\bs{v}^{k}$ is called proximity operator \cite{Moreau:1965}, denoted as
\[
\bs{v}^{k} = \mbox{prox}_{c_k}^{f \circ \mbox{Exp}_{\bs{x}^k}}\left( -\frac{1}{c_k}\partial_{\bs{x}}h(\bs{x}^k,\bs{y}^k) \right).
\]
Similar claims apply to problem of Eq.~\eqref{Subprob_y}, implying the well-definiteness of $\bs{x}^{k+1}$ and $\bs{y}^{k+1}$. Solving Eqs.~\eqref{Subprob_x} and \eqref{Subprob_y} alternately yields the algorithm PALMR outlined in Algorithm~\ref{Opt_Manifold_Alg}.
\begin{algorithm}[!t]
\caption{(\textbf{PALMR}): PALM on Riemannian manifolds} \label{Opt_Manifold_Alg}
\begin{algorithmic}[1]
\REQUIRE $\mu_1 > 1$ and $\mu_2 > 1$.
\ENSURE the sequence $\{(\bs{x}^k, \bs{y}^k)\}_{k\in\mathbb{N}}$.
\STATE Initialization: $(\bs{x}^0, \bs{y}^0)$ and $k=0$.
\WHILE{stopping criterion not satisfied}
  \STATE Set $c_k = \mu_1 L_1(\bs{y}^k)$ and compute $\bs{x}^{k+1}$ as in Eq.~\eqref{Subprob_x}.
  \STATE Set $d_k = \mu_2 L_2(\bs{x}^{k+1})$ and compute $\bs{y}^{k+1}$ as in Eq.~\eqref{Subprob_y}.
\ENDWHILE
\end{algorithmic}
\end{algorithm}

To analyze the convergence of PALMR, we need the following assumptions.
\begin{assumption}\label{Assump1}
$\Psi(\bs{x},\bs{y})$ satisfies the following conditions:
\begin{itemize}
\item[(i)] $\inf f > -\infty$, $\inf g > -\infty$ and $\inf \Psi > -\infty$.
\item[(ii)] For any fixed $\bs{y}$, the function $\bs{x}\to h(\bs{x},\bs{y})$ has $L_1(\bs{y})$-Lipschitz gradient. Likewise, for any fixed $\bs{x}$, the function $\bs{y}\to h(\bs{x},\bs{y})$ has $L_2(\bs{x})$-Lipschitz gradient. Moreover, there exist real scalars $\lambda_i^{-}, \lambda_i^{+} >0$ for $i=1,2$, such that
\begin{align*}
\inf_{k\in\mathbb{N}}\{L_1(\bs{y}^k)\} \geq \lambda_1^{-}, \quad \inf_{k\in\mathbb{N}}\{L_2(\bs{x}^k)\} \geq \lambda_2^{-}, \\
\sup_{k\in\mathbb{N}}\{L_1(\bs{y}^k)\} \leq \lambda_1^{+}, \quad \sup_{k\in\mathbb{N}}\{L_2(\bs{x}^k)\} \geq \lambda_2^{+}
\end{align*}
\item[(iii)] $\partial h$ is Lipschitz continuous on bounded subset of $\mathcal{M}_1 \times \mathcal{M}_2$. More specifically, for bounded subset $A_1\times A_2 \in \mathcal{M}_1 \times \mathcal{M}_2$, there exists constant $L>0$ such that for all $(\bs{x}_i,\bs{y}_i) \in A_1\times A_2$, $i=1,2$, we have
\begin{align*}
\|\partial_{\bs{x}} h(\bs{x}_1,\bs{y}_1) - \partial_{\bs{x}} h(\bs{x}_1,\bs{y}_2)\| \leq L d_{\mathcal{M}_2}(\bs{y}_1,\bs{y}_2), \\
\|\partial_{\bs{y}} h(\bs{x}_1,\bs{y}_1) - \partial_{\bs{y}} h(\bs{x}_2,\bs{y}_1)\| \leq L d_{\mathcal{M}_1}(\bs{x}_1,\bs{x}_2).
\end{align*}
\item[(iv)] $\Psi(\bs{x},\bs{y})$ has the Kurdyka--{\L}ojasiewicz (K-L) property on Hadamard manifolds.
\end{itemize}
\end{assumption}

{Assumption (i) establishes that proximal operators in Eqs.~\eqref{Subprob_x} and \eqref{Subprob_y} are well-defined, leading to the well-definedness of algorithm PALMR. Assumption (ii) provides that $h$ is locally block-Lipschitz continuous, and the boundedness of Lipschitz constants are to ensure sufficient decrease of objective function value over iterations. Assumption (iii) considers the partial gradients of $h$ being Lipschitz continuous, which would be used to derive lower bound for the iteration gap $d(\bs{x}^{k+1}, \bs{x}^k)$ + $d(\bs{y}^{k+1}, \bs{y}^k)$.  Assumption (iv) guarantees that $\{(\bs{x}^k, \bs{y}^k)\}$ form a Cauchy sequence.}

Under Assumption~\ref{Assump1} we have the following theorem, whose proof is provided in the supplementary.
\setcounter{theorem}{0}
\begin{theorem}\label{PALMR_Conv}
Suppose Assumption~\ref{Assump1} holds. Let $\{(\bs{x}^k, \bs{y}^k)\}_{k\in\mathbb{N}}$ be a sequence generated by PALMR. Then either the sequence $\{d_{\mathcal{M}_1\times \mathcal{M}_2}((\bs{x}^0, \bs{y}^0),(\bs{x}^k, \bs{y}^k))\}$ is unbounded or the following assertions hold:
\begin{itemize}
\item[1)] The sequence $\{(\bs{x}^k, \bs{y}^k)\}_{k\in\mathbb{N}}$ has finite length, i.e.
$\sum\limits_{k} d_{\mathcal{M}_1}(\bs{x}^{k+1},\bs{x}^{k}) < \infty$ and $\sum\limits_{k} d_{\mathcal{M}_2}(\bs{y}^{k+1},\bs{y}^{k}) < \infty$.
\item[2)] The sequence $\{(\bs{x}^k, \bs{y}^k)\}_{k\in\mathbb{N}}$ converges to a critical point $(\bs{x}^*, \bs{y}^*)$ of $\Psi$.
\end{itemize}
\end{theorem}
{Based on Theorem \ref{PALMR_Conv}, we know that the sequence $\{(\bs{x}^k, \bs{y}^k)\}$ generated by PALMR converges to a critical point of $\Psi$, provided the boundedness of the sequence. As shown in~\cite{BolSabTeb:mp14}, there are many scenarios where such assumption holds. For example, when functions $f$ and $g$ are convex and $h(\bs{x}, \bs{y}) = \|A\bs{x} - B\bs{y}\|$ where $A$ and $B$ are matrices, then the sequence $\{(\bs{x}^k, \bs{y}^k)\}$ is bounded.}

In what follows, we specifically investigate the dedicated realization of PALMR to solve the optimization problem of Eq.~\eqref{GR_Gross_OP}.
To simplify the notation, the resulting algorithm is also referred to as PALMR when there is no confusion.

\subsection{Applying PALMR to optimization problem of Eq.~\eqref{GR_Gross_OP}}\label{subsec:gross}

Optimization problem of Eq.~\eqref{GR_Gross_OP} in our context can be reformulated as
{\small
\begin{equation*}
\min\limits_{\begin{array}{c}(\bs{p},\{\bs{v}_j\})\in \mathcal{M}_1\\ \{\bs{g}_i\}\in \mathcal{M}_2 \end{array} } \underbrace{E(\bs{p},\{\bs{v}_j\},\{\bs{g}_i\})}_{h(\bs{p},\{\bs{v}_j\},\{\bs{g}_i\})} + \underbrace{\lambda \sum_{j=1}^d \norm{ \bs{v}_j }_{\bs{p}}}_{f(\bs{p},\{\bs{v}_j\})} + \underbrace{\rho \sum_{i=1}^N \norm{ \bs{g}_i }_{\bs{y}_i} }_{g(\{\bs{g}_i\})},
\end{equation*}
}
which is of the form in Eq.~\eqref{Opt_generic} with $\mathcal{M}_1 = \mathcal{M} \times  T\mathcal{M}$ and $\mathcal{M}_2 = T_{\bs{y}_1}\mathcal{M}\times\cdots\times T_{\bs{y}_N}\mathcal{M}$. To apply PALMR to solve problem of Eq.~\eqref{GR_Gross_OP}, we need to evaluate the gradients of $E(\bs{p},\{\bs{v}_j\},\{\bs{g}_i\})$. To simplify the notation, we further denote the prediction $\hat{\bs{y}}_i := \mathrm{Exp}_{\bs{p}} \big(\sum_j x_i^{j} \bs{v}_j \big)$, as well as the derivatives of the exponential map with respect to $\bs{p}$ and $\bs{v}$ as $d_{\bs{p}}\mathrm{Exp}_{\bs{p}}(\bs{v})$ and $d_{\bs{v}}\mathrm{Exp}_{\bs{p}}(\bs{v})$, respectively. Now, the partial gradient of $E$ with respect to $\bs{p}$ amounts to
\begin{equation}\label{grad_Ep}
\partial_{\bs{p}} E = -\sum_i \Big( d_{\bs{p}}\mathrm{Exp}_{\bs{p}} \big( \sum_j x_i^{j} \bs{v}_j \big) \Big)^{\dag} \mathrm{Exp}_{\hat{\bs{y}}_i}^{-1} \bs{y}_i^c \in T_{\bs{p}} \mathcal{M},
\end{equation}
where $(\cdot)^{\dag}$ is the adjoint derivative of the exponential map~\cite{Fle:ijcv13} defined by
$\big\langle \bs{\mu},  d_{\bs{p}}\mathrm{Exp}_{\bs{p}}(\bs{v}) \bs{w} \big\rangle_{\mathrm{Exp}_{\bs{p}}(\bs{v})} = \big\langle \big(d_{\bs{p}}\mathrm{Exp}_{\bs{p}}(\bs{v})\big)^{\dag} \bs{\mu}, \bs{w} \big\rangle_{\bs{p}}$
with $\bs{\mu} \in T_{\mathrm{Exp}_{\bs{p}}(\bs{v})} \mathcal{M}$, $\bs{w} \in T_{\bs{p}} \mathcal{M}$. The adjoint derivative operator maps $\mathrm{Exp}_{\hat{\bs{y}}_i}^{-1}(\bs{y}_i^c)$ from the tangent space of $\hat{\bs{y}}_i$ to the tangent space of $\bs{p}$. Thus $\partial_{\bs{p}} E\in T_{\bs{p}} \mathcal{M}$. Similarly, the partial gradient of $E$ with respect to $\bs{v}_j$ and $\bs{g}_i$ are given by
\vspace{-10pt}
\begin{align}\label{grad_Ev}
\partial_{\bs{v}_j}E &= -\sum_i x_i^{j} \bigg( d_{\bs{v}}\mathrm{Exp}_{\bs{p}} \big( \sum_{j'} x_i^{j'} \bs{v}_{j'} \big)  \bigg)^{\dag} \mathrm{Exp}_{\hat{\bs{y}}_i}^{-1} \bs{y}_i^c \in T_{\bs{p}} \mathcal{M},
\end{align}
\vspace{-10pt}
and
\begin{align}\label{grad_Eg}
\partial_{\bs{g}_i}E &= - \bigg( d_{\bs{v}} \mathrm{Exp}_{\bs{y}_i}( \bs{g}_i) \bigg)^{\dag} \mathrm{Exp}_{\bs{y}_i^c}^{-1} \hat{\bs{y}}_i \in T_{\bs{y}_i} \mathcal{M},
\end{align}
respectively.

\begin{algorithm}[!t]
\caption{PALMR for multivariate regression with gross error on manifolds} \label{GR_Gross_Alg}
\begin{algorithmic}[1]
\REQUIRE $\{(\bs{x}_i,\bs{y}_i)\}$, $\lambda \geq 0$, $\rho\geq 0$, $\mu_1 > 1$, $\mu_2>1$, and $k=0$.
\ENSURE $\tilde{\bs{p}}$, $\{\tilde{\bs{v}}_j\}$, and $\{\tilde{\bs{g}}\}_i$.
\STATE Initialize $\bs{p}$, $\{\bs{v}_j\}$, and $\{\bs{g}\}_i$.
\WHILE{stopping criterion not satisfied}
  \STATE $\bs{p}^{k+1} = \mathrm{Exp}_{\bs{p}^k} \left( -\frac{1}{c_k}\partial_{\bs{p}}E^k \right)$.
  \STATE $\hat{\bs{v}}_j^k = \mathrm{prox}_{\frac{\lambda}{d_k}}^{R_v} (\bs{s}_j^k)$.
  \STATE $\bs{v}_j^{k+1} = P_{\bs{p}^{k}\bs{p}^{k+1}}(\hat{\bs{v}}_j^k)$.
  \STATE $\bs{g}_i^{k+1} = \mathrm{prox}_{\frac{\rho}{e_k}}^{R_g} (\bs{t}_i^k)$.
\ENDWHILE
\STATE \textbf{return} $\tilde{\bs{p}}\leftarrow \bs{p}^{k+1}$, $\tilde{\bs{v}}_j\leftarrow \bs{v}_j^{k+1}$ and $\tilde{\bs{g}}_i\leftarrow \bs{g}_i^{k+1}$.
\end{algorithmic}
\end{algorithm}


The PALMR algorithm for problem of Eq.~\eqref{GR_Gross_OP} proceeds as follows: To update $(\bs{p},\{\bs{v}_j\})$, we let $\partial_{\bs{p}}E^k := \partial_{\bs{p}}E(\bs{p}^k,\{\bs{v}_j^k\},\{\bs{g}_i^k\})$ and $\partial_{\bs{v}_j}E^k := \partial_{\bs{v}_j}E(\bs{p}^k,\{\bs{v}_j^k\},\{\bs{g}_i^k\})$ and solve
\begin{align*}
(\bs{p}^{k+1},\{\bs{v}_j^{k+1}\}) = \argmin_{\bs{p},\{\bs{v}_j\}}  & ~ \left\langle\mathrm{Exp}_{\bs{p}^k}^{-1} \bs{p}, \partial_{\bs{p}}E^k\right\rangle + \frac{c_k}{2} d^2(\bs{p},\bs{p}^k) \\
& + \sum_{j=1}^d \left(\left\langle P_{\bs{p}\bs{p}^k}(\bs{v}_j) - \bs{v}_j^k, \partial_{\bs{v}_j}E^k\right\rangle \right. \\
& \left. + \lambda\norm{ \bs{v}_j }_{\bs{p}} + \frac{c_k}{2} \|P_{\bs{p}\bs{p}^k}(\bs{v}_j) - \bs{v}_j^k\|^2 \right),
\end{align*}
where $P_{\bs{p}\bs{p}^k}$ is the parallel transport from $\bs{p}$ to $\bs{p}^k$ along the unique geodesic between them. {Due to the constraint $\bs{v}\in T_{\bs{p}}\mathcal{M}$, it is difficult to solve $\bs{p}$ and $\bs{v}_j$ together. Instead, the above subproblem is solved by alternating minimization over $\bs{p}$ and $\bs{v}_j$.} Specifically, to update $\bs{p}$, we solve
\[
\bs{p}^{k+1} = \mbox{arg}\min_{\bs{p}\in\mathcal{M}} \left\langle\mathrm{Exp}_{\bs{p}^k}^{-1} \bs{p}, \partial_{\bs{p}}E^k\right\rangle + \frac{c_k}{2} d^2(\bs{p},\bs{p}^k)
\]
which, by a change of variable $\bs{u}=\mathrm{Exp}_{\bs{p}^k}^{-1} \bs{p}$, is equivalent to solving
\[
\bs{u}^k = \argmin\limits_{\bs{u}\in T_{\bs{p}^k}\mathcal{M}} \left\langle \bs{u}, \partial_{\bs{p}}E^k\right\rangle + \frac{c_k}{2} \norm{ \bs{u} }_{\bs{p}^k}^2 = - \frac{1}{c_k}\partial_{\bs{p}}E^k,
\]
and $\bs{p}^{k+1} = \mathrm{Exp}_{\bs{p}^k}(\bs{u}^k)$.

To update $\{\bs{v}_j\}$, we need to first obtain $\hat{\bs{v}}_j^{k} $ by
{\small
\begin{align*}
\hat{\bs{v}}_j^{k} = & \argmin\limits_{\bs{v}_j \in T_{\bs{p}^{k}}\mathcal{M}} \left\langle \bs{v}_j-\bs{v}_j^k, \partial_{\bs{v}_j}E^k \right\rangle  + \frac{c_k}{2} \norm{ \bs{v}_j-\bs{v}_j^k }_{\bs{p}^{k}}^2 + \lambda \norm{ \bs{v}_j }_{\bs{p}^{k}} \\
 = & \argmin\limits_{\bs{v}_j \in T_{\bs{p}^{k}}\mathcal{M}} \frac{1}{2} \norm{ \bs{v}_j - \bs{s}_j^k }_{\bs{p}^{k}}^2 + \frac{\lambda}{c_k}\norm{ \bs{v}_j }_{\bs{p}^{k}},
\end{align*}
}
where $\bs{s}_j^k=\bs{v}_j^k - \frac{1}{c_k}\partial_{\bs{v}_j}E^k$. Notice that the above optimization problem have closed form solution of
\[
\hat{\bs{v}}_j^{k} = \mathrm{prox}_{\frac{\lambda}{c_k}}^{\norm{ \cdot }_{\bs{p}^{k}}} (\bs{s}_j^k) := \left(1-\frac{\lambda}{c_k \norm{ \bs{s}_j^k }_{\bs{p}^{k+1}}}\right)_{+}\bs{s}_j^k,
\]
where $(\alpha)_+ = \alpha$ if $\alpha > 0$ and $0$ otherwise. Since $\{\hat{\bs{v}}_j^{k}\}$ lie on the tangent space at $\bs{p}^{k}$, we need to parallel transport them to $T_{\bs{p}^{k+1}}\mathcal{M}$ by $\bs{v}_j^{k+1} = P_{\bs{p}^{k}\bs{p}^{k+1}}(\hat{\bs{v}}_j^k)$ along the unique geodesic between $\bs{p}^{k}$ and $\bs{p}^{k+1}$.

Similarly, update $\{\bs{g}_i\}$ by
\begin{align*}
\bs{g}_i^{k+1} = & \mbox{arg}\min\limits_{\bs{g}_i \in T_{\bs{y}_i}\mathcal{M}} \frac{1}{2} \norm{ \bs{g}_i - \bs{t}_i^k }_{\bs{y}_i}^2 + \frac{\rho}{e_k} \norm{ \bs{g}_i }_{\bs{y}_i} \\
 = & \left(1-\frac{\rho}{e_k \norm{ \bs{t}_i^k }_{\bs{y}_i}}\right)_{+}\bs{t}_i^k,
\end{align*}
where $\bs{t}_i^k=\bs{g}_i^k - \frac{1}{e_k}\partial_{\bs{g}_i}E(\bs{p}^{k+1},\{\bs{v}_j^{k+1}\},\{\bs{g}_i^k\})$.

Now, we are ready to present our algorithm for multivariate regression with grossly corrupted manifold-valued data, as shown in \textbf{Algorithm}~\ref{GR_Gross_Alg}. Notice that when letting $\lambda=0$ and $\rho=+\infty$, \textbf{Algorithm}~\ref{GR_Gross_Alg} alternately updates the values of $\bs{p}$ and $\bs{v}_j$ via three steps: (1) $\bs{p}^{k+1} = \mathrm{Exp}_{\bs{p}^k} \left( -\frac{1}{c_k}\partial_{\bs{p}}E^k \right)$, (2) $\hat{\bs{v}}_j^k = \bs{v}_j^k - \frac{1}{c_k}\partial_{\bs{v}_j}E^k$, (3) $\bs{v}_j^{k+1} = P_{\bs{p}^{k}\bs{p}^{k+1}}(\hat{\bs{v}}_j^k)$, which recovers the gradient descent method proposed in~\cite{KimEtAl:cvpr14}.
%

\subsection{Implementation of Algorithm~\ref{GR_Gross_Alg}}

\begin{figure}[!t]
\centering
\includegraphics[width=0.7\columnwidth, height=0.3\columnwidth]{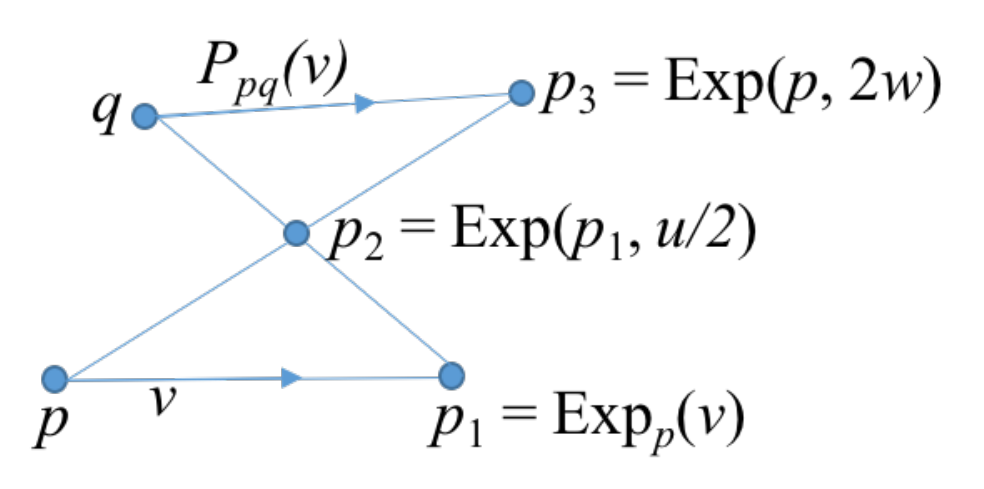}
\vspace{-10pt}
\caption{{An illustration of the Schild's ladder approximation of parallel transport of the tangent vector $\bs{v}$ from $\bs{p}$ to $\bs{q}$. It consists of four steps: (1) Obtain $\bs{p}_1$; (2) Compute tangent vector $\bs{u}=\text{Exp}_{\bs{p}_1}^{-1}(\bs{q})$ and take half step along $\bs{u}$ to arrive at $\bs{p}_2$; (3) Compute tangent vector $\bs{w}=\text{Exp}_{\bs{p}}^{-1}(\bs{p}_2)$ and take two steps along $\bs{w}$ to have $\bs{p}_3$; (4) Compute tangent vector joining $\bs{q}$ and $\bs{p}_3$ $P_{\bs{p}\bs{q}}(\bs{v}) = \text{Exp}_{q}^{-1}(\bs{p}_3)$. If the distance between $\bs{p}$ and $\bs{q}$ is large, the above process can be iterated over points along the geodesic path joining $\bs{p}$ and $\bs{q}$.}}
\label{SchildLadder}
\vspace{-10pt}
\end{figure}

{During each iteration of \textbf{Algorithm}~\ref{GR_Gross_Alg}, the partial derivatives $\partial_{\bs{p}}E$, $\partial_{\bs{v}_j}E$ and $\partial_{\bs{g}_i}E$ of Eqs.~\eqref{grad_Ep}, \eqref{grad_Ev}, and \eqref{grad_Eg} are evaluated. Their detailed derivations are provided in Section 3 of the supplementary file. Nevertheless, these terms could be practically intractable to compute for some manifolds, due to the presence of adjoint derivatives of the exponential map. As a remedy to this issue, we adopt the variational technique of~\cite{KimEtAl:cvpr14,KimEtAl:icml2015} for computing derivatives, which basically replaces the adjoint derivative operators by parallel transports:
{\small
\begin{align}\label{grad_Eapprx}
&\partial_{\bs{p}} E \approx -\sum_i P_{\hat{\bs{y}}_i\bs{p}}(\mathrm{Exp}_{\hat{\bs{y}}_i}^{-1} \bs{y}_i^c), \\
&\partial_{\bs{v}_j}E \approx -\sum_i x_i^{j} P_{\hat{\bs{y}}_i\bs{p}}(\mathrm{Exp}_{\hat{\bs{y}}_i}^{-1} \bs{y}_i^c), \\
&\partial_{\bs{g}_i}E \approx - P_{\bs{y}^c_i\bs{y}_i}(\mathrm{Exp}_{\bs{y}_i^c}^{-1} \hat{\bs{y}}_i).
\end{align}
}
One advantage of such approximation is that for some special manifolds, including manifold of SPD matrices $\mathcal{S}_{++}(n)$, parallel transports have analytical expressions and can be computed directly. For general manifolds that have no analytical expressions for parallel transports, approximation approaches such as Schild's ladder approximation~\cite{Kheyfets2000, LorenziP14} can be used. The method approximates parallel transport by constructing geodesic parallelograms, which requires three exponential maps and two inverse exponential maps, as shown in \figurename~\ref{SchildLadder}.}

\section{Experiments}\label{sec:exp}
In this section, we empirically evaluate the performance of the proposed approach (i.e. \emph{PALMR}) in working with synthetic and real DTI data sets, which lies in the $\mathcal{S}_{++}(3)$ manifold of SPD matrices.
Throughout all experiments, we fix $\lambda = 0.1$ and choose the optimal $\rho$ from set $\{0.05,0.1,\cdots, 0.95, 1\}$ by a validation process using a validation data set consisting of the same number of data points as the testing data. As our algorithm is iterative by nature, in practice it stops if either of the two stopping criteria is met: (1) the difference between consecutive objective function values is below 1e-5, or (2) maximum number of iterations (100) is reached.

\begin{figure}[!t]
\includegraphics[width=\columnwidth, height=0.6\columnwidth]{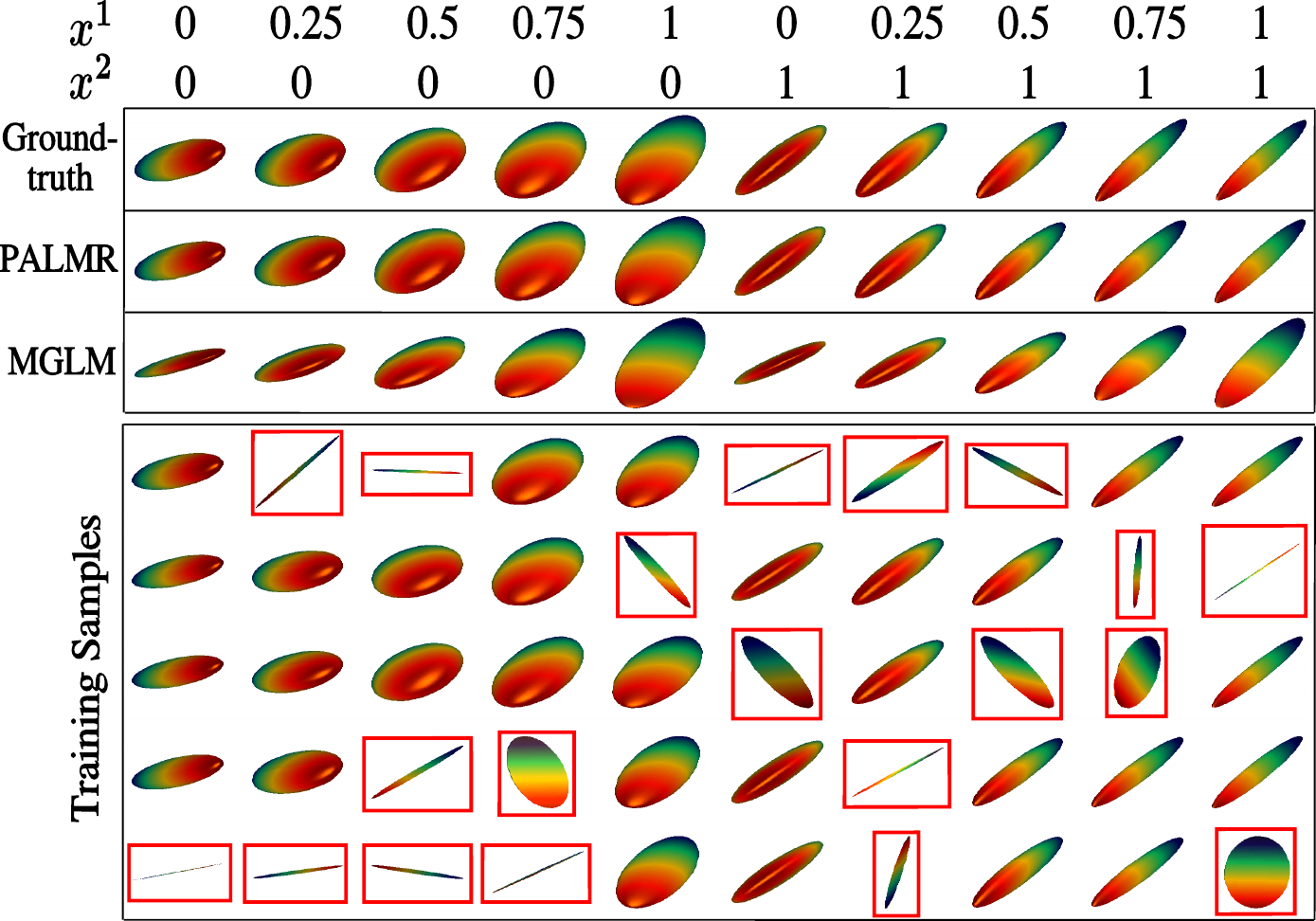}
\vspace{-5pt}
\caption{Visualization of the synthesized training samples and the predictions of PALMR and MGLM. The two row vectors on the top give the values of $X$ generating the data, red boxes identify the samples with gross error. The rows indexed by PALMR and MGLM display the predictions of corresponding method on the training data. All objects are viewed directly from overhead. Best viewed in color.}
\label{Vis_training}
\vspace{-10pt}
\end{figure}

\begin{figure*}[!t]
\begin{center}
\subfigure[Predictions of PALMR and MGLM on the testing data]{\includegraphics[width=\columnwidth, height=0.3\columnwidth]{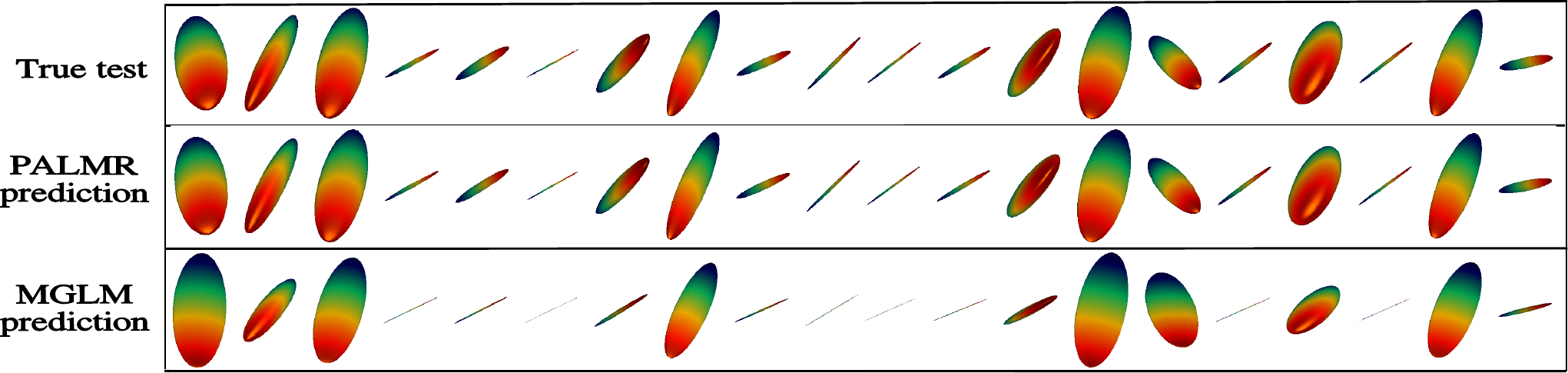}}  
\hspace{5pt}
\subfigure[Training data correction by our model]{\includegraphics[width=\columnwidth, height=0.3\columnwidth]{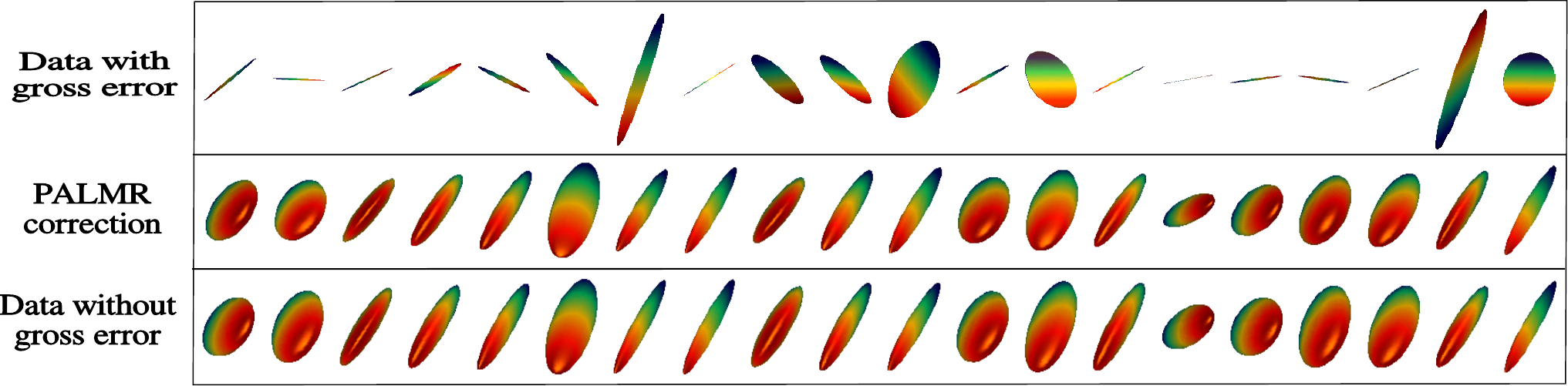}}
\end{center}
\vspace{-10pt}
\caption{Visual results of PALMR and MGLM. (a) Predictions for 20 testing data. (b) From top to bottom: training samples corrupted by gross error (i.e. samples marked by red boxes in \figurename~\ref{Vis_training}), correction results of PALMR, and the true data without gross error. Best viewed in color.}
\label{Vis_Result}
\vspace{-5pt}
\end{figure*}

\subsection{Synthetic DTI data}\label{synthetic}
Synthetic DTI data sets are constructed with known ground-truths and gross errors as follows:
First, we randomly generate $\bs{p} \in \mathcal{S}_{++}(3)$, symmetric matrices $\{\bs{v}_j\}_{j=1}^d \subseteq \mathcal{S}(3)$ and $\{\bs{x}_i\}_{i=1}^N \subseteq \bbR^{d}$ where entries of $\bs{x}_i$ are sampled from standard normal distribution $\mathcal{N}(0,1)$. Then the ground-truth DTI data is obtained as $\bs{y}^t_{i} := \text{Exp}_{\bs{p}}\left(\sum_{j=1}^d x_i^j\bs{v}_j\right)$. This is followed by DTI data with stochastic noise as $\bs{y}^s_{i} := \text{Exp}_{\bs{y}^t_{i}}\left(\bs{z}_i\right)$, where $\bs{z}_i$ is a random matrix in $\mathcal{S}(3)$ with its entries being sampled from $\mathcal{N}(0,1)$ and satisfies $\norm{ \bs{z}_i }_{\bs{y}^t_{i}} \leq 0.1$. Meanwhile, the gross errors are generated by a two-step process: (a) Randomly select an index subset $I_g$ from $\{1,2,\cdots,N\}$, such that $|I_g| = \beta*N$ with $0 \leq \beta \leq 1$. (b) For $i\in I_g$, its grossly corrupted response is attained by $\bs{y}_i = \text{Exp}_{\bs{y}^s_{i}}\left(\bs{g}_i \right)$, where $\bs{g}_i$ is a random matrix in $\mathcal{S}(3)$ satisfying $\norm{ \bs{g}_i }_{\bs{y}^s_{i}} = \sigma_g$. The rest of the training data remain unchanged, i.e. $\bs{y}_i = \bs{y}^s_{i}$ for $i\notin I_g$ . Thus, among all $N$ manifold-valued data, the percentage of grossly corrupted data is $\beta$. With the same $\bs{p}$ and $\{\bs{v}_j\}$, we also generate $N_t$ pairs of testing data $\{(\bs{x}^{test}_i, \bs{y}^{test}_i)\}$ and validation data.

\begin{figure*}[!t]
\centering
\subfigure{\includegraphics[width = 0.5\textwidth]{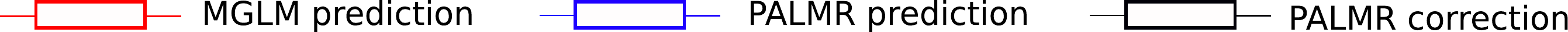}} \\ 
\begin{subfigure}
    \centering
    \hspace{7pt}
    \makebox[0pt][r]{\makebox[30pt]{\raisebox{35pt}{\rotatebox[origin=c]{90}{$\text{MSGE}_{train}$}}}}%
    \includegraphics[width = 0.22\textwidth, height = 0.14\textwidth]{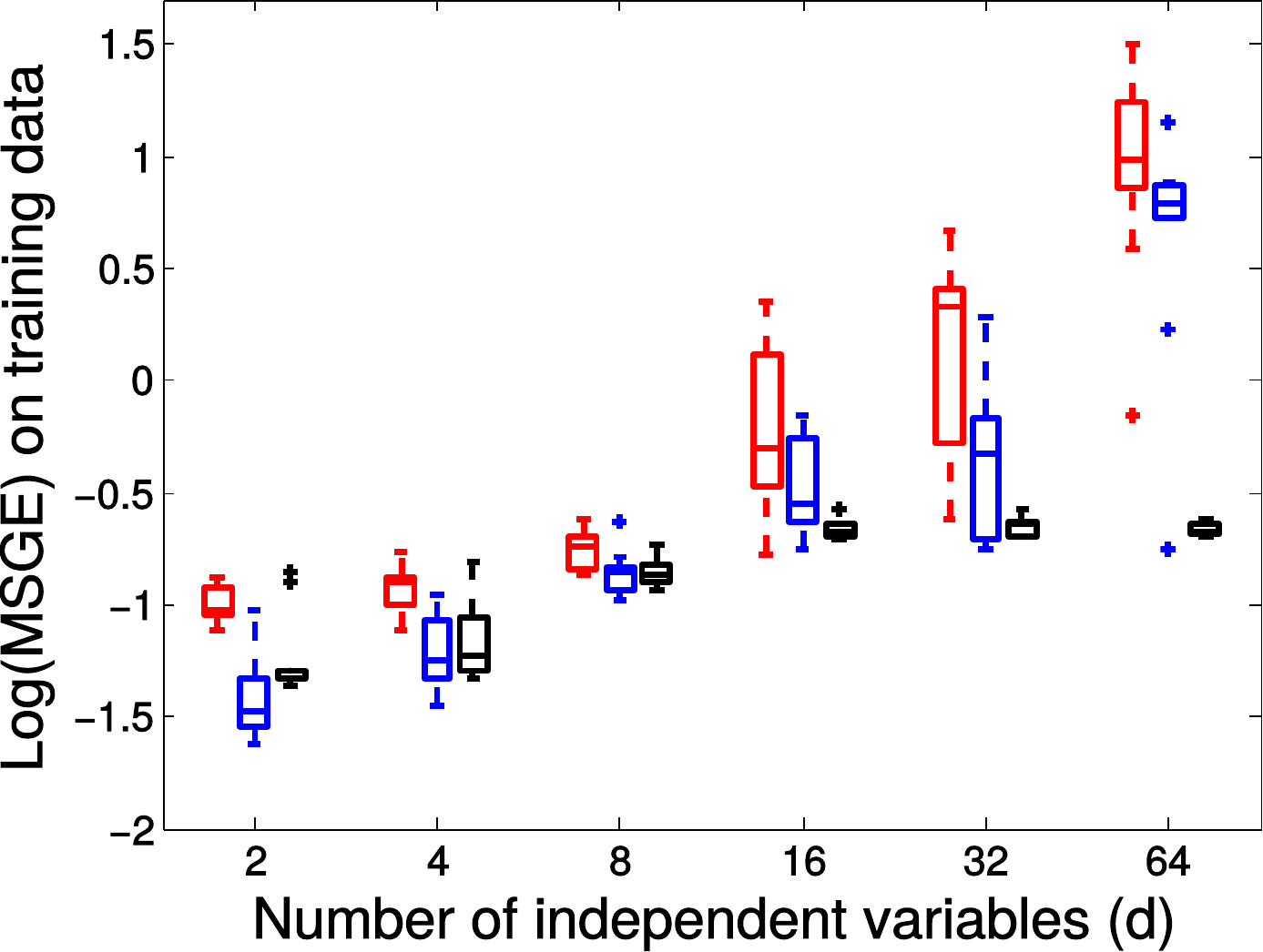} \hspace{10pt}
    \includegraphics[width = 0.22\textwidth, height = 0.14\textwidth]{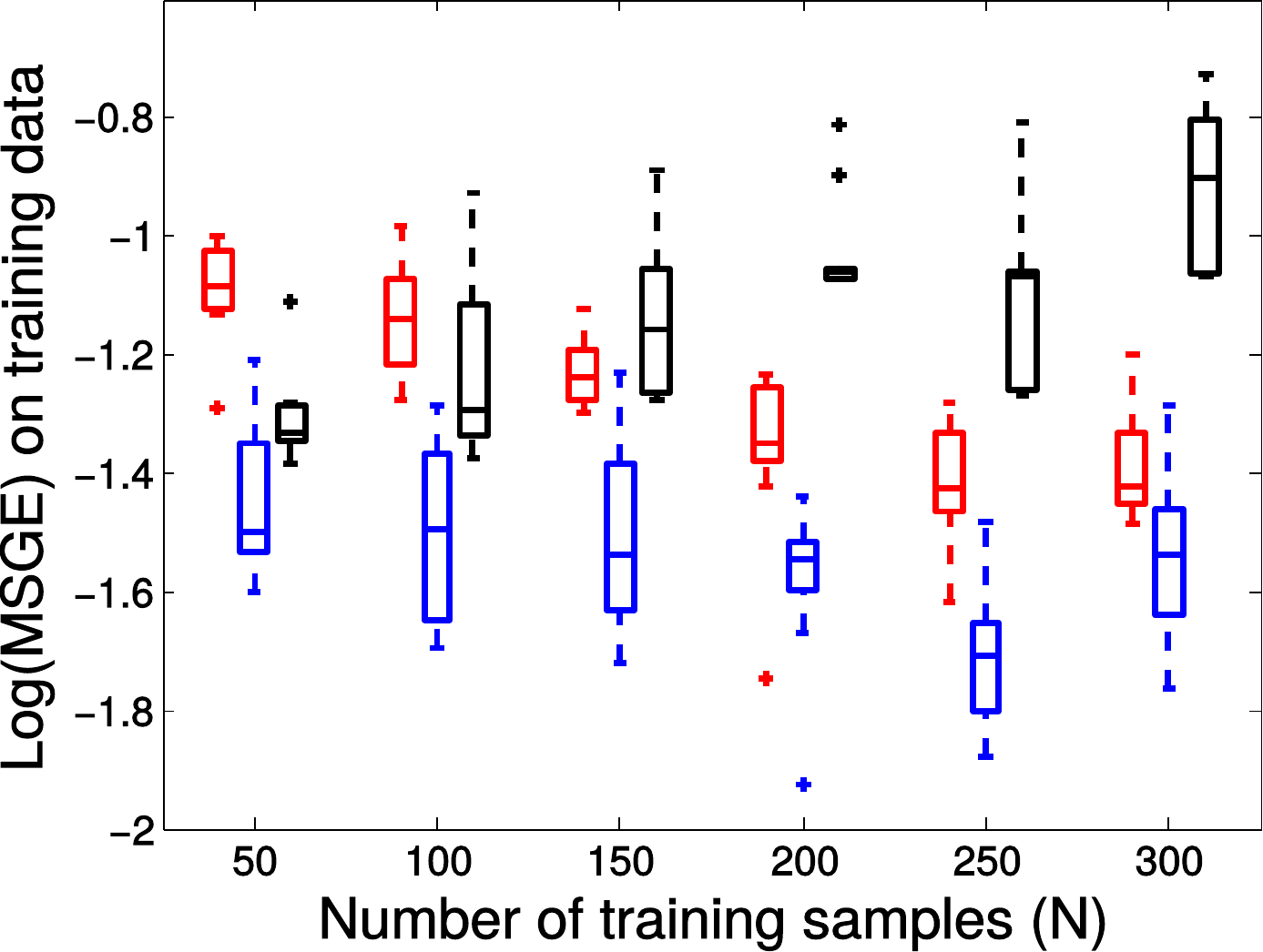} \hspace{10pt}
    \includegraphics[width = 0.22\textwidth, height = 0.14\textwidth]{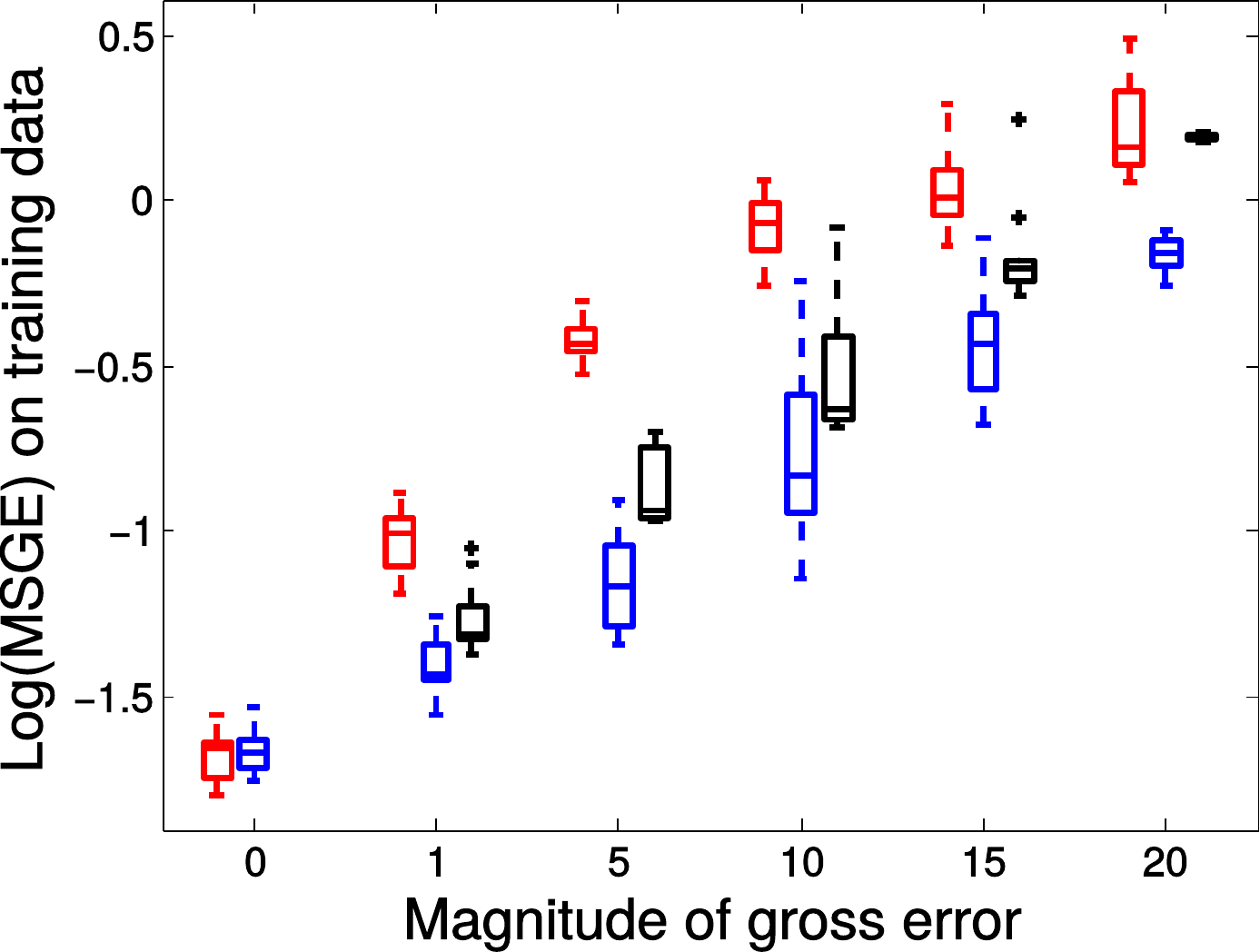} \hspace{10pt}
    \includegraphics[width = 0.22\textwidth, height = 0.14\textwidth]{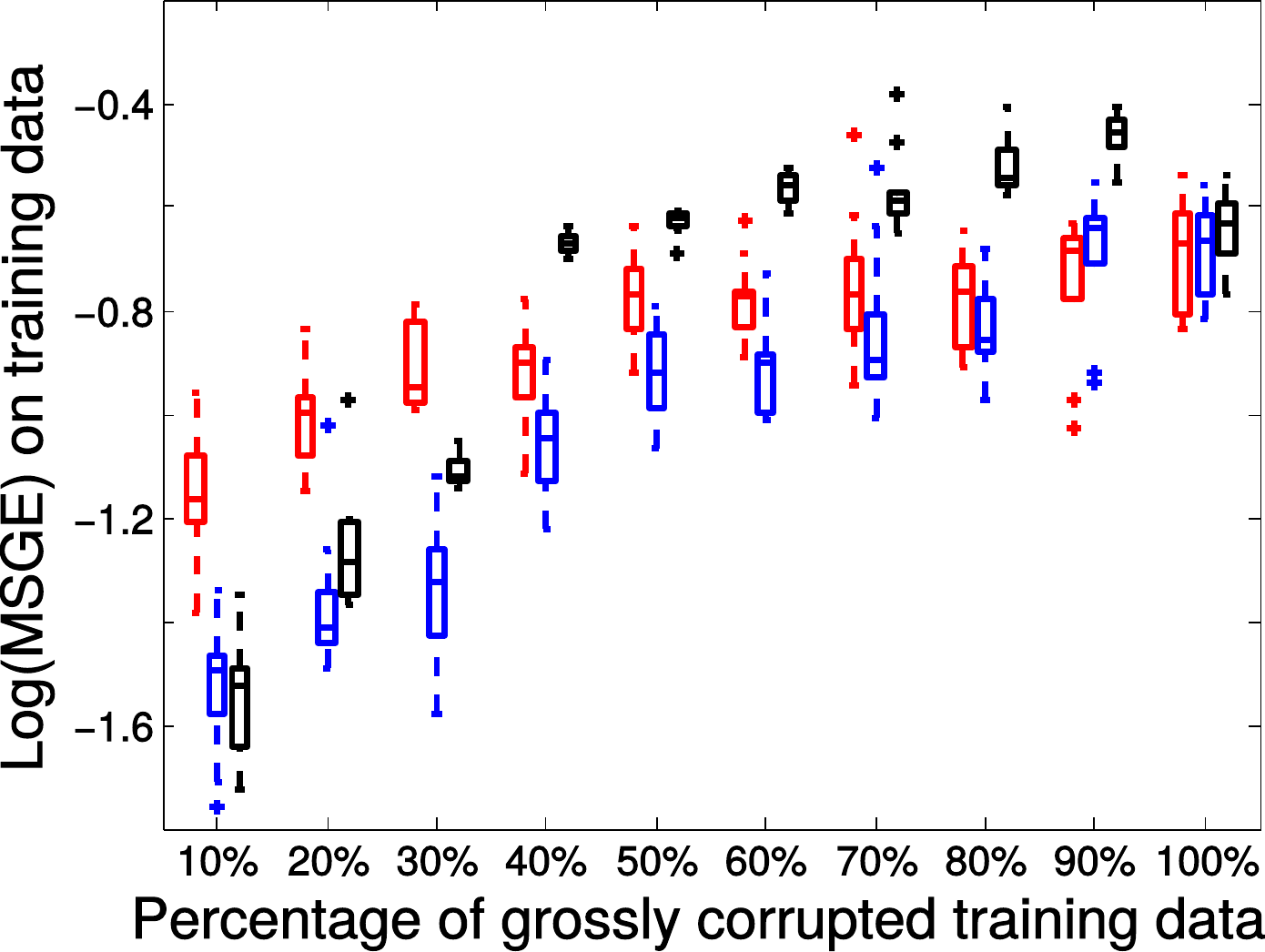}
\end{subfigure}\\
\begin{subfigure}
    \centering
    \hspace{7pt}
    \makebox[0pt][r]{\makebox[30pt]{\raisebox{35pt}{\rotatebox[origin=c]{90}{$\text{MSGE}_{test}$}}}}%
    \includegraphics[width = 0.22\textwidth, height = 0.14\textwidth]{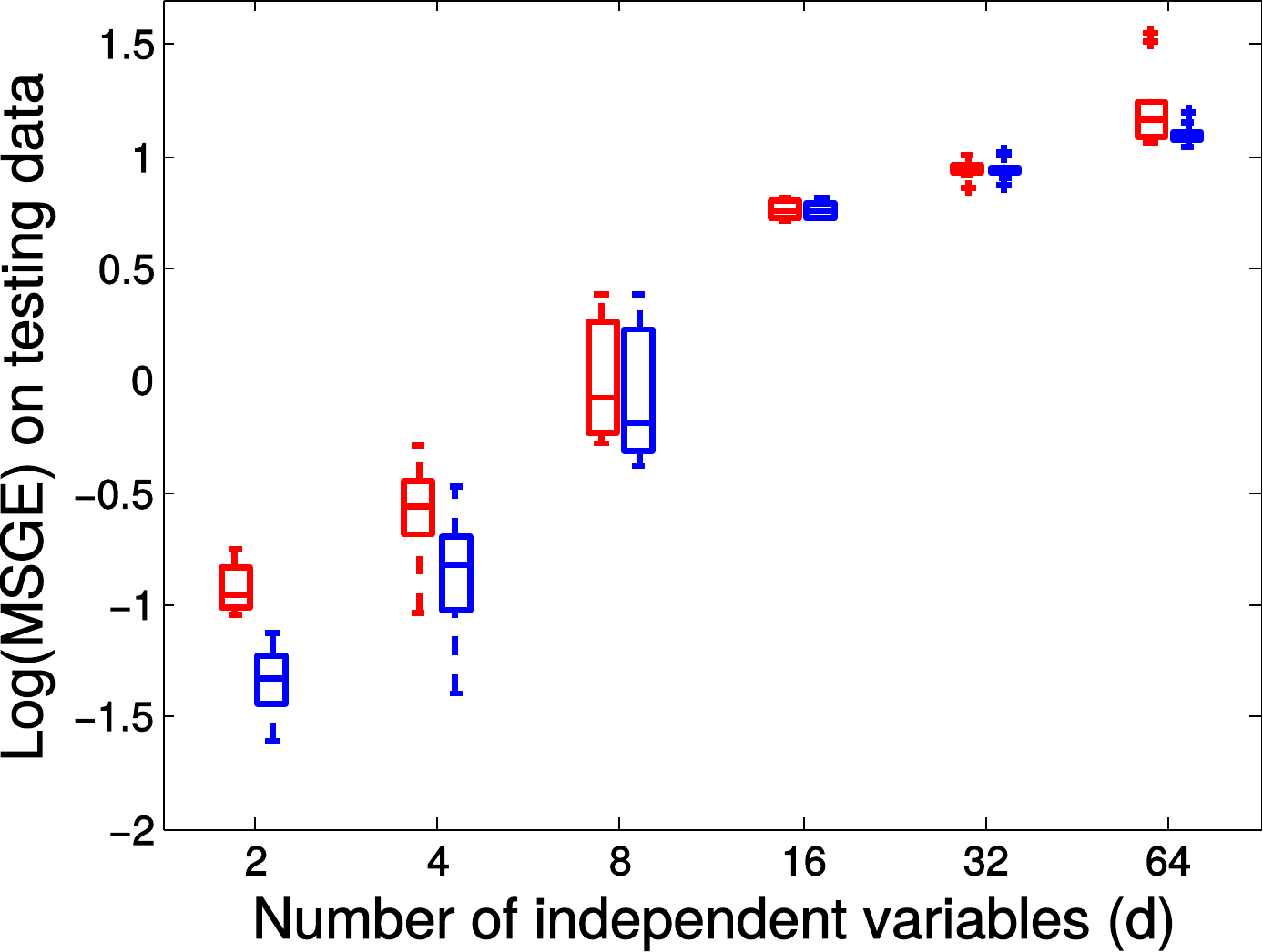} \hspace{10pt}
    \includegraphics[width = 0.22\textwidth, height = 0.14\textwidth]{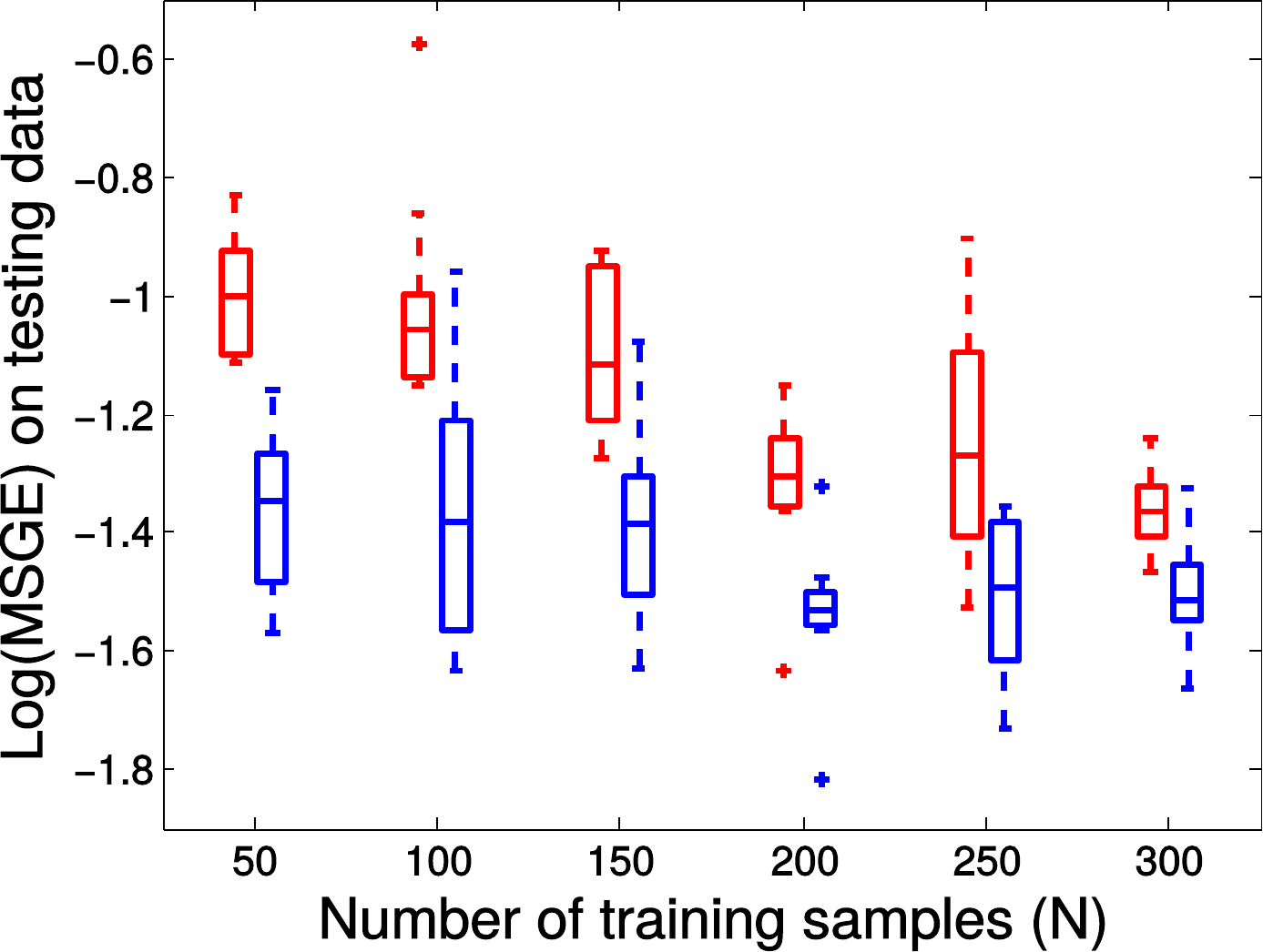} \hspace{10pt}
    \includegraphics[width = 0.22\textwidth, height = 0.14\textwidth]{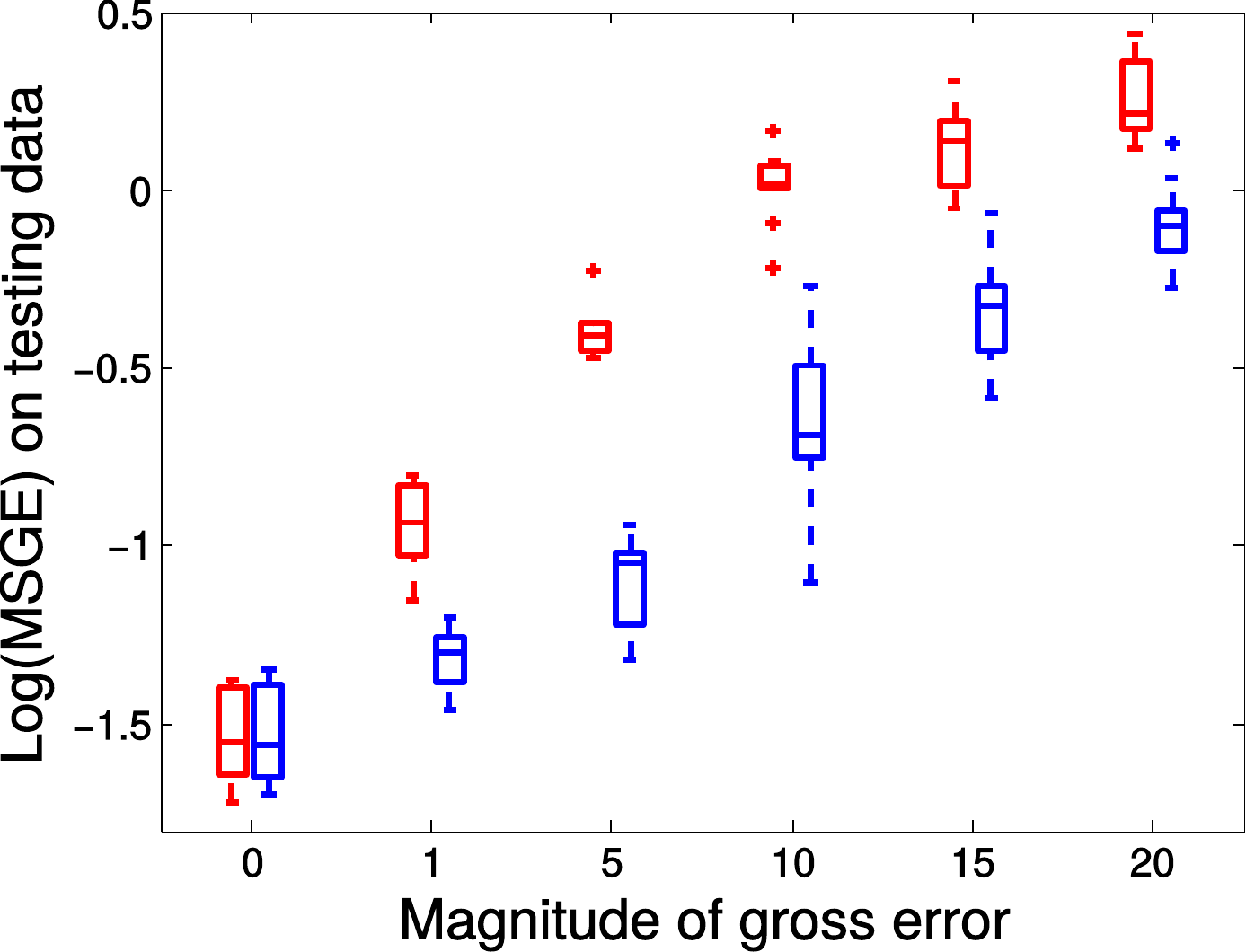} \hspace{10pt}
    \includegraphics[width = 0.22\textwidth, height = 0.14\textwidth]{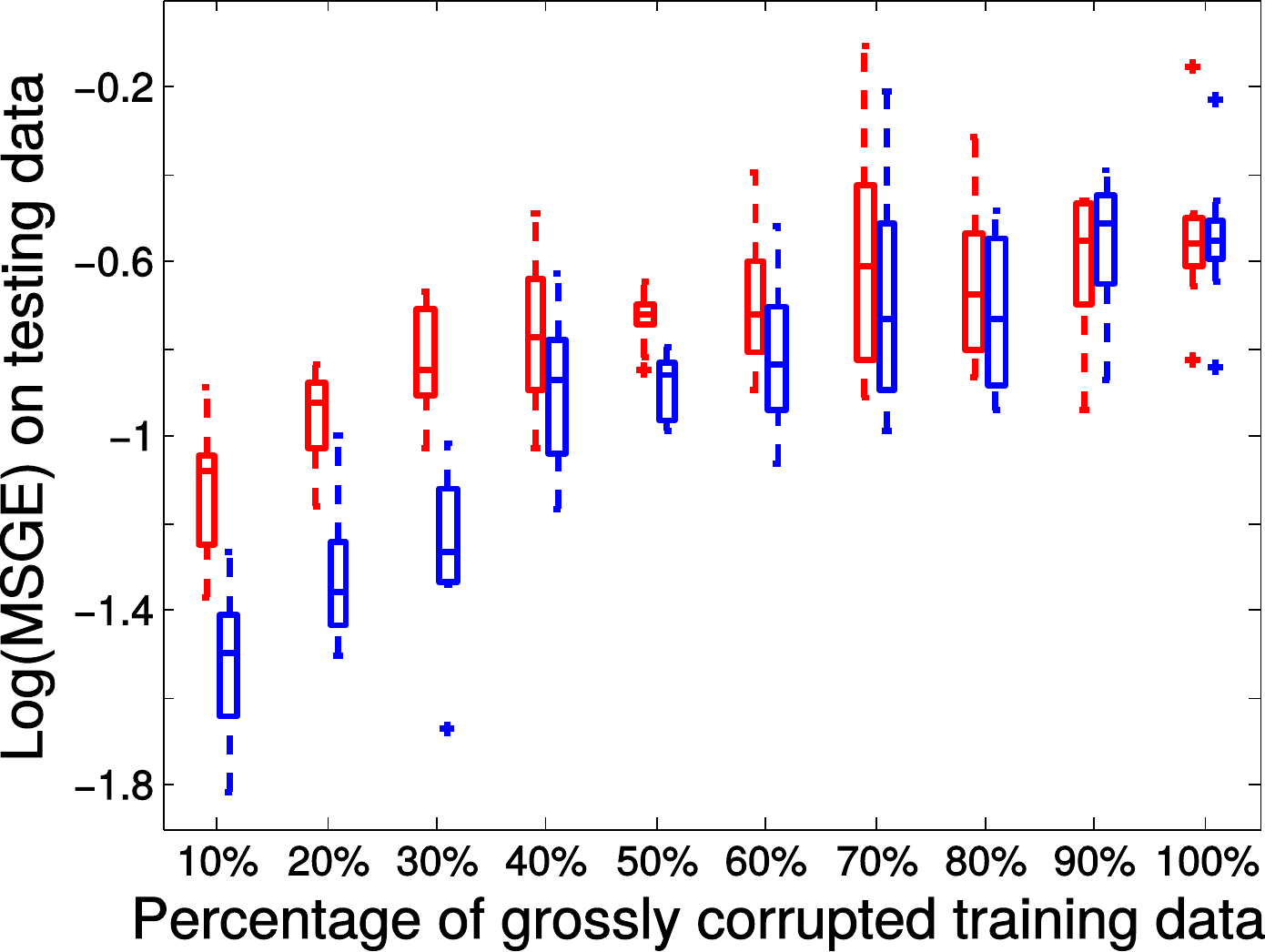}
\end{subfigure}\\
\begin{subfigure}
    \centering
    \hspace{7pt}
    \makebox[0pt][r]{\makebox[30pt]{\raisebox{35pt}{\rotatebox[origin=c]{90}{$\text{MSGE}_{\bs{p}}$}}}}%
    \includegraphics[width = 0.22\textwidth, height = 0.14\textwidth]{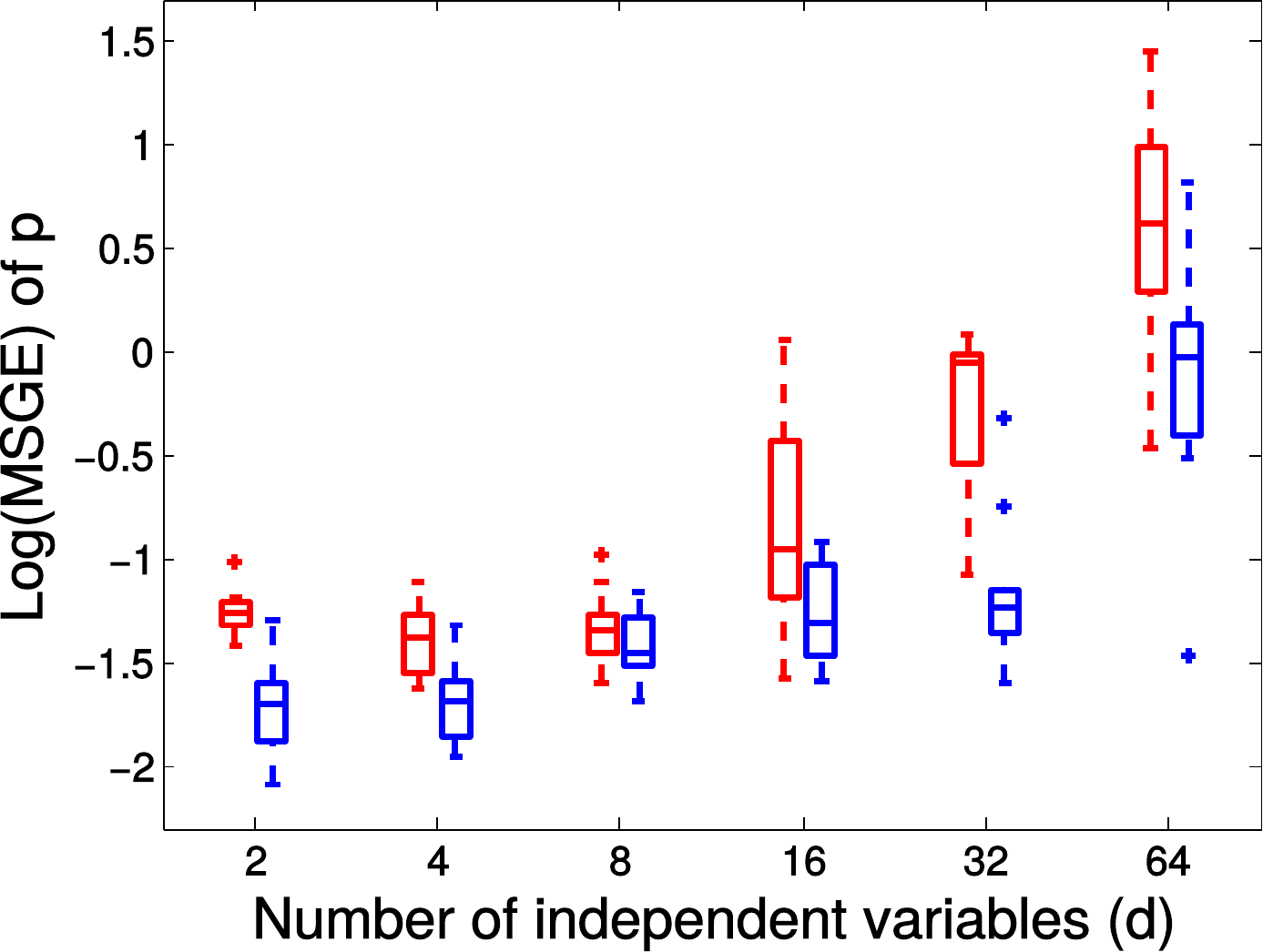} \hspace{10pt}
    \includegraphics[width = 0.22\textwidth, height = 0.14\textwidth]{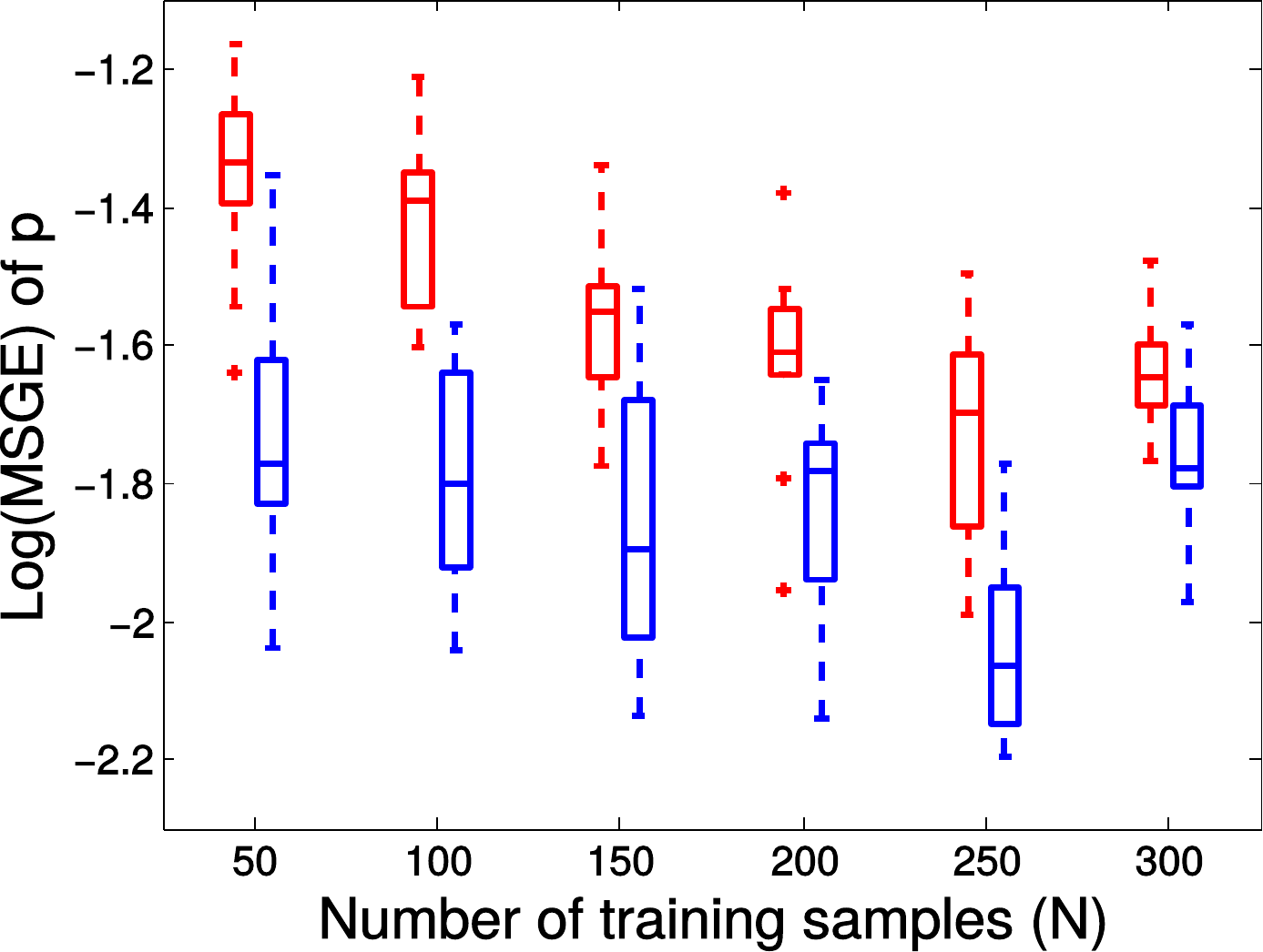} \hspace{10pt}
    \includegraphics[width = 0.22\textwidth, height = 0.14\textwidth]{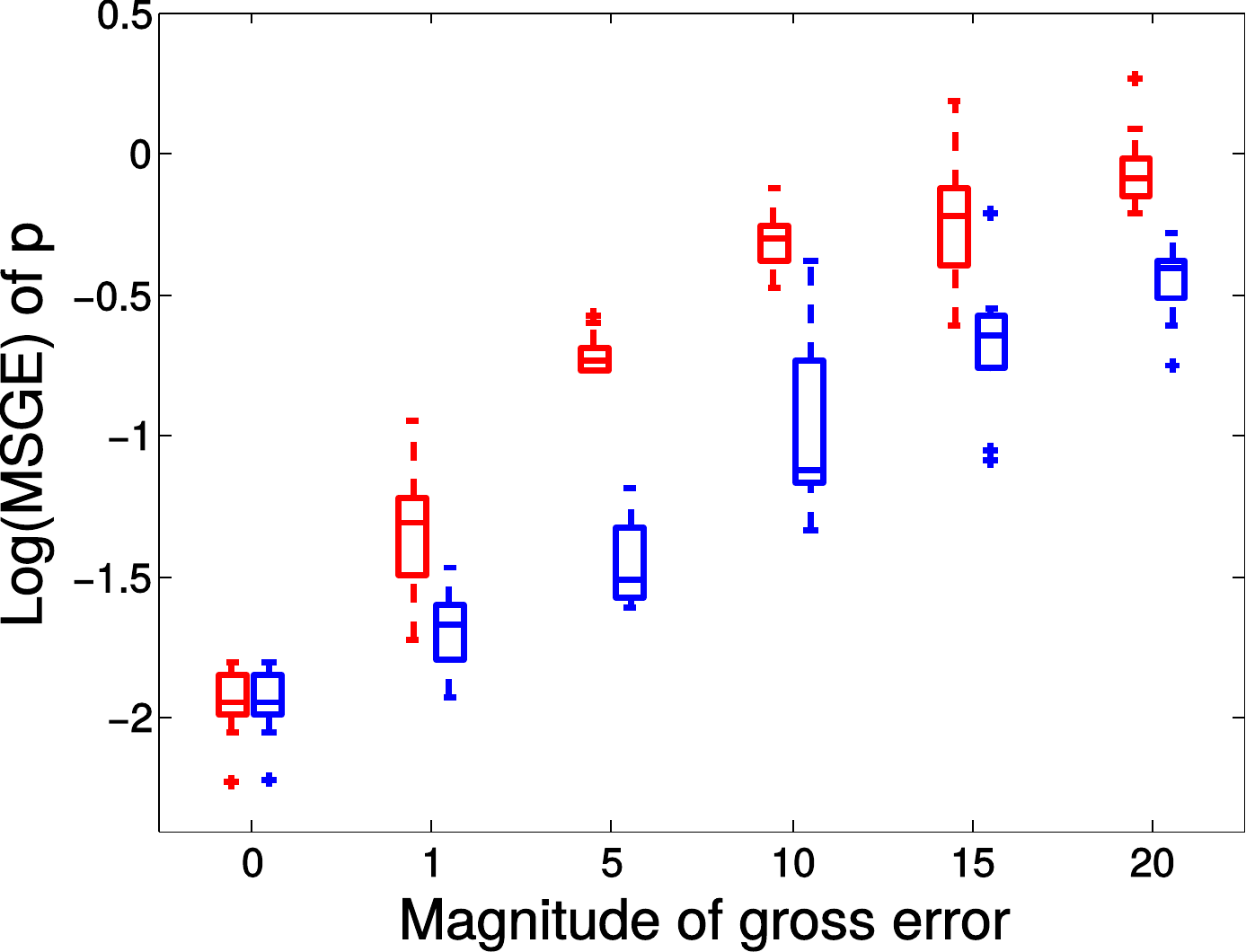} \hspace{10pt}
    \includegraphics[width = 0.22\textwidth, height = 0.14\textwidth]{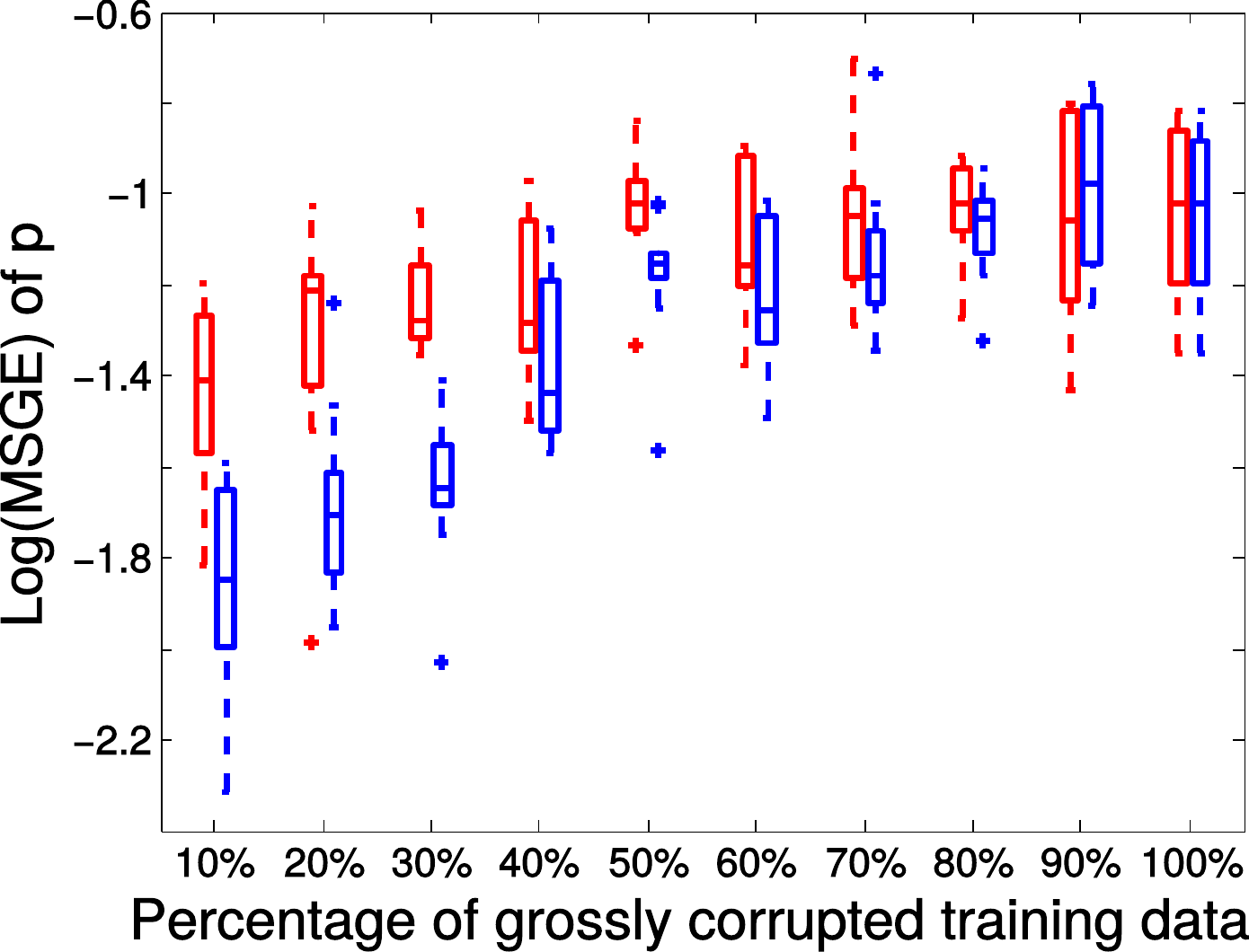}
\end{subfigure}\\
\begin{subfigure}
    \centering
    \hspace{7pt}
    \makebox[0pt][r]{\makebox[30pt]{\raisebox{35pt}{\rotatebox[origin=c]{90}{$\text{MSGE}_{V}$}}}}%
    \includegraphics[width = 0.22\textwidth, height = 0.14\textwidth]{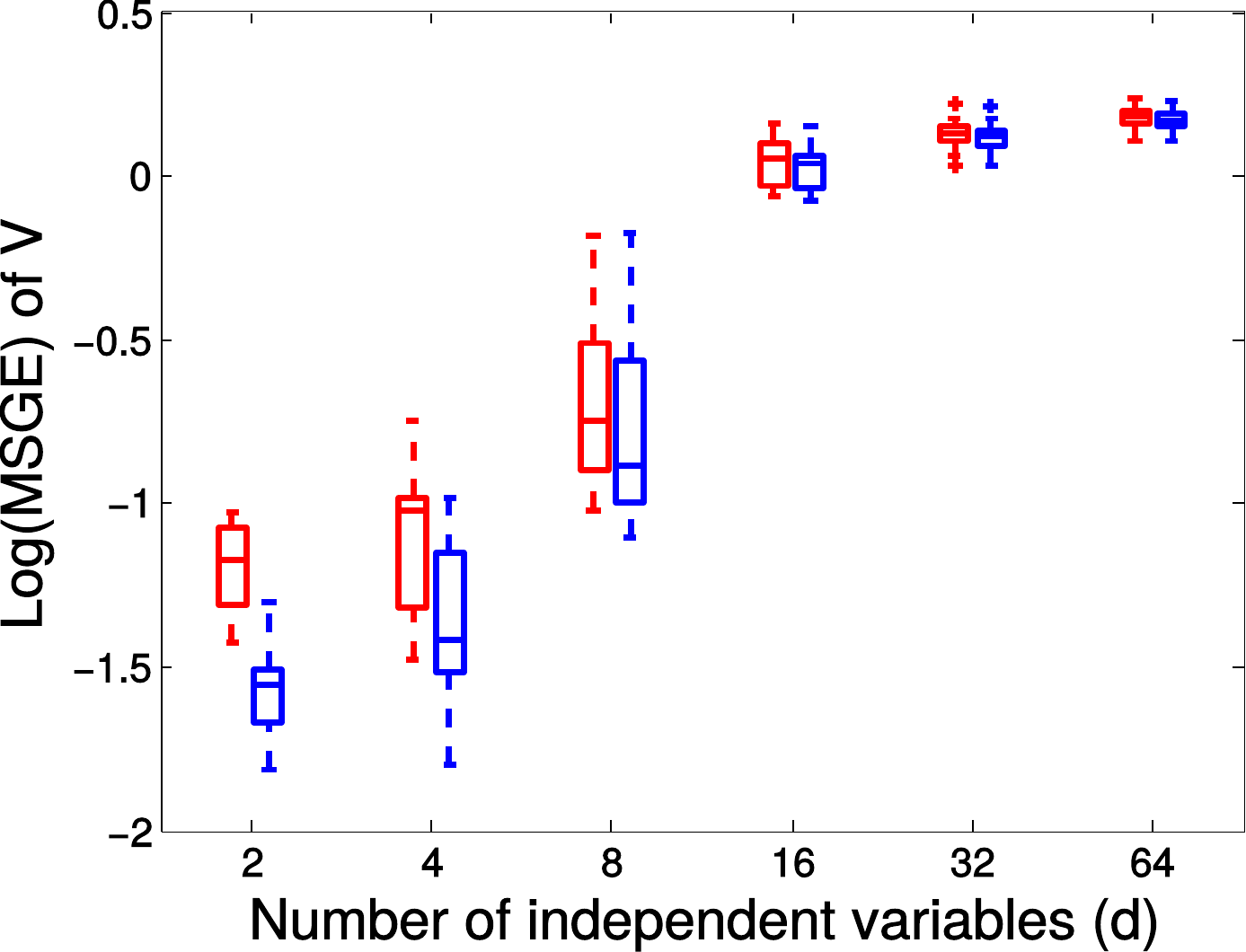} \hspace{10pt}
    \includegraphics[width = 0.22\textwidth, height = 0.14\textwidth]{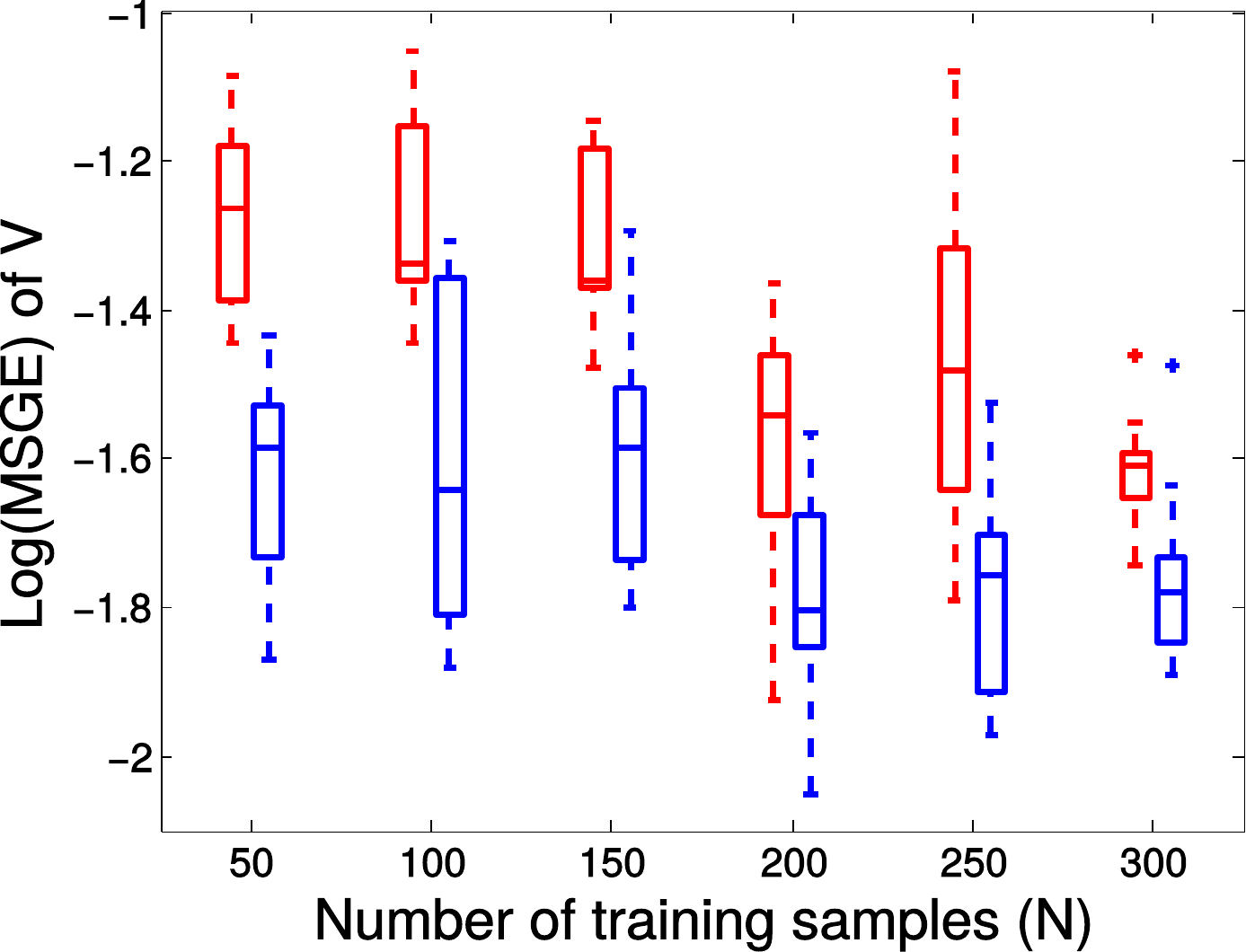} \hspace{10pt}
    \includegraphics[width = 0.22\textwidth, height = 0.14\textwidth]{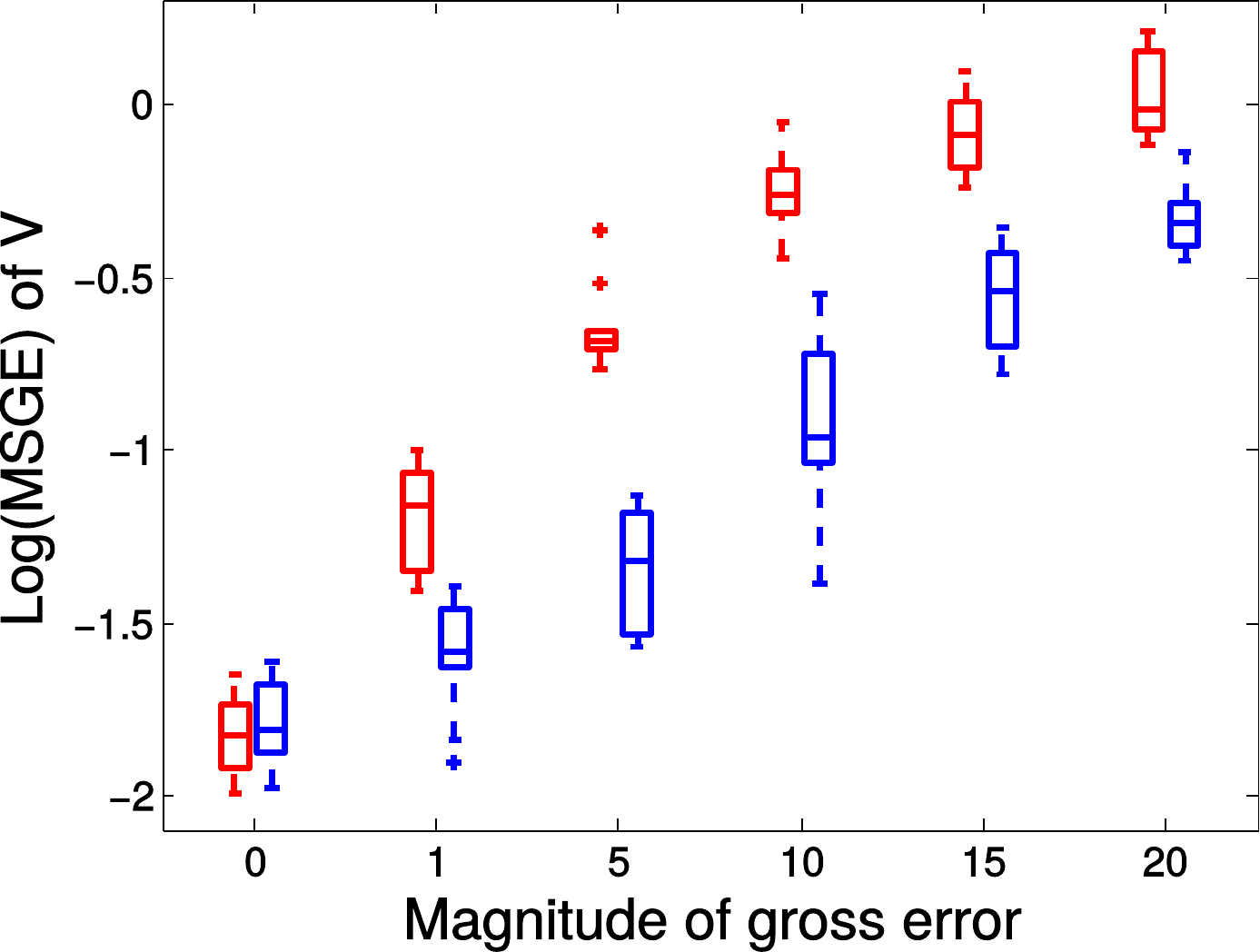} \hspace{10pt}
    \includegraphics[width = 0.22\textwidth, height = 0.14\textwidth]{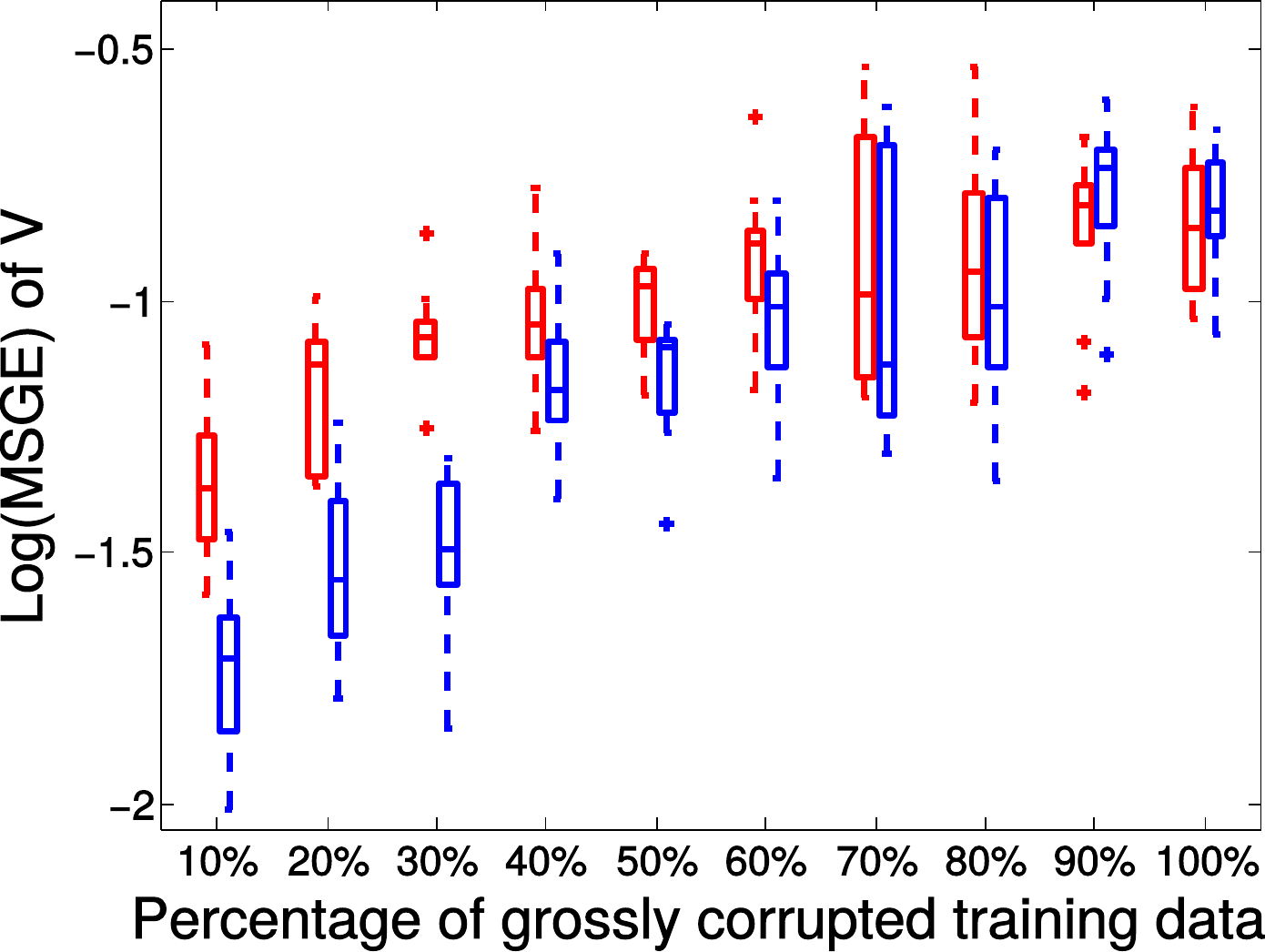}
\end{subfigure}\\
\setcounter{subfigure}{0}
\hspace{7pt}
\subfigure[\scriptsize Number of independent variables $d$]{
\makebox[0pt][r]{\makebox[30pt]{\raisebox{45pt}{\rotatebox[origin=c]{90}{$Rate_G$\quad $\text{MSGE}_{G}$}}}}%
\includegraphics[width = 0.22\textwidth]{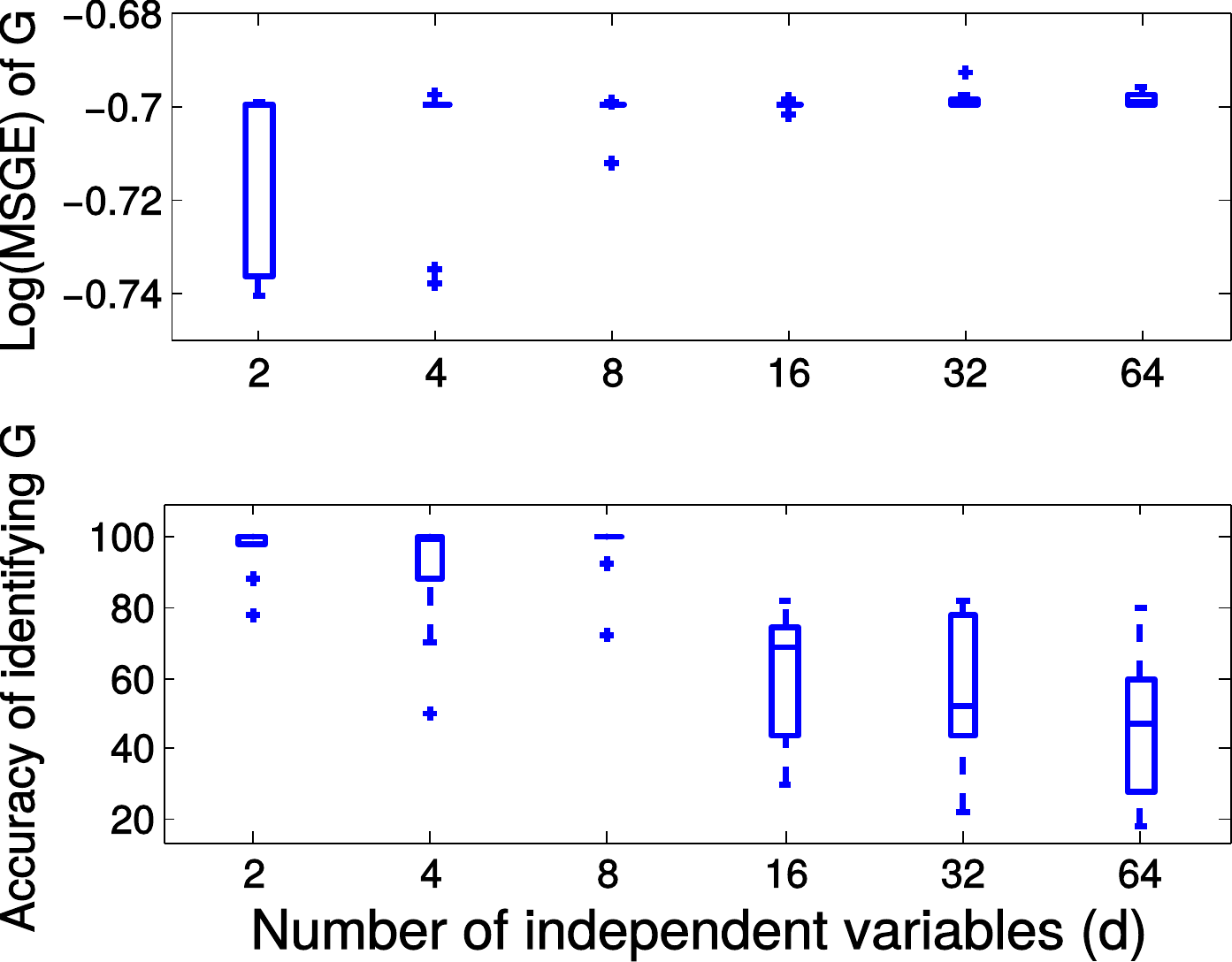}
} \hspace{6pt}
\subfigure[\scriptsize Training size $N$]{\includegraphics[width = 0.22\textwidth]{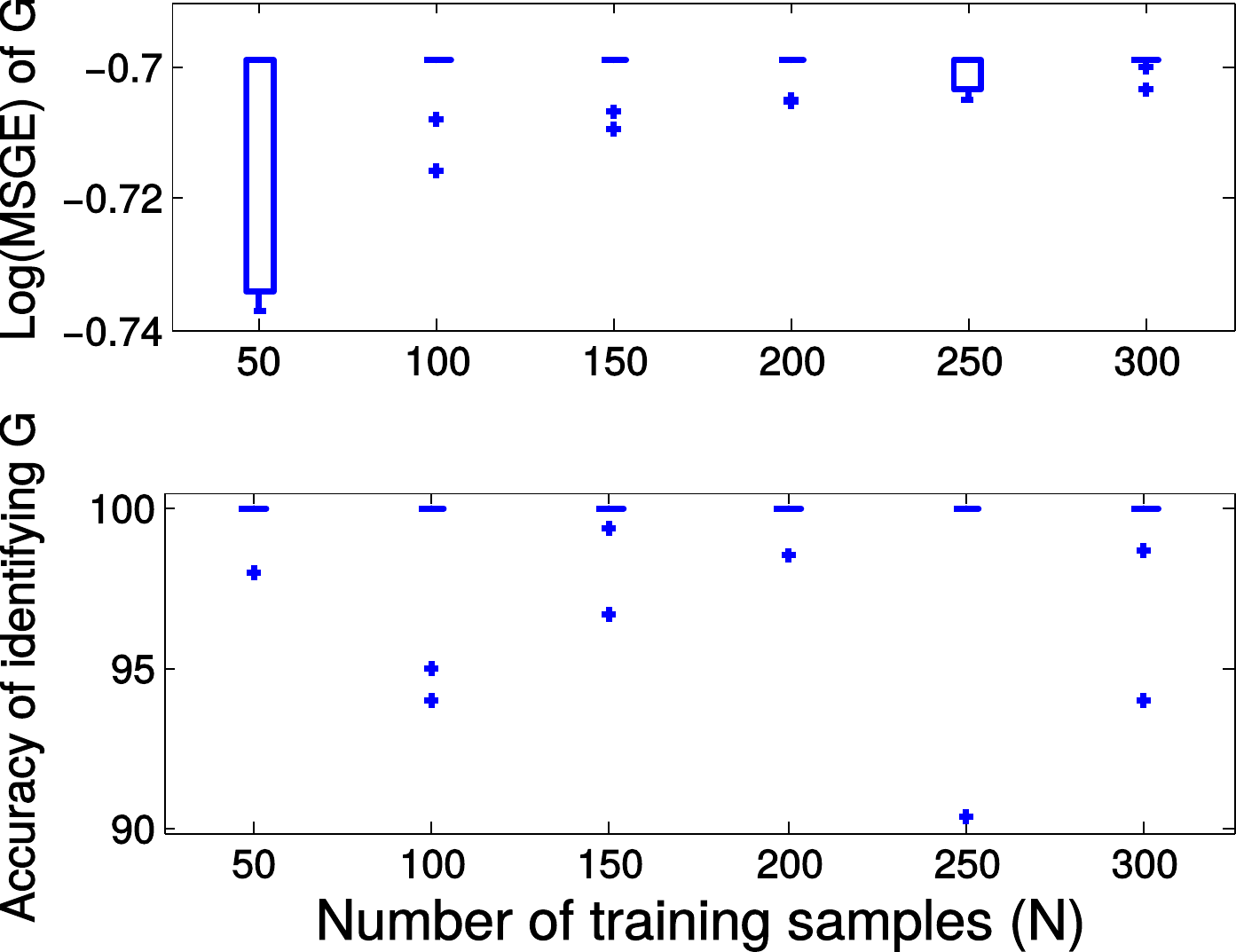}
} \hspace{6pt}
\subfigure[\scriptsize Magnitude of gross error $\sigma_g$]{\includegraphics[width = 0.22\textwidth]{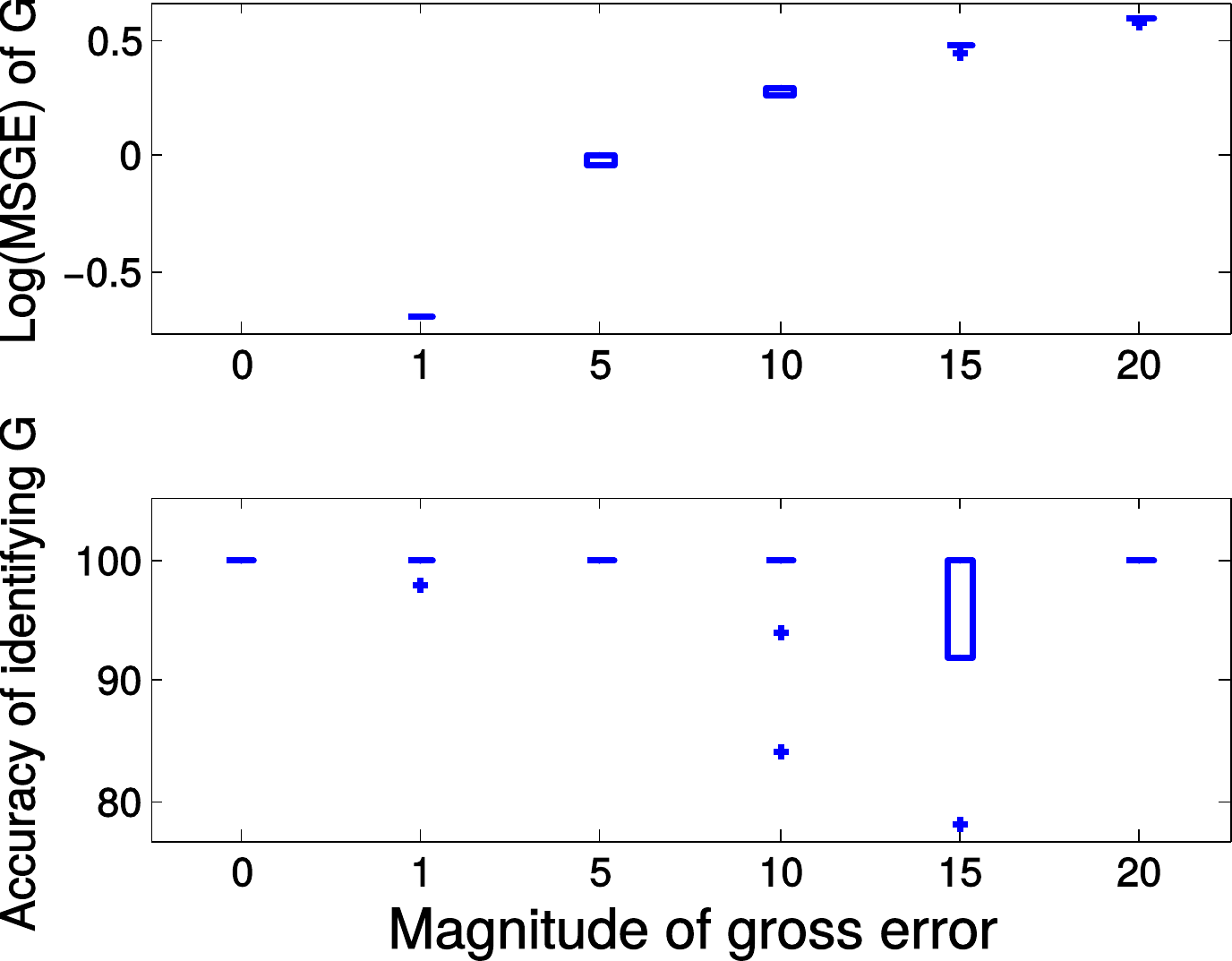}
} \hspace{6pt}
\subfigure[\scriptsize Ratio of corrupted data $\beta$]{\includegraphics[width = 0.22\textwidth]{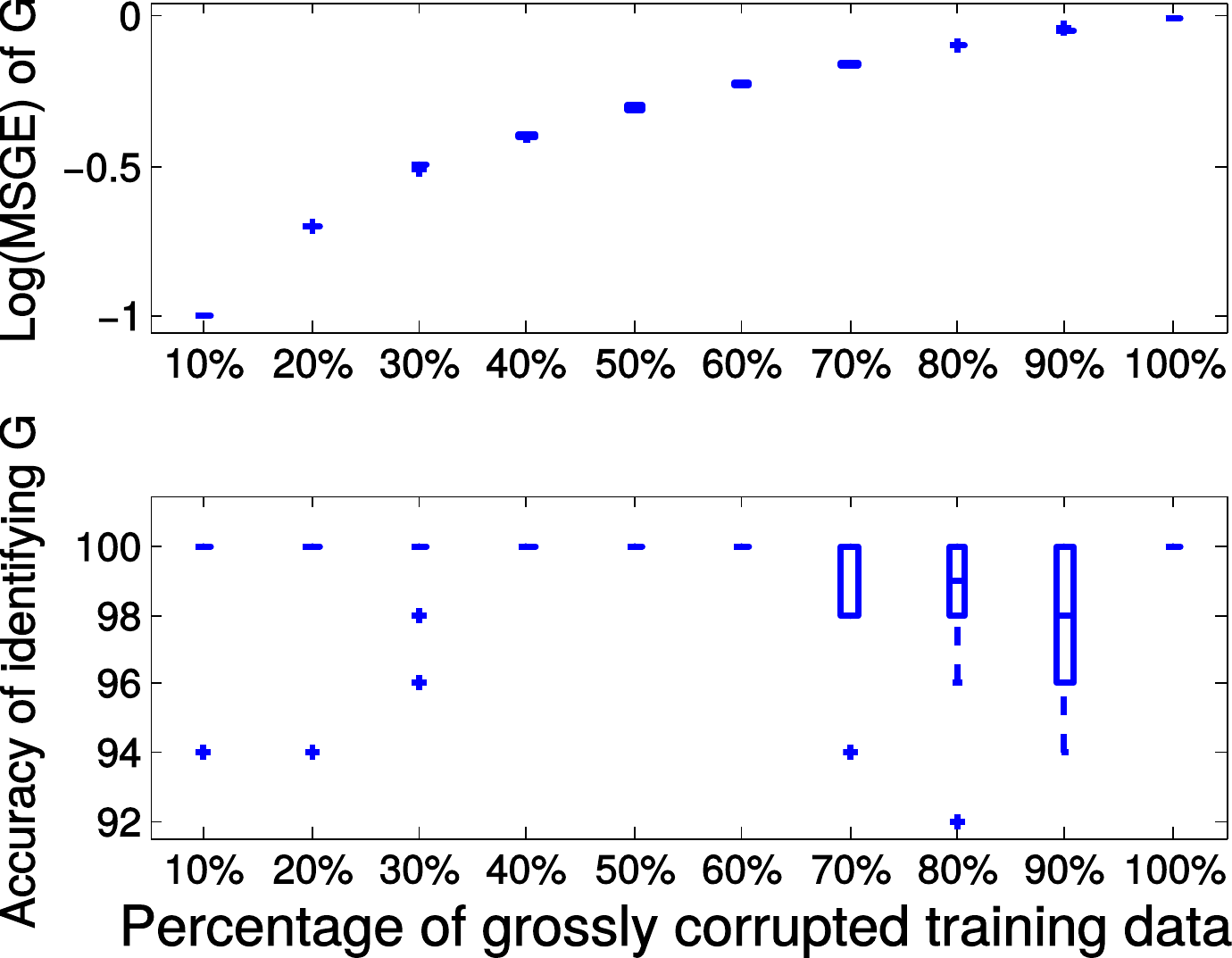}
}
\vspace{-10pt}
\caption{Box plots showing the effect of four different parameters: the number of tangent basis $d$, the number of training data $N$, the magnitude of gross error $\sigma_g$, and the percentage of grossly corrupted training data $\beta$, corresponding to four columns accordingly. Plots in each row show results using the same metric. Since MGLM does not consider gross error, the last two rows only show results of PALMR. See text for details. Best viewed in color.}
\label{Effect_para}
\vspace{-10pt}
\end{figure*}

We first conduct experiments on a data set with $d=2$, $N=50$, $\beta = 40\%$ and $\sigma_g = 5$, and compared with the multivariate general linear model (MGLM) of~\cite{KimEtAl:cvpr14} which has not considered gross error. Training samples are displayed in \figurename~\ref{Vis_training}, where we also show the predictions of PALMR and MGLM on the training data. Visual results of PALMR and MGLM on 20 testing data and training data correction by PALMR are presented in \figurename~\ref{Vis_Result}(a) and \figurename~\ref{Vis_Result}(b), respectively. Collectively, the results suggest that PALMR indeed is capable of correctly identifying the gross errors during training. This enables the delivery of a better-behaved model. \figurename~\ref{Vis_Result}(b) shows that PALMR can effectively recover the original data (i.e. true data without gross error). It also produces improved regression results on testing data as displayed in \figurename~\ref{Vis_Result}(a).

Next we quantitatively evaluate the effect of varying the internal parameters of PALMR, which include the number of independent variables $d$, the number of training data $N$, magnitude of gross error $\sigma_g$, and percentage of grossly corrupted training data $\beta$. To see the effect of one specific parameter, synthetic DTI data are generated by varying this parameter value while keeping rest parameters at their default values. The following default values are used: $d=2$, $N=50$, $\beta=20\%$, and $\sigma_g=1$. {To evaluate performance of PALMR, the following mean squared geodesic error (MSGE) metrics are considered: $\text{MSGE}_{train} := \frac{1}{N}\sum_{i}d^2(\bs{y}_i,\hat{\bs{y}}_i)$, $\text{MSGE}_{test} := \frac{1}{N_t}\sum_{i}d^2(\bs{y}^{test}_i,\hat{\bs{y}}^{test}_i)$, $\text{MSGE}_{\bs{p}} :=d^2(\bs{p}, \tilde{\bs{p}})$, $\text{MSGE}_{V} :=\frac{1}{d}\sum_{j}\|\bs{v}_j - P_{\tilde{\bs{p}}\bs{p}}(\tilde{\bs{v}}_j)\|_{\bs{p}}^2$, and $\text{MSGE}_{G} :=\frac{1}{N}\sum_{i}\|\mbox{Exp}_{\bs{y}_i}^{-1}(\bs{y}^s_i) - \tilde{\bs{g}}_i\|_{\bs{y}_i}^2$,
where $\tilde{\bs{p}}$, $\tilde{\bs{v}}_j$ and $\tilde{\bs{g}}_i$ are the outputs of \textbf{Algorithm}~\ref{GR_Gross_Alg}.} The data correction error is measured as  $\frac{1}{N}\sum_{i}d(\bs{y}^s_i,\bs{y}^c_i)^2$. In addition, we say that gross error $\bs{g}_i$ is correctly identified if both $\bs{g}_i$ and $\tilde{\bs{g}}_i$ are either zero or nonzero, and compute the rate $Rate_G:= \mbox{\textit{number of correctly identified gross errors}} / N$. Results averaged over 10 repetitions are presented in \figurename~\ref{Effect_para}, where each column corresponds to the effect of one parameter and each row corresponds to the results using one metric.

From \figurename~\ref{Effect_para}, we have four observations: (1) PALMR has lower MSGE for all values of $d$, and our correction performs well on training data, cf. column \figurename~\ref{Effect_para}(a). (2) PALMR has large advantage over MGLM for all values of training size ($N$) and magnitude of gross error ($\sigma_g$), cf. columns \figurename~\ref{Effect_para}(b-c). (3) PALMR can handle training data with up to $80\%$ being grossly corrupted, and delivers better result than MGLM. On the other hand, the performance is slightly worse if more than $80\%$ of training data are corrupted, cf. column \figurename~\ref{Effect_para}(d). (4) PALMR can reliably identify most of the gross errors. Still it may not always correctly recover the true value of the error. This is evidenced in the last row of \figurename~\ref{Effect_para}, where the MSGE on $G$ increases as $\sigma_g$ or $\beta$ increases, and our correction error starts to stand out (i.e. being larger than both prediction errors of PALMR and MGLM) when over $\beta=30\%$ of the training samples are grossly corrupted. We believe this is acceptable as in most practical situations, only small fraction of the training examples would be contaminated by gross errors.

\begin{figure}[!t]
\centering
\subfigure{\includegraphics[width = 0.48\columnwidth, height = 0.3\columnwidth]{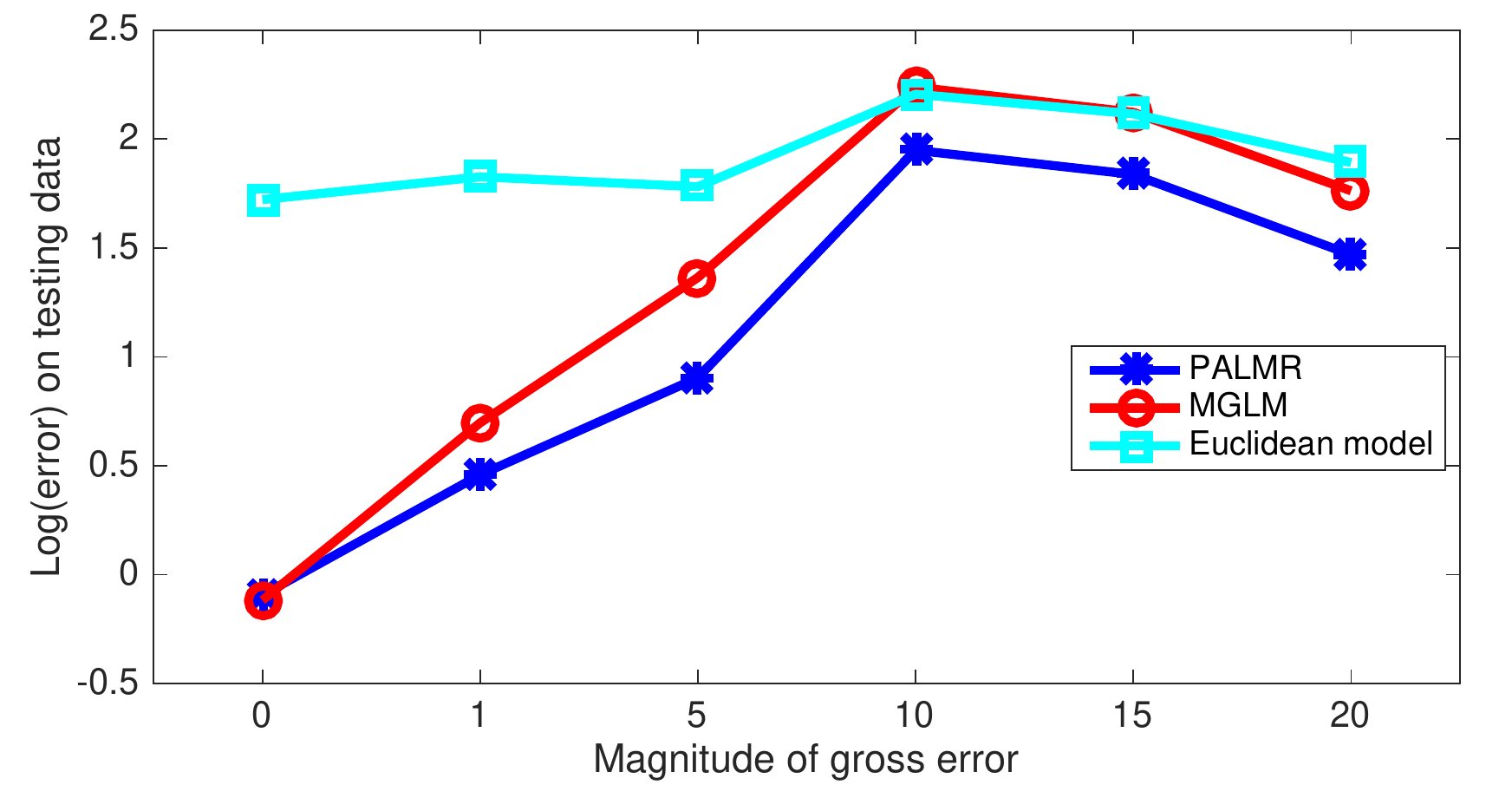}} \hspace{-1pt}
\subfigure{\includegraphics[width = 0.48\columnwidth, height = 0.3\columnwidth]{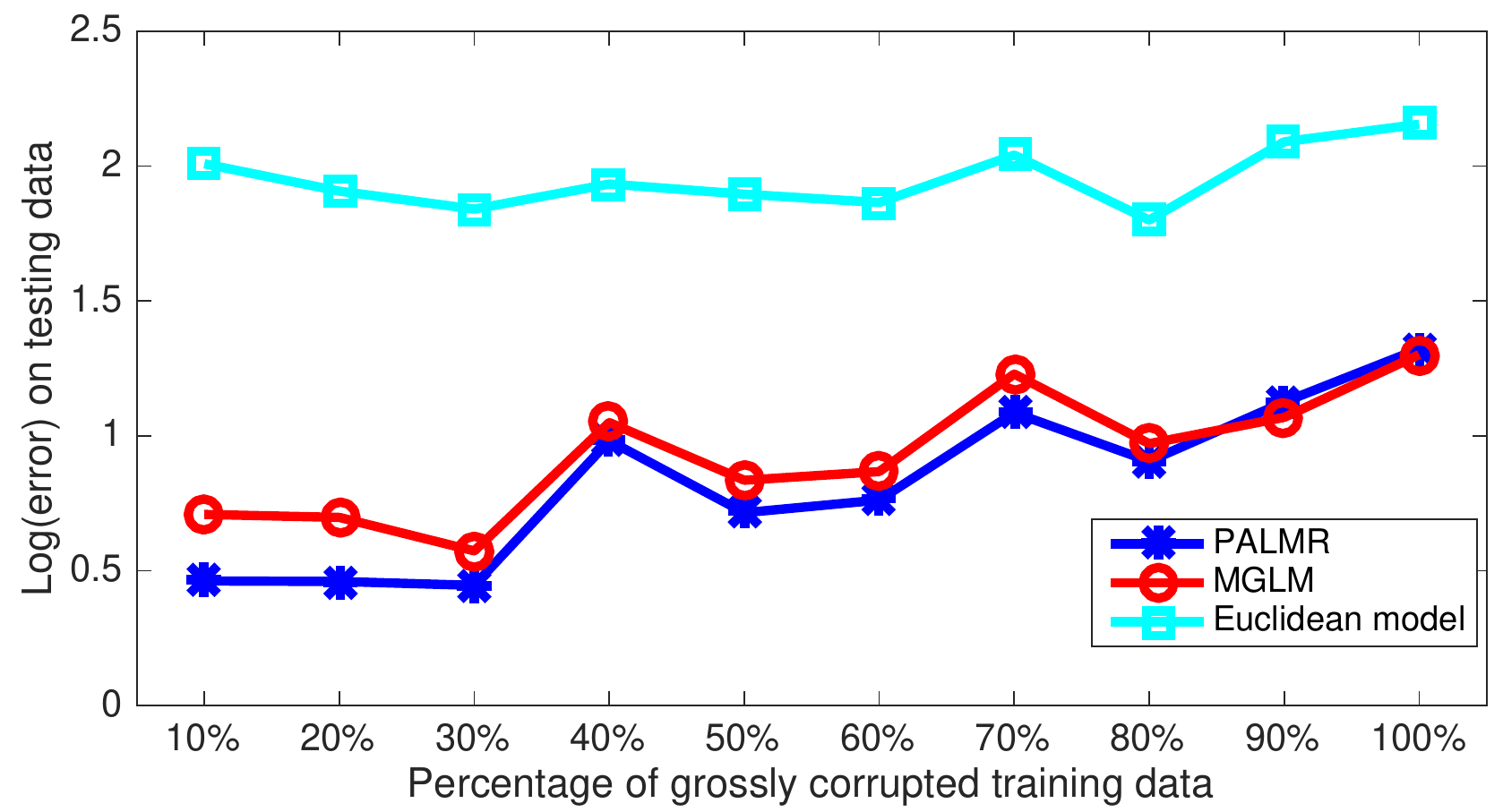}}
\vspace{-10pt}
\caption{Results of comparing PALMR and MGLM with Euclidean model \eqref{GR_Gross_RMTL} under different magnitude of gross error (left) and different ratio of gross error in the training data (right). For each plot, the $y$-axis denotes the log-scale of median error over 10 reptitions measured by Frobenious norm.}\label{fig:compare_euclid}
\vspace{-10pt}
\end{figure}

{
Finally, we compare the proposed method PALMR and MGLM with an Euclidean multivariate linear regression model with gross errors described in equation \eqref{GR_Gross_RMTL} of \textbf{Example}~\ref{ex1}. All experimental settings are the same as above except three aspects: (i) Since the Euclidean model can not deal with DTI tensors directly,  for each tensor $\bs{y}$, we vectorize its upper triangle part into a 6-dimensional vector. Therefore, $X\in\mathbb{R}^{50\times 2}$ and $Y\in\mathbb{R}^{50\times 6}$ in model \eqref{GR_Gross_RMTL}. (ii) Since predictions of the Euclidean model are not guaranteed to lie on the SPD manifold, the geodesic metrics are not applicable. As alternate, we adopt Frobenious norm distance $\|\bs{y} - \hat{\bs{y}}\|_F$ to measure the distance between prediction $\hat{\bs{y}}$ and ground-truth $\bs{y}$. (iii) We only investigate the effect of the magnitude of gross errors and the ratio of gross errors in the training data. Results are shown in  \figurename~\ref{fig:compare_euclid}, where the $y$-axis in each plot denotes the log-scale of median error over 10 reptitions measured by Frobenious norm. We observe that PALMR achieves the best performance and outporforms the Euclidean model by a large margin under various settings. MGLM also performs better than the Euclidean model, but when there are large gross errors in the training data, its advantage disappears, as can be seen in the left plot. These observations are within our expectation, since the Euclidean model does not respect the intrinsic structure of the DTI data.
}

\begin{figure*}[!t]
\centering
\includegraphics[scale = 0.7]{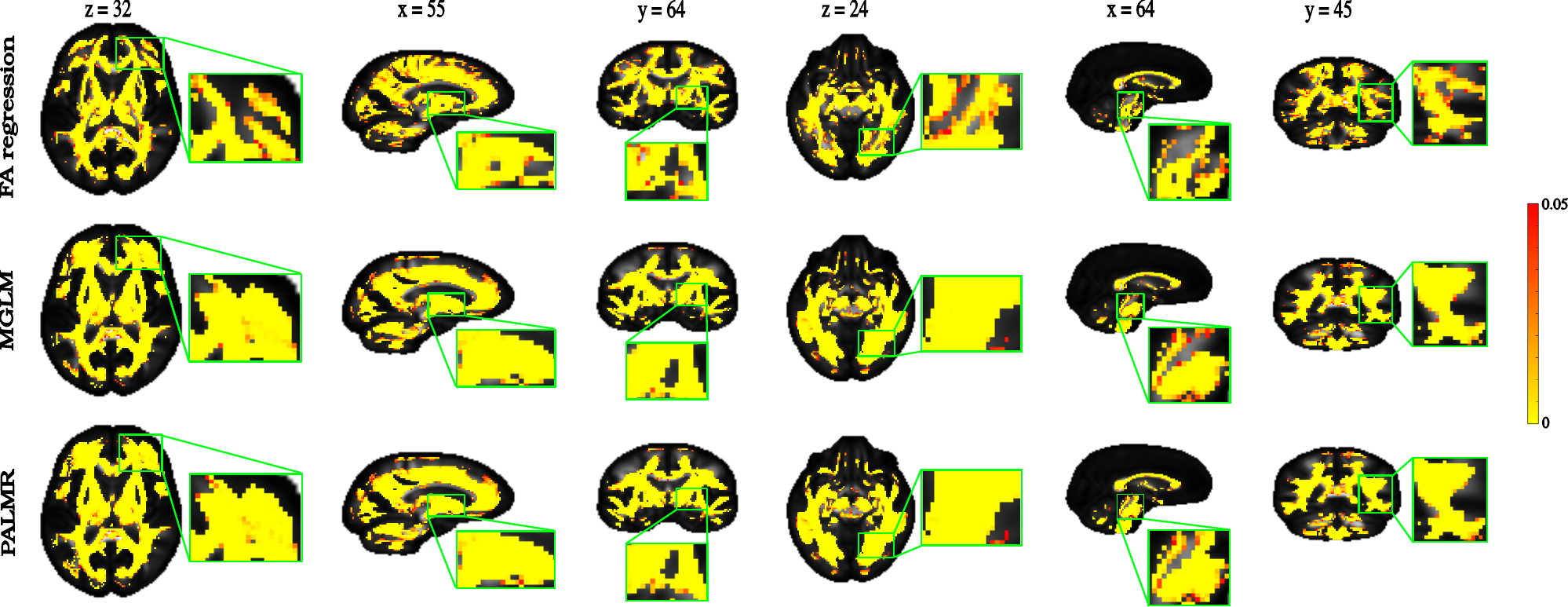}
\vspace{-10pt}
\caption{$p$-value maps obtained by three methods: FA regression (top), MGLM (middle) and PALMR (bottom). $p$-value is only illustrated for voxels with $p$-value $\leq$ 0.05. Best viewed in color.}
\label{Pvalue_map}
\vspace{-10pt}
\end{figure*}

\begin{figure}[!t]
\centering
\includegraphics[width=0.6\columnwidth, height=0.4\columnwidth]{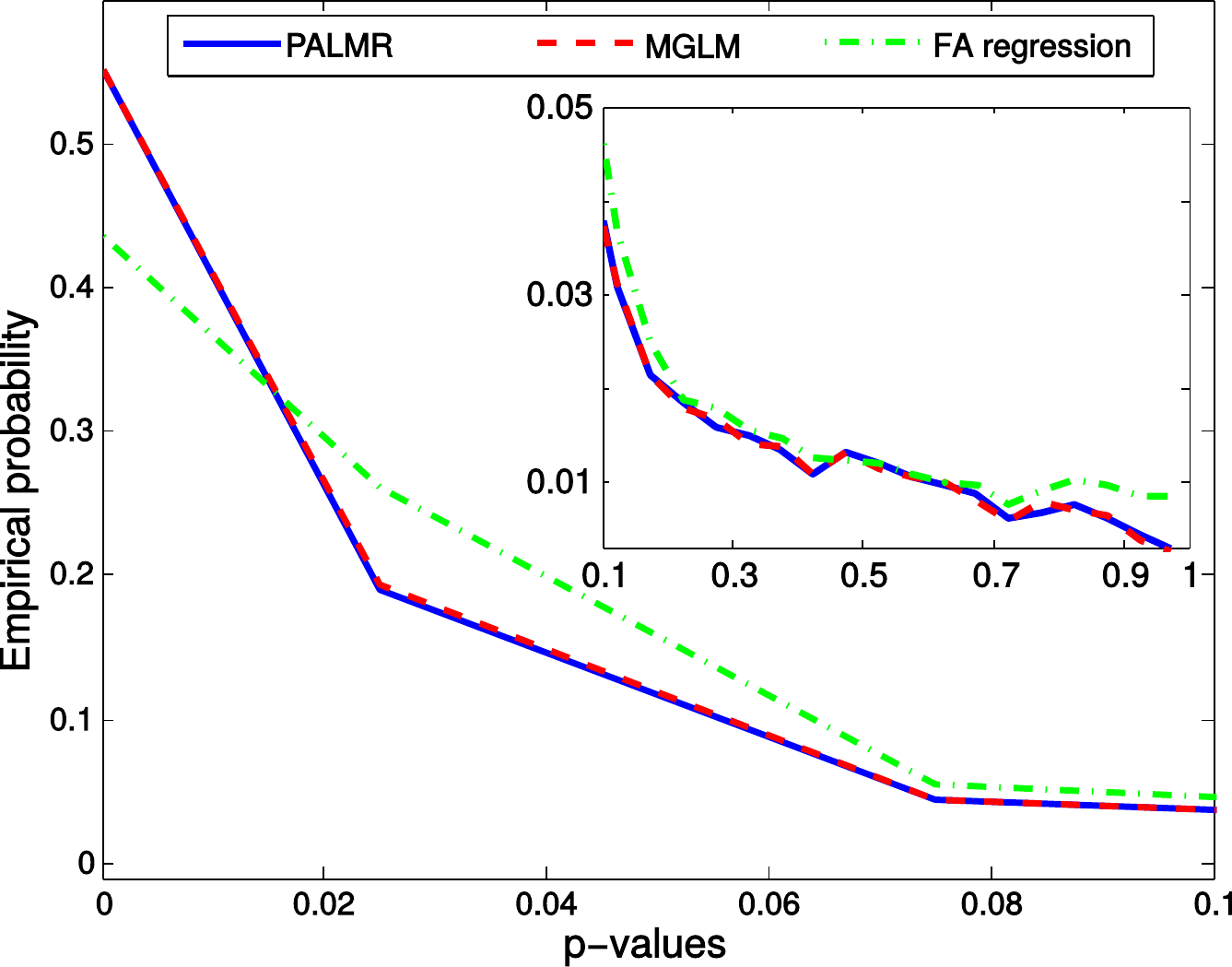}
\vspace{-10pt}
\caption{Distribution of $p$-values for white matter tensors in all six slices. The inlet plot shows distribution of $p$-values over range $[0.1, 1]$.}\label{Pvalue_distribution}
\vspace{-10pt}
\end{figure}

\subsection{Real DTI data}\label{subsec:real}

In this section, we apply PALMR to examine the effect of age and gender on human brain white matter. We experiment with the C-MIND database~\footnote{\url{https://cmind.research.cchmc.org}} released by Cincinnati Children's Hospital Medical Center (CCHMC) with the purpose of investigating brain development in children from infants and toddlers (0 $\sim$ 3 years) through adolescence (18 years).
We use the imaging data of participants who were scanned at CCHMC at year one and
whose age were between 8 and 18 (2947 to 6885 days), consisting of 27 female and 31 male. The DTI data of each subject are first manually inspected and corrected for subject movements and eddy current distortions using FSL's eddy tool \cite{JenkinsonBBWS:12}, then passed to FSL's brain extraction tool to delete non-brain tissue~\footnote{\url{http://fsl.fmrib.ox.ac.uk/fsl/fslwiki/}}. After the pre-processing, we use FSL's DTIFIT tool to reconstruct DTI tensors. Finally, all DTIs are registered to a population specific template constructed using DTI-TK~\footnote{\url{http://dti-tk.sourceforge.net/pmwiki/pmwiki.php}}. We investigate six exemplar slices that have been identified as typical slices by domain experts and have been also similarly used by many existing works such as~\cite{KimEtAl:cvpr14,DuGKQ14}.
And in particular, we are interested in the white matter region.
At each voxel within the white matter region, the following multivariate regression model
\begin{equation}\label{CMIND}
\bs{y} = \mbox{Exp}_{\bs{p}}(\bs{v}_{1} \times \mbox{age} + \bs{v}_{2} \times \mbox{gender})
\end{equation}
is adopted to describe the relation between the DTI data $\bs{y}$ and variables `age' and `gender'.

In DTI studies, another frequently used measure of a tensor is fractional anisotropy (FA)~\cite{Basser:95, Basser1996} defined as
\begin{align*}
FA = \sqrt{\frac{(\lambda_1 - \lambda_2)^2 + (\lambda_2 - \lambda_3)^2 + (\lambda_1 - \lambda_3)^2}{2(\lambda_1^2 + \lambda_2^2 + \lambda_2^2)}},
\end{align*}
where $\lambda_1$, $\lambda_2$ and $\lambda_3$ are eigenvalues of the tensor. FA is an important measurement of diffusion asymmetry within a voxel and reflects fiber density, axonal diameter, and myelination in white matter. In our experiments, we also compared three models: two geodesic regression models, MGLM and PALMR, and the FA regression model which uses FA value to replace tensor $\bs{y}$ in Eq.~\eqref{CMIND}.  {The relative FA error metric is employed to compare the results of geodesic regressions and FA regression, as follows: Since the responses of geodesic regression are tensors, the FA values of the tensors can be computed. The relative FA error metric is then evaluated on testing data, which is defined as the mean relative error between the FA values of the predicted tensors and the true tensors. Besides this relative FA error metric, the aforementioned mean squared geodesic error (MSGE) on testing data as in subsection~\ref{synthetic} is still engaged to compare the performance of MGLM and PALMR.}

\begin{figure}[!t]
\centering
\subfigure{\includegraphics[scale = 0.32]{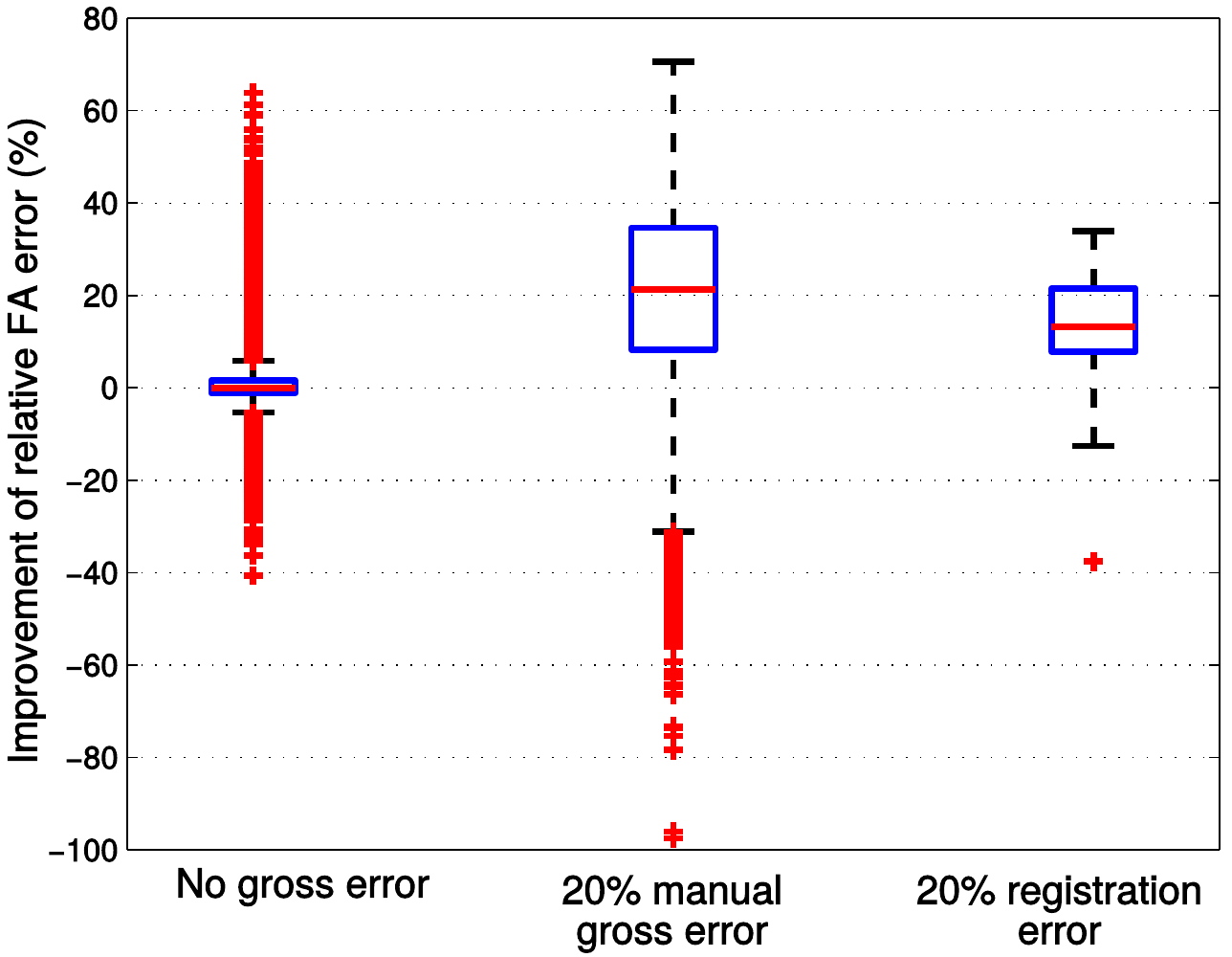}} \hspace{5pt}
\subfigure{\includegraphics[scale = 0.32]{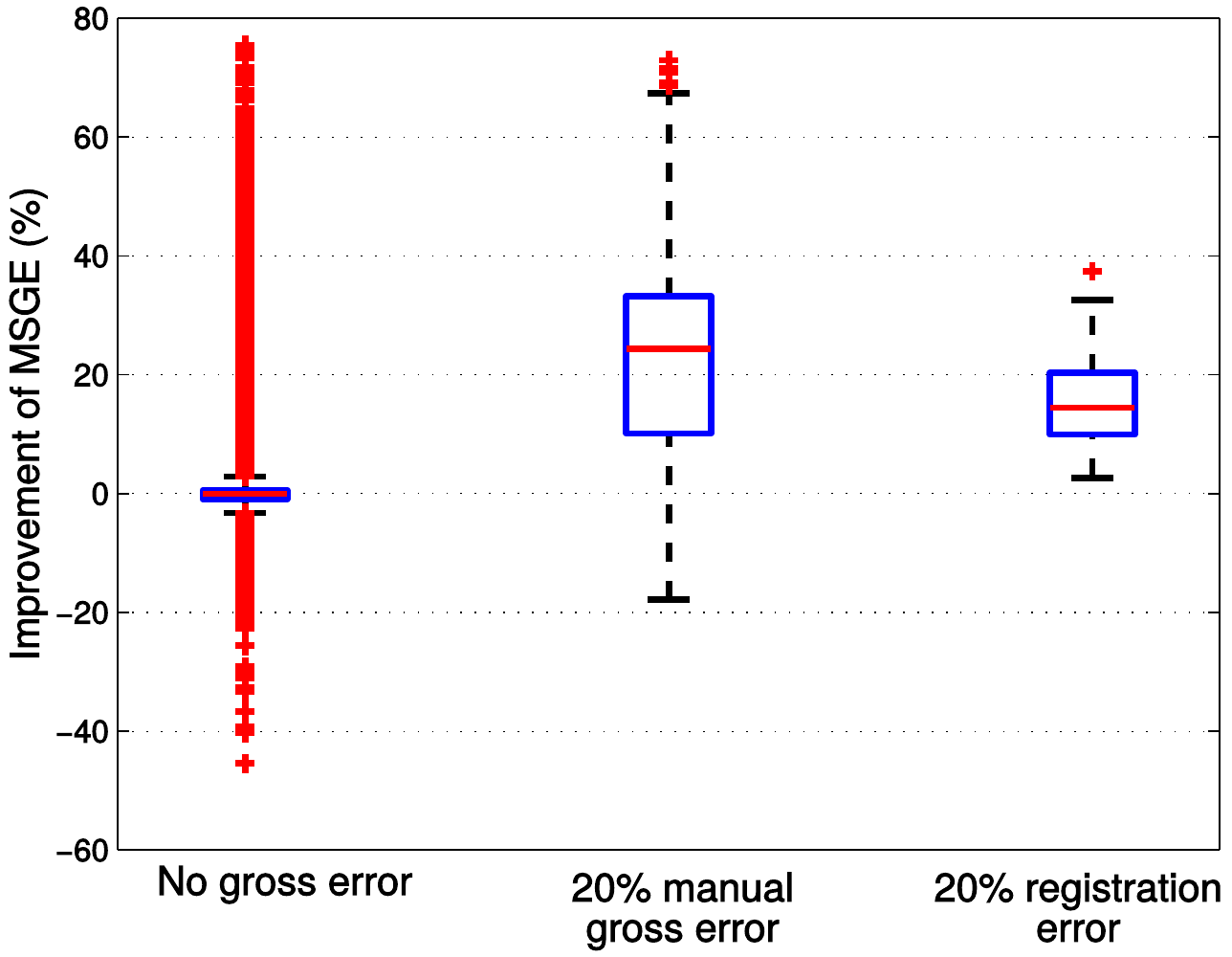}}
\vspace{-10pt}
\caption{Performance improvement obtained by PALMR measured with the relative FA error (left) and the MSGE (right). A positive value means that PALMR is better than the best competitor, and a negative value means that PALMR is worse. We first compute the performance improvement of PALMR on each voxel of all six slices to get a percentage value, then put all values under the same metric and experimental setting to plot a box plot.}\label{Perf-Impr}
\vspace{-10pt}
\end{figure}

\subsubsection{Model significance}
To examine the significance of the statistical model of Eq.~\eqref{CMIND} considered in our approach, the following hypothesis test is performed. The null hypothesis is $H_0: \bs{v}_{1}=0$, which means, under this hypothesis, age has no effect on the DTI data. We randomly permute the values of age \footnote{Empirical results investigating the effect of `gender' are provided in Section 5 of the supplementary file.} among all samples and fix the DTI data, then apply model of Eq.~\eqref{CMIND} to the permuted data and compute the mean squared geodesic error $MSGE_{perm}=\frac{1}{N}\sum_{i}\mbox{dist}(\bs{y}_i,\hat{\bs{y}}^p_i)^2$, where $\hat{\bs{y}}^p_i$ is the prediction of PALMR on the permuted data. Repeat the permutation $T=1000$ times, we get a sequence of errors  $\{MSGE_{perm}^i\}_{i=1}^T$ and calculate a $p$-value at each voxel using $p\mbox{-value} := \frac{|\{i \mid MSGE_{perm}^i < MSGE_{train}\}|}{T}$. \figurename~\ref{Pvalue_map} presents the maps of voxel-wise $p$-values for three models using six typical slices, and \figurename~\ref{Pvalue_distribution} displays the distribution of $p$-values for all six slices collectively.

As shown in \figurename~\ref{Pvalue_map} and \figurename~\ref{Pvalue_distribution}, geodesic regression models are able to capture more white matter regions with aging effects than FA regression model. In addition, voxels satisfying $p$-value $\leq 0.05$ are more spatially contiguous when geodesic regression models are used, as can be seen from the zoom-in plot for each slice in \figurename~\ref{Pvalue_map}. This may be attributed to the fact that geodesic regression models preserve more geometric information of tensor images than that of FA regression. We also observe that PALMR and MGLM obtain very similar results. This is to be expected, as both methods use model of Eq.~\eqref{CMIND} and adopt geodesic regression on manifolds. The main difference is that PALMR considers gross error while MGLM does not, and in this experiment, there is no gross error in the DTI data.

\begin{table}[!t]
\setlength{\tabcolsep}{2pt} 
\caption{Median values of prediction errors on all six slices of testing data. We use two metrics, relative FA error and MSGE, to measure the prediction error. The best results in each setting are highlighted in bold.}\label{median_error}
\vspace{-15pt}
\begin{center}
\begin{tabular}{|c|c|c|c|c||c|}
\hline
& Metrics & Methods & \begin{tabular}{@{}c@{}}No\\gross error\end{tabular} & \begin{tabular}{@{}c@{}}20\% manual\\gross error\end{tabular} & \begin{tabular}{@{}c@{}}20\%\\registration\\error\end{tabular} \\
\hline\hline
 \multirow{5}{*}{\rotatebox[origin=c]{90}{Slice $z=32$}} & \multirow{3}{*}{$\begin{array}{c}\mbox{Relative} \\ \mbox{FA~ error} \end{array}$} & FA regression & 0.9376 & 1.0414 & 0.9467 \\
 & & MGLM & 0.3223 & 0.4349 & 0.1654 \\
 & & PALMR & \textbf{0.3210} & \textbf{0.3409} & \textbf{0.1316} \\
\cline{2-6}
 & \multirow{2}{*}{MSGE} & MGLM & 0.1475 & 0.3530 & 0.1949\\
 & & PALMR & \textbf{0.1386} & \textbf{0.2196} & \textbf{0.1508} \\
\hline\hline
 \multirow{5}{*}{\rotatebox[origin=c]{90}{Slice $x=55$}} &
\multirow{3}{*}{$\begin{array}{c}\mbox{Relative} \\ \mbox{FA~ error} \end{array}$} & FA regression & 0.9238 & 1.0362 & 0.8688 \\
 & & MGLM & 0.3298 & 0.5089 & 0.2067 \\
 & & PALMR & \textbf{0.3279} & \textbf{0.3682} & \textbf{0.1882}\\
\cline{2-6}
 & \multirow{2}{*}{MSGE} & MGLM & 0.1606 & 0.3631 & 0.3513 \\
 & & PALMR & \textbf{0.1602} & \textbf{0.2562} & \textbf{0.2915} \\
\hline\hline
 \multirow{5}{*}{\rotatebox[origin=c]{90}{Slice $y=64$}} &
\multirow{3}{*}{$\begin{array}{c}\mbox{Relative} \\ \mbox{FA~ error} \end{array}$} & FA regression & 0.8822 & 1.0136 & 0.9528 \\
 & & MGLM & \textbf{0.3162} & 0.4564 & 0.1917 \\
 & & PALMR & 0.3166 & \textbf{0.3665} & \textbf{0.1562}\\
\cline{2-6}
 & \multirow{2}{*}{MSGE} & MGLM & 0.1687 & 0.3720 & 0.2449 \\
 & & PALMR & \textbf{0.1614} & \textbf{0.2843} & \textbf{0.1906} \\
\hline\hline
 \multirow{5}{*}{\rotatebox[origin=c]{90}{Slice $z=24$}} &
\multirow{3}{*}{$\begin{array}{c}\mbox{Relative} \\ \mbox{FA~ error} \end{array}$} & FA regression & 0.8478 & 1.0066 & 0.8144 \\
 & & MGLM & 0.3570 & 0.7342 & 0.2140 \\
 & & PALMR & \textbf{0.3564} & \textbf{0.5081} & \textbf{0.1581}\\
\cline{2-6}
 & \multirow{2}{*}{MSGE} & MGLM & 0.1227 & 0.3466 & 0.2954 \\
 & & PALMR & \textbf{0.1160} & \textbf{0.2530} & \textbf{0.2445} \\
\hline\hline
 \multirow{5}{*}{\rotatebox[origin=c]{90}{Slice $x=64$}} &
\multirow{3}{*}{$\begin{array}{c}\mbox{Relative} \\ \mbox{FA~ error} \end{array}$} & FA regression & 0.9723 & 1.0526 & 0.9067 \\
 & & MGLM & 0.2142 & 0.4053 & 0.5023 \\
 & & PALMR & \textbf{0.2114} & \textbf{0.3318} & \textbf{0.4318}\\
\cline{2-6}
 & \multirow{2}{*}{MSGE} & MGLM & 0.1646 & 0.3663 & 0.2436 \\
 & & PALMR & \textbf{0.1639} & \textbf{0.2779} & \textbf{0.2226} \\
\hline\hline
 \multirow{5}{*}{\rotatebox[origin=c]{90}{Slice $y=45$}} &
\multirow{3}{*}{$\begin{array}{c}\mbox{Relative} \\ \mbox{FA~ error} \end{array}$} & FA regression & 0.9715 & 1.0695 & 0.9379 \\
 & & MGLM & 0.3779 & 0.5976 & 0.1739 \\
 & & PALMR & \textbf{0.3767} & \textbf{0.5319} & \textbf{0.1664}\\
\cline{2-6}
 & \multirow{2}{*}{MSGE} & MGLM & 0.2162 & 0.4205 & 0.2928 \\
 & & PALMR & \textbf{0.2113} & \textbf{0.3780} & \textbf{0.2593} \\
\hline
\end{tabular}
\end{center}
\vspace{-10pt}
\end{table}

\begin{figure*}[!t]
\centering
\subfigure{\includegraphics[scale = 0.7]{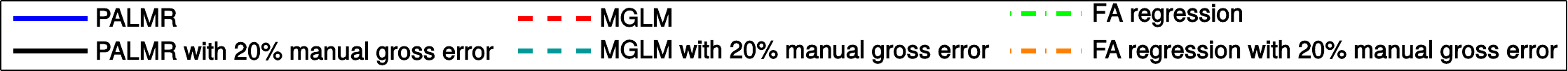}} \\
\vspace{-5pt}
\setcounter{subfigure}{0}
\subfigure[Slice $z=32$]{\includegraphics[width=0.55\columnwidth, height=0.35\columnwidth]{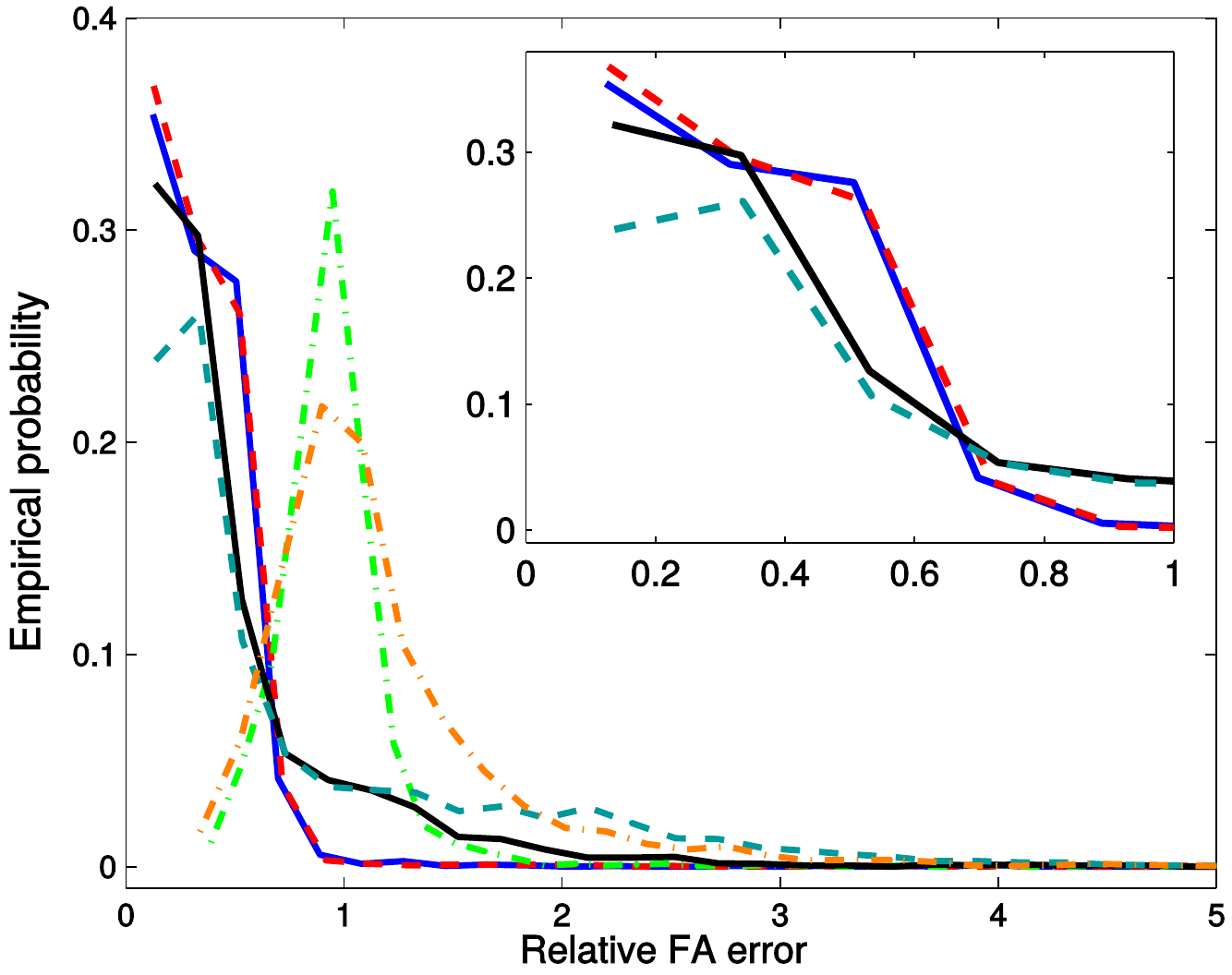}} \hspace{15pt}
\subfigure[Slice $x=55$]{\includegraphics[width=0.55\columnwidth, height=0.35\columnwidth]{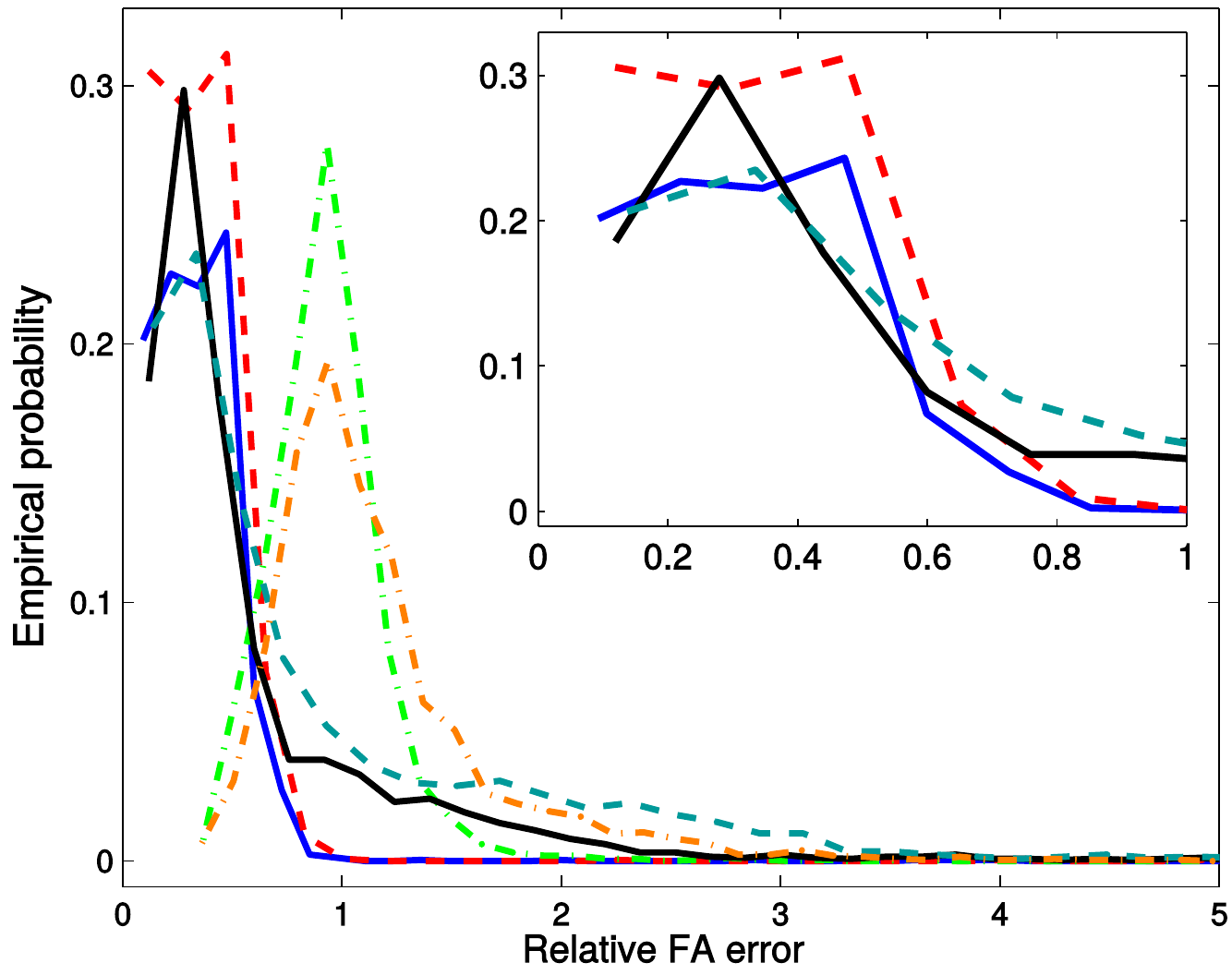}} \hspace{15pt}
\subfigure[Slice $y=64$]{\includegraphics[width=0.55\columnwidth, height=0.35\columnwidth]{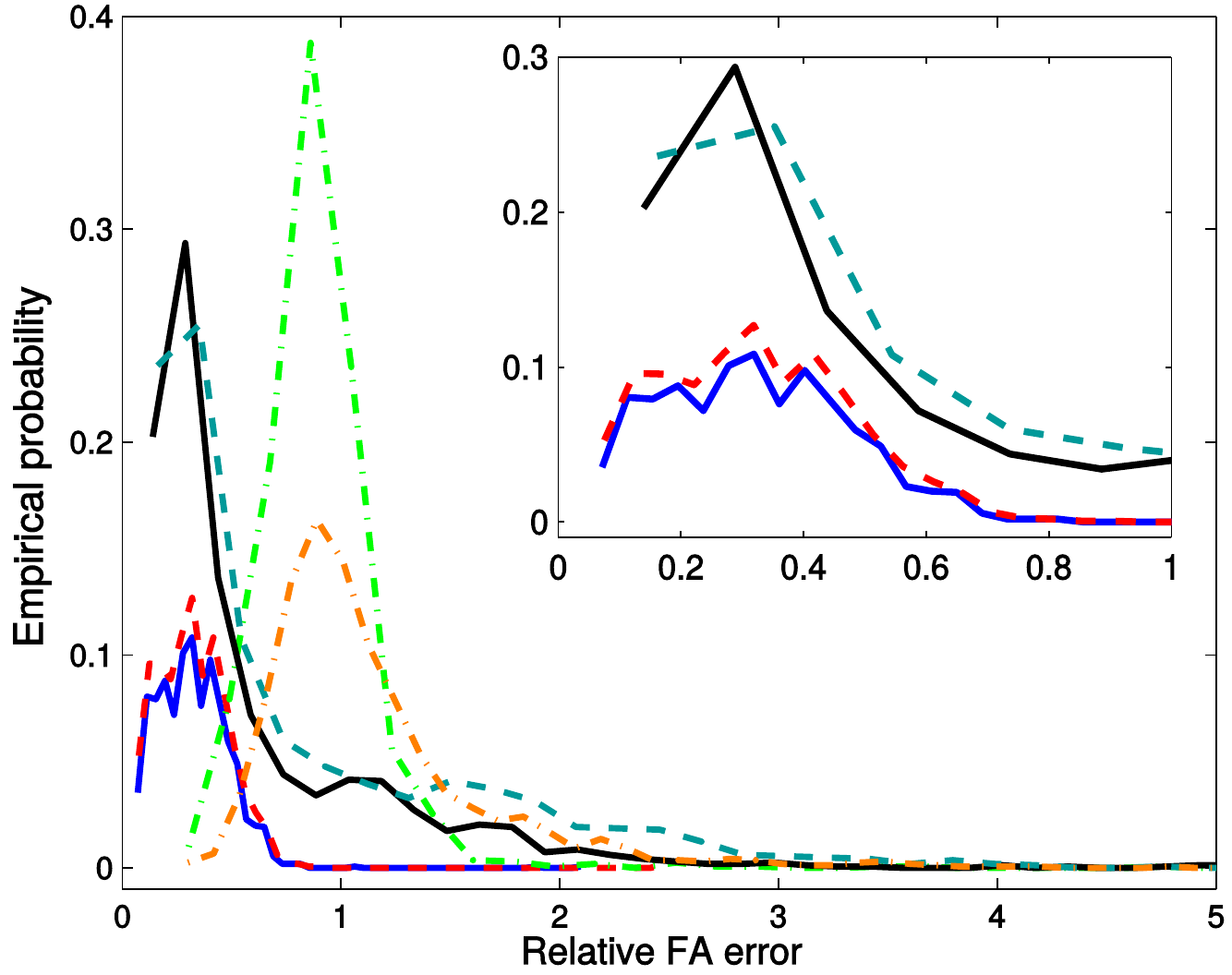}} \\
\vspace{-5pt}
\subfigure[Slice $z=24$]{\includegraphics[width=0.55\columnwidth, height=0.35\columnwidth]{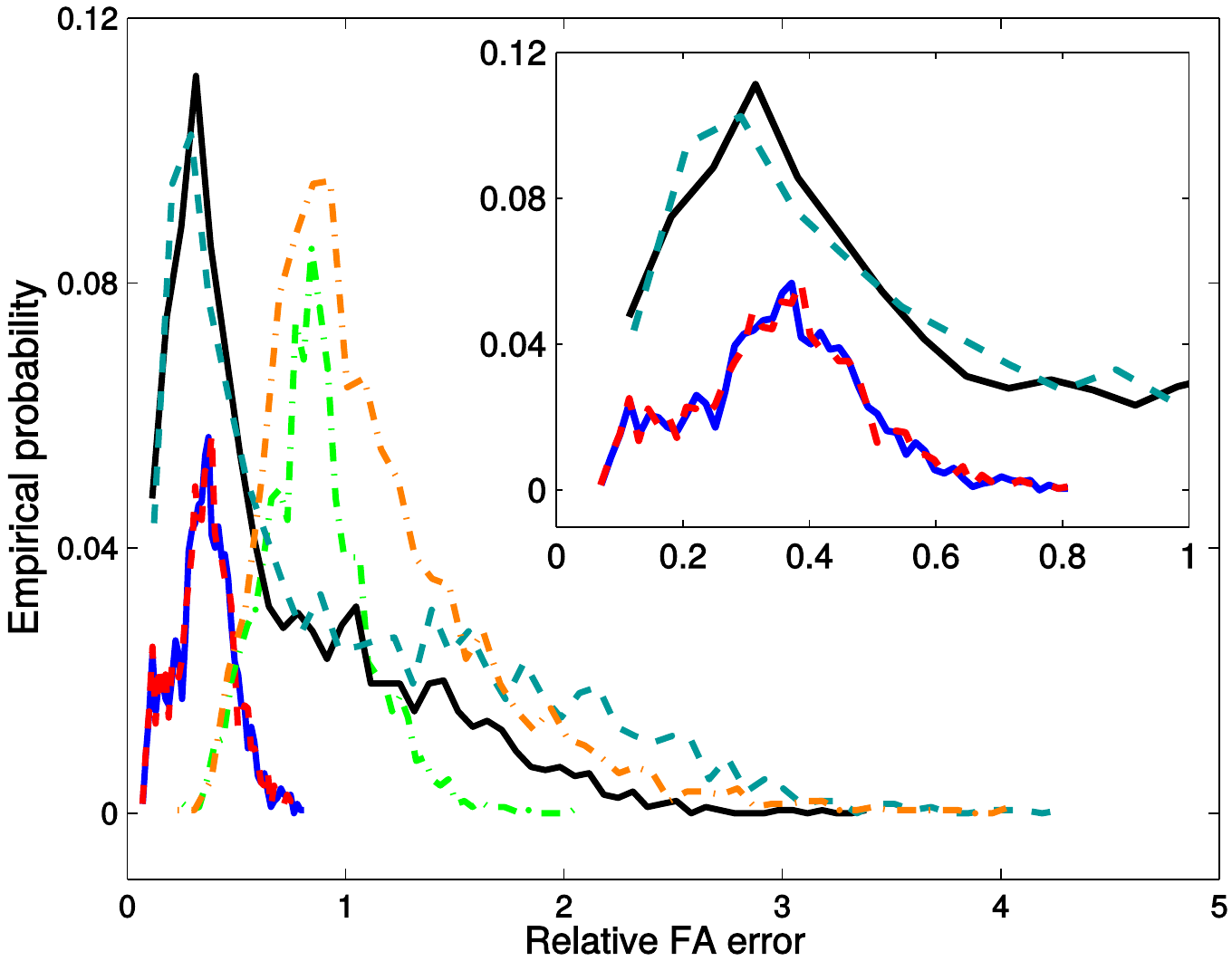}} \hspace{15pt}
\subfigure[Slice $x=64$]{\includegraphics[width=0.55\columnwidth, height=0.35\columnwidth]{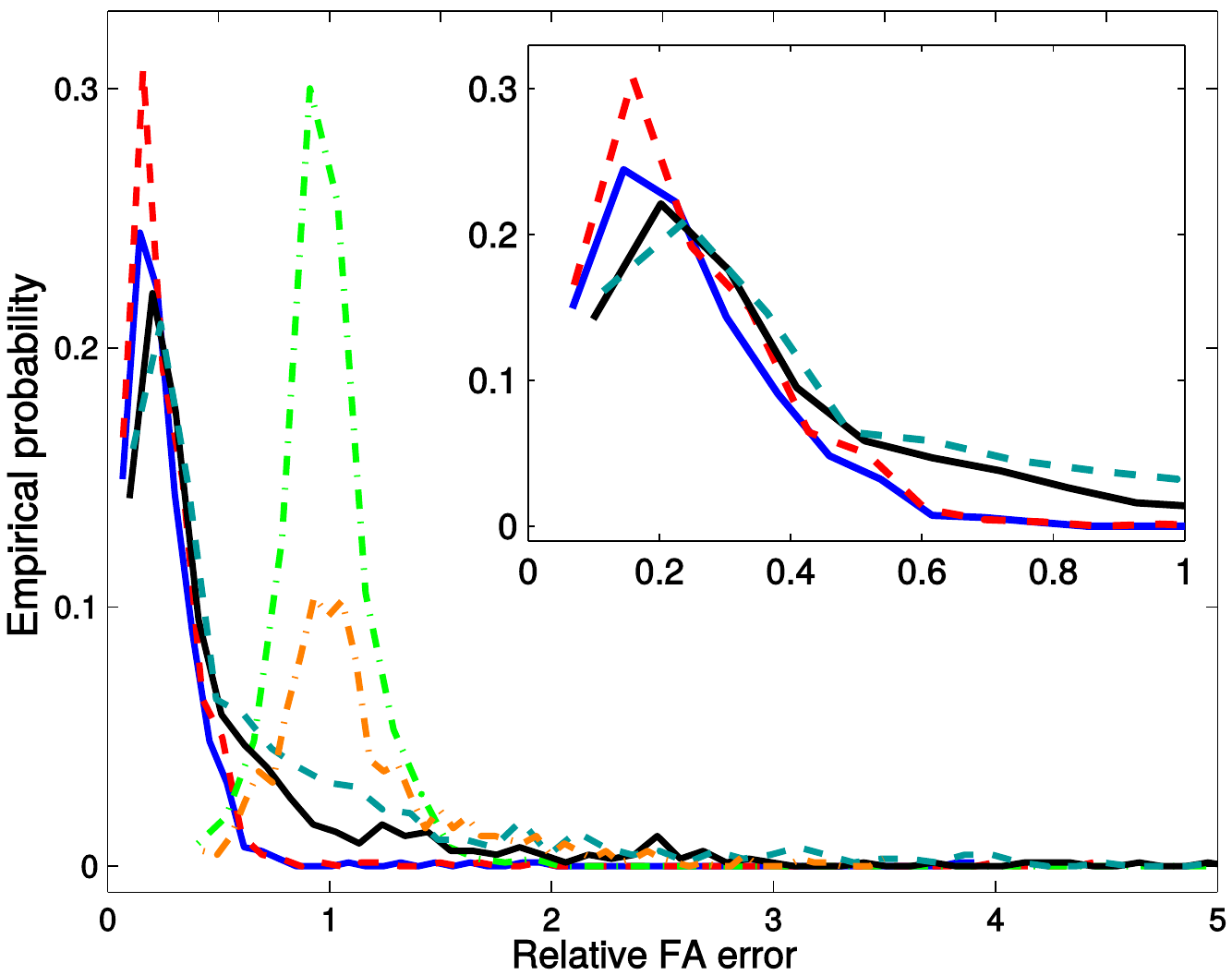}} \hspace{15pt}
\subfigure[Slice $y=45$]{\includegraphics[width=0.55\columnwidth, height=0.35\columnwidth]{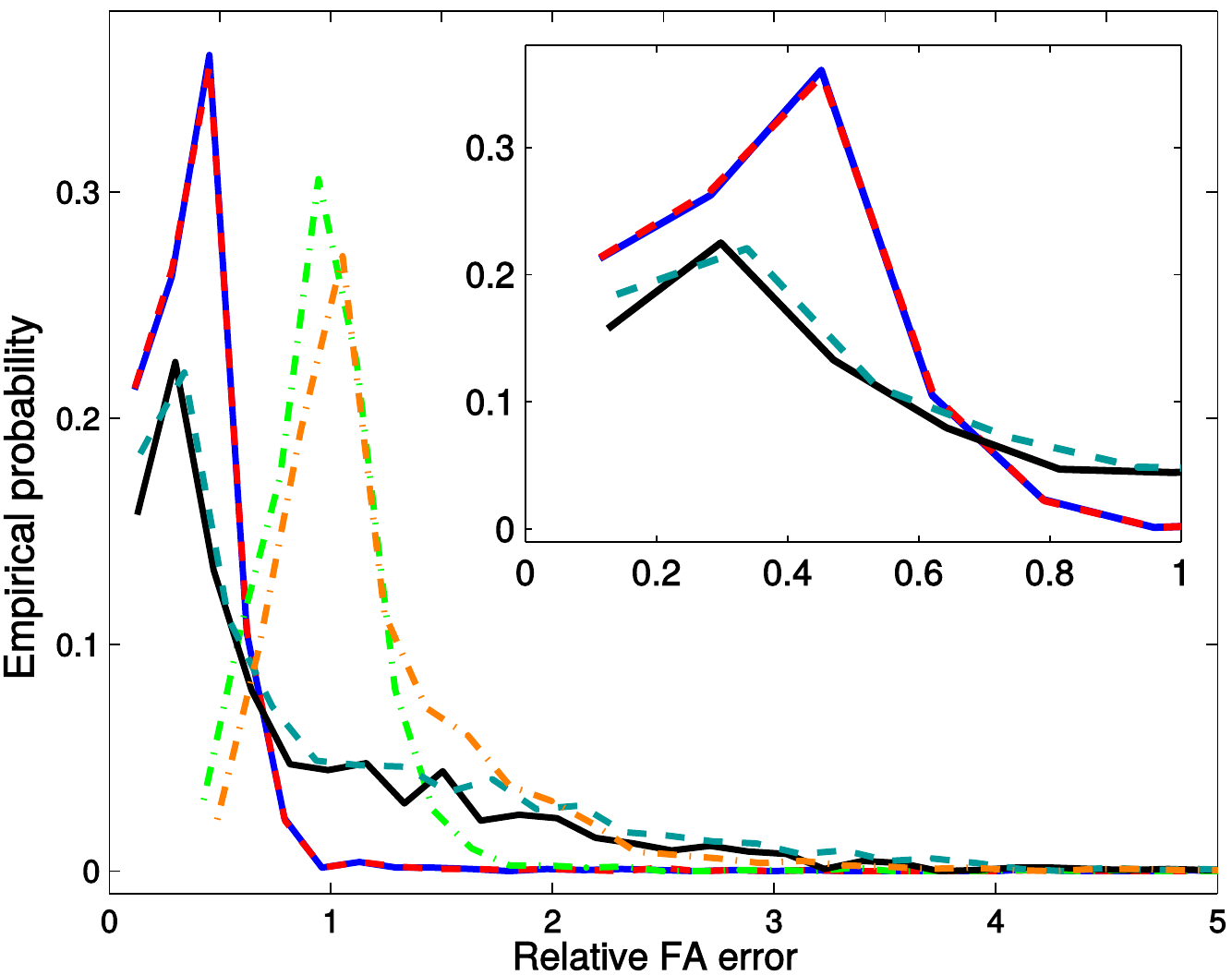}}
\vspace{-10pt}
\caption{Distribution of relative FA errors on testing data. The inset figures show zoom-in plots of the prediction errors by MGLM and PALMR over the error interval $[0, 1]$. Better viewed in color.}
\label{FA_Error_distribution}
\vspace{-5pt}
\end{figure*}

\subsubsection{Model predictability}
We proceed to investigate the predictability of PALMR when compared with existing methods such as FA regression and MGLM. For each of the six slices, we randomly partition our data into 40 training (20 female + 20 male) and 18 testing (7 female + 11 male) data, then train all three methods on each voxel within the white matter region. To test the ability of PALMR in handling gross errors, we consider three different experimental settings: (1) No gross error, where all training data are fully preprocessed as described at the beginning of subsection \ref{subsec:real}; (2) 20\% manual gross error, where for each voxel we randomly select $20\%$ of training instances and insert gross error with magnitude $\sigma_g = 5$; (3) 20\% registration error, where $20\%$ of the patients in the training data are randomly selected to undergo an incomplete registration processing. Compared with fully preprocessed data, DTI data with registration error are obtained by skipping the diffeomorphic registration step in DTI-TK. The purpose of experimenting on data with registration error is to imitate the realistic scenario that gross error can be caused by improper preprocessing of the data. We should remark that registration error is more challenging to handle than the manual gross error, since its magnitude varies dramatically for different voxels and patients. A heat map of registration error for each slice is provided in Fig.~1 of the supplementary file. In this case, instead of considering all voxels on each slice, we set a threshold value $\omega$ and consider those voxels whose minimum registration error is greater than $\omega$. For the first four slices, we set $\omega = 0.7$ and for the last two slices we set $\omega = 0.5$. The three comparison methods are examined on the three types of training data, and for each voxel the experiments are repeated 10 times.

\tablename~\ref{median_error} provides the median values of prediction errors measured with both relative FA error and MSGE on all voxels and over all six slices. As clearly indicated in \tablename~\ref{median_error}, geodesic regression models again outperform FA regression model, which is to be expected. Moreover, when there is no gross error in the training data, both MGLM and PALMR achieve similar results. This is consistent with the claim that MGLM is a special case of PALMR when there is no gross error. In addition, the `20\% manual gross error' column shows that when $20\%$ of the training data contain gross errors PALMR outperforms MGLM by a large margin. For the challenging case of 20\% registration error, the last column of \tablename~\ref{median_error} shows that PALMR is still much better than its competitors. In \figurename~\ref{Perf-Impr}, we use box plots to demonstrate the performance advantage of PALMR over its competitors. For each metric, the performance improvement is computed as \textit{(error of the best competitor - error of PALMR) / error of the best competitor * 100\%}. \figurename~\ref{Perf-Impr} displays the same results as in \tablename~\ref{median_error} from a different perspective and with more details. We first compute the performance improvement of PALMR on each voxel of all six slices to get a percentage value, then put all values under the same metric and experimental setting to plot a box plot. \figurename~\ref{Perf-Impr} shows that PALMR improves the median prediction error by at least 20\% and 15\% in the case of manual gross error and registration error, respectively.

\begin{figure*}[!t]
\centering
\subfigure{\includegraphics[scale = 0.7]{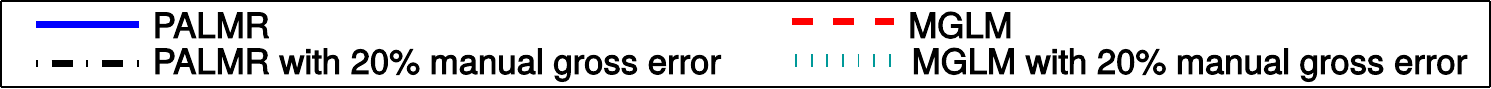}} \\
\vspace{-5pt}
\setcounter{subfigure}{0}
\subfigure[Slice $z=32$]{\includegraphics[width=0.55\columnwidth, height=0.35\columnwidth]{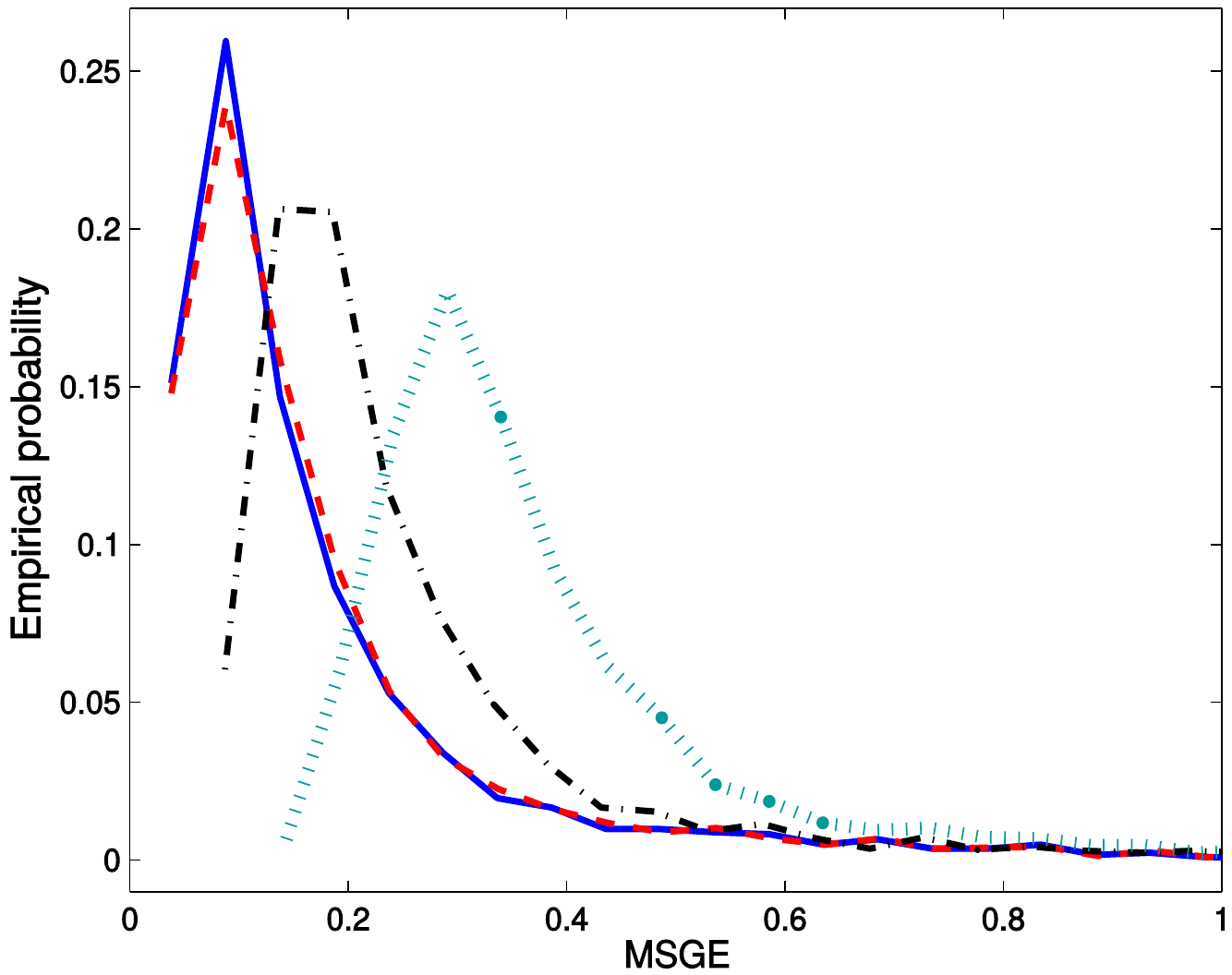}} \hspace{15pt}
\subfigure[Slice $x=55$]{\includegraphics[width=0.55\columnwidth, height=0.35\columnwidth]{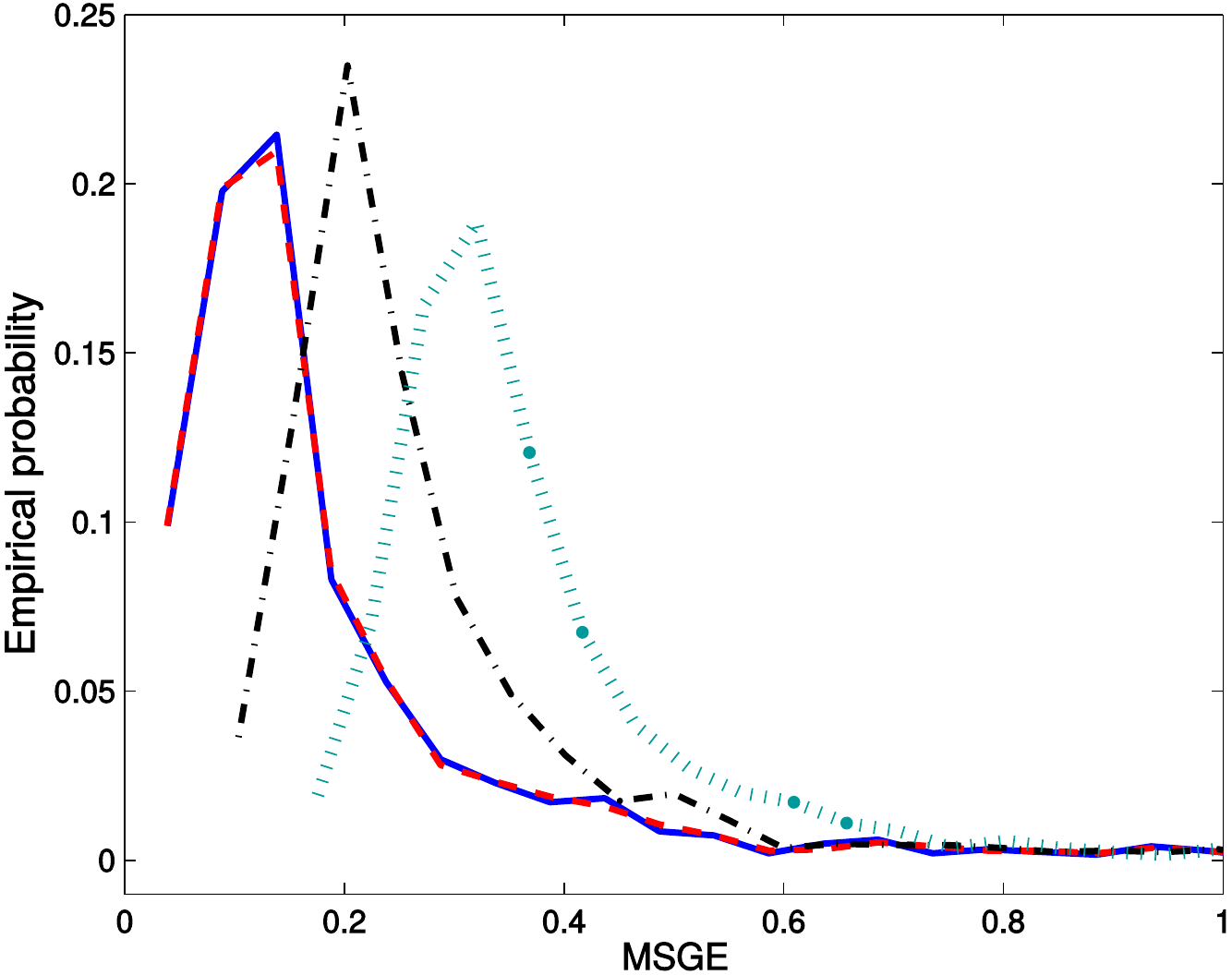}} \hspace{15pt}
\subfigure[Slice $y=64$]{\includegraphics[width=0.55\columnwidth, height=0.35\columnwidth]{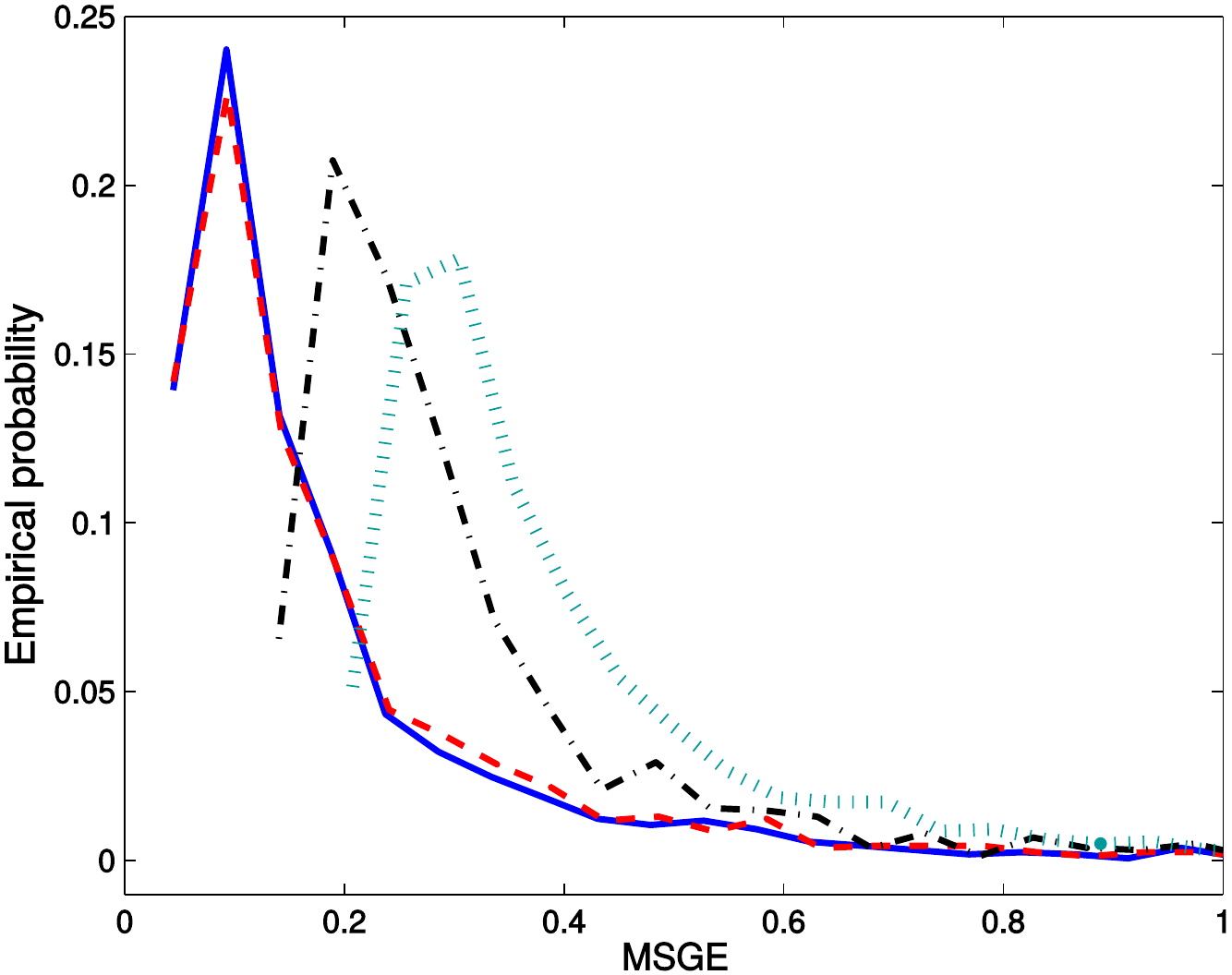}} \\
\vspace{-5pt}
\subfigure[Slice $z=24$]{\includegraphics[width=0.55\columnwidth, height=0.35\columnwidth]{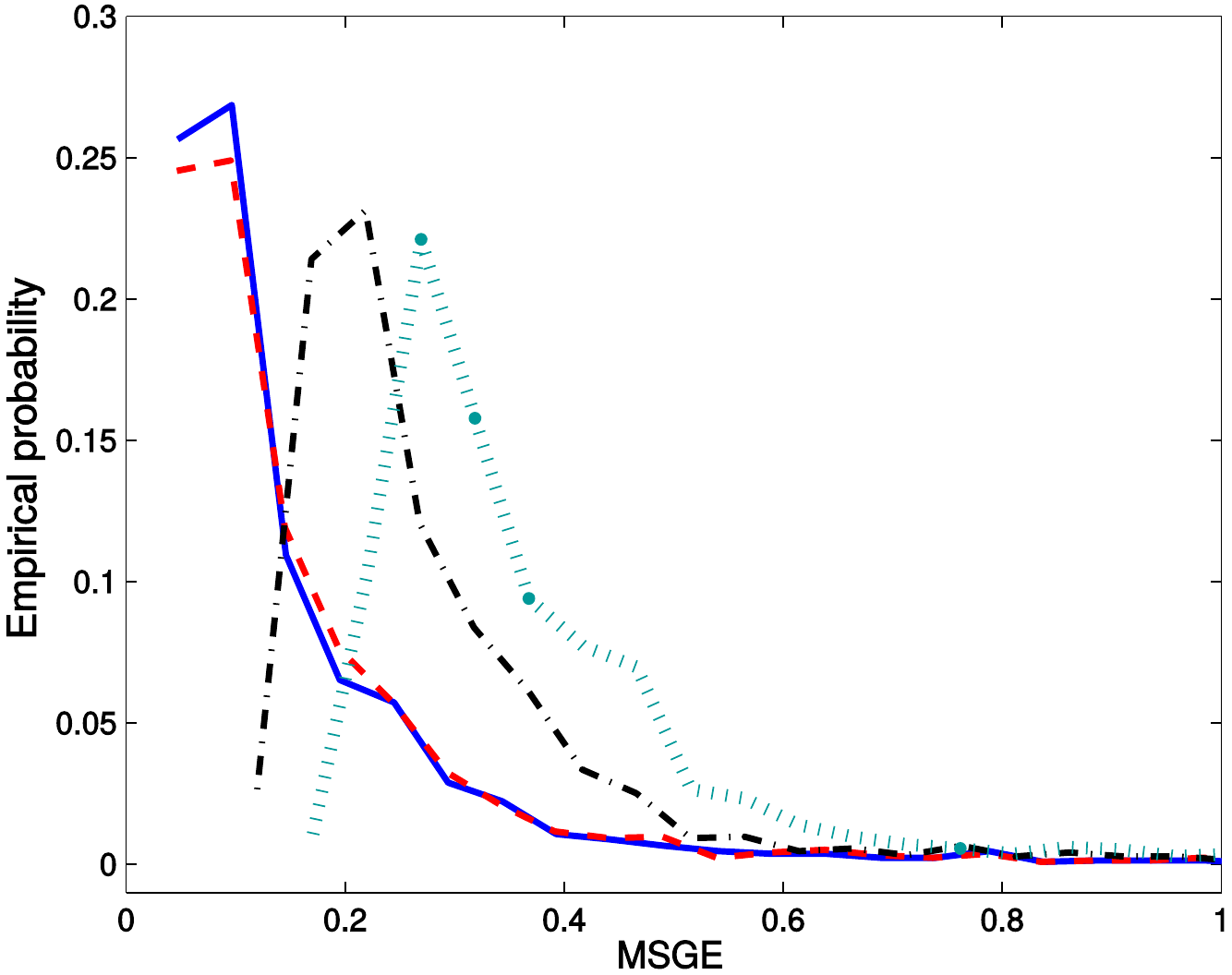}} \hspace{15pt}
\subfigure[Slice $x=64$]{\includegraphics[width=0.55\columnwidth, height=0.35\columnwidth]{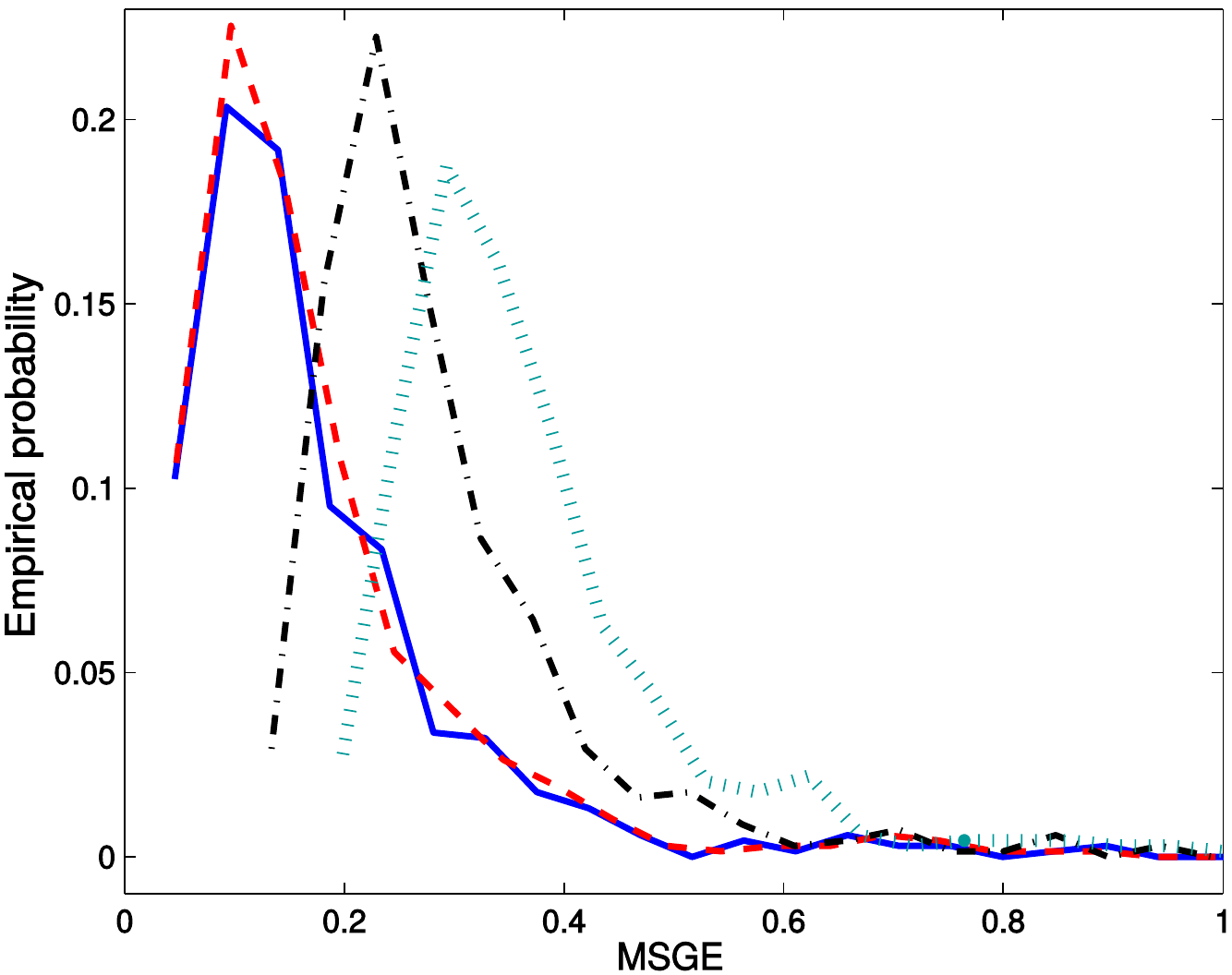}} \hspace{15pt}
\subfigure[Slice $y=45$]{\includegraphics[width=0.55\columnwidth, height=0.35\columnwidth]{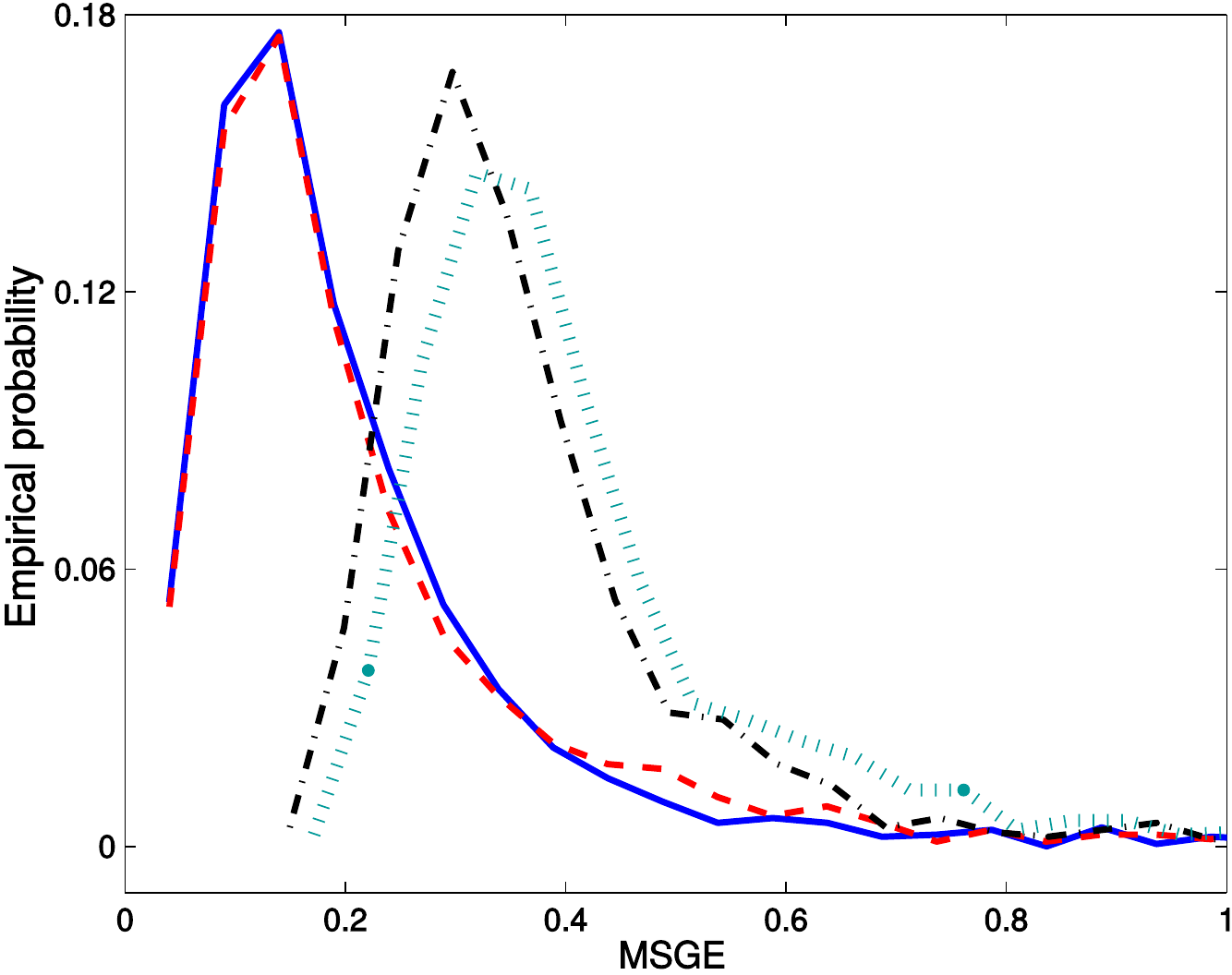}}
\vspace{-10pt}
\caption{Distribution of mean squared geodesic errors on testing data for each of the six slices. Better viewed in color.}
\label{MSGE_distribution}
\vspace{-10pt}
\end{figure*}

The distribution of prediction errors measured by the relative FA error and the MSGE on each slice is shown in \figurename~\ref{FA_Error_distribution} and \figurename~\ref{MSGE_distribution}, respectively. In each plot, the method with corresponding distribution on the left is better than the one with corresponding distribution on the right. From both \figurename~\ref{FA_Error_distribution} and \figurename~\ref{MSGE_distribution}, we get similar observation as in \tablename~\ref{median_error}. Moreover, \figurename~\ref{MSGE_distribution} shows that PALMR is more robust to gross errors than its competitors. In Fig.~2 of the supplementary file, we also show the comparison of prediction errors of MGLM and PALMR on each voxel of all slices. We observe that on most of the voxels PALMR is better than MGLM when gross errors are present. More experimental results on real DTI data are available in Section 5 of the supplementary file.


\section{Conclusion and Future Work}\label{sec:conclusion}
This paper focuses on the interesting problem of multivariate regression on manifolds with gross error contamination, where mathematical formulation nevertheless resides in a challenging landscape concerning a nonconvex and nonsmooth optimization on manifolds. A new algorithm, PALMR, is proposed to address this problem and its convergence property is analyzed. Through empirical studies, PALMR is shown to be capable of dealing with the presence of gross error and produces reliable results.
{For future work, there are several directions to explore. In terms of theoretical study, it remains to investigate the recoverbility of the proposed model, that is, to study conditions under which our model can correctly locate gross errors and recover their magnitude. It is also of interest to analyze the asymptotic behaviour of the resulting estimators. In terms of applications, in addition to age and gender, one may also consider the influence of handedness (i.e. left- or right-handed) on DTI responses.} We also plan to apply our framework to different applications including shape analysis and robotics, where the manifolds of interest could be $SO(3)$ and $SE(3)$.

{
\bibliography{main}

\begin{thebibliography}{10}
\providecommand{\url}[1]{#1}
\csname url@samestyle\endcsname
\providecommand{\newblock}{\relax}
\providecommand{\bibinfo}[2]{#2}
\providecommand{\BIBentrySTDinterwordspacing}{\spaceskip=0pt\relax}
\providecommand{\BIBentryALTinterwordstretchfactor}{4}
\providecommand{\BIBentryALTinterwordspacing}{\spaceskip=\fontdimen2\font plus
\BIBentryALTinterwordstretchfactor\fontdimen3\font minus
  \fontdimen4\font\relax}
\providecommand{\BIBforeignlanguage}[2]{{%
\expandafter\ifx\csname l@#1\endcsname\relax
\typeout{** WARNING: IEEEtran.bst: No hyphenation pattern has been}%
\typeout{** loaded for the language `#1'. Using the pattern for}%
\typeout{** the default language instead.}%
\else
\language=\csname l@#1\endcsname
\fi
#2}}
\providecommand{\BIBdecl}{\relax}
\BIBdecl

\bibitem{DavisFBJ:ijcv10}
B.~C. Davis, T.~Fletcher, E.~Bullitt, and S.~C. Joshi, ``Population shape
  regression from random design data,'' \emph{International Journal of Computer
  Vision}, vol.~90, no.~2, pp. 255--66, 2010.

\bibitem{KimEtAl:cvpr14}
H.~Kim, N.~Adluru, M.~Collins, M.~Chung, B.~Bendlin, S.~Johnson, R.~Davidson,
  and V.~Singh, ``Multivariate general linear models ({MGLM}) on {R}iemannian
  manifolds with applications to statistical analysis of diffusion weighted
  images,'' in \emph{CVPR}, 2014.

\bibitem{Cornea:RSSB17}
E.~Cornea, H.~Zhu, P.~Kim, J.~Ibrahim, and the Alzheimer's Disease
  Neuroimaging~Initiative, ``Regression models on {Riemannian} symmetric
  spaces,'' \emph{Journal of the Royal Statistical Society: Series B
  (Statistical Methodology)}, vol.~79, no.~2, pp. 463--482, 2017.

\bibitem{Muralidharan:CVPR12}
P.~Muralidharan and P.~T. Fletcher, ``Sasaki metrics for analysis of
  longitudinal data on manifolds,'' in \emph{CVPR}, 2012.

\bibitem{Hsu:2008}
J.~Hsu, A.~Leemans, C.~Bai, C.~Lee, Y.~Tsai, H.~Chiu, and W.~Chen, ``Gender
  differences and age-related white matter changes of the human brain: A
  diffusion tensor imaging study,'' \emph{NeuroImage}, vol.~39, no.~2, pp.
  566--577, 2008.

\bibitem{WuCWLBMP:miccai08}
M.~Wu, L.-C. Chang, L.~Walker, H.~Lemaitre, A.~Barnett, S.~Marenco, and
  C.~Pierpaoli, ``Comparison of {EPI} distortion correction methods in
  diffusion tensor {MRI} using a novel framework,'' in \emph{MICCAI}, 2008.

\bibitem{Zalesky:11}
A.~Zalesky, ``Moderating registration misalignment in voxelwise comparisons of
  {DTI} data: a performance evaluation of skeleton projection,'' \emph{Magnetic
  Resonance Imaging}, vol.~29, no.~1, pp. 111--125, 2011.

\bibitem{Bastin:98}
M.~Bastin, P.~Armitage, and I.~Marshall, ``A theoretical study of the effect of
  experimental noise on the measurement of anisotropy in diffusion imaging,''
  \emph{Magnetic Resonance Imaging}, vol.~16, no.~7, pp. 773--785, 1998.

\bibitem{BasuFW:miccai06}
S.~Basu, T.~Fletcher, and R.~Whitaker, ``Rician noise removal in diffusion
  tensor {MRI},'' in \emph{MICCAI}, 2006.

\bibitem{Wright:TIT10}
J.~Wright and Y.~Ma, ``Dense error correction via $\ell^1$-minimization,''
  \emph{IEEE Trans. Inf. Theor.}, vol.~56, no.~7, pp. 3540--60, 2010.

\bibitem{Candes:RPCA11}
E.~J. Cand\`{e}s, X.~Li, Y.~Ma, and J.~Wright, ``Robust principal component
  analysis?'' \emph{J. ACM}, vol.~58, no.~3, pp. 11:1--11:37, 2011.

\bibitem{ChenJSC:TIT13}
Y.~Chen, A.~Jalali, S.~Sanghavi, and C.~Caramanis, ``Low-rank matrix recovery
  from errors and erasures,'' \emph{{IEEE} Trans. Inf. Theor.}, vol.~59, no.~7,
  pp. 4324--37, 2013.

\bibitem{NguTra:tit13}
N.~Nguyen and T.~Tran, ``robust lasso with missing and grossly corrupted
  observations,'' \emph{IEEE Trans. Info. Theory}, vol.~59, no.~4, pp.
  2036--58, 2013.

\bibitem{XuLen:aistat12}
H.~Xu and C.~Leng, ``robust multi-task regression with grossly corrupted
  observations,'' in \emph{AISTAT}, 2012.

\bibitem{Bhatia15}
K.~Bhatia, P.~Jain, and P.~Kar, ``Robust regression via hard thresholding,'' in
  \emph{NIPS}, 2015.

\bibitem{BolSabTeb:mp14}
J.~Bolte, S.~Sabach, and M.~Teboulle, ``Proximal alternating linearized
  minimization for nonconvex and nonsmooth problems,'' \emph{Math. Program.},
  vol. 146, no. 1-2, pp. 459--94, 2014.

\bibitem{Basser02}
P.~Basser and D.~Jones, ``Diffusion-tensor {MRI}: theory, experimental design
  and data analysis -- a technical review,'' \emph{NMR Biomed.}, vol.~15, pp.
  456--467, 2002.

\bibitem{Srivastava:PAMI2005}
A.~Srivastava, S.~Joshi, W.~Mio, and X.~Liu, ``Statistical shape analysis:
  Clustering, learning, and testing,'' \emph{IEEE Trans. Pattern Anal. Mach.
  Intell.}, vol.~27, no.~4, pp. 590--602, 2005.

\bibitem{FletcherLPJ04}
T.~P. Fletcher, C.~Lu, S.~M. Pizer, and S.~C. Joshi, ``Principal geodesic
  analysis for the study of nonlinear statistics of shape,'' \emph{IEEE Trans.
  Med. Imaging}, vol.~23, pp. 995--1005, 2004.

\bibitem{HinkleFJ:jmiv14}
J.~Hinkle, P.~T. Fletcher, and S.~C. Joshi, ``Intrinsic polynomials for
  regression on {R}iemannian manifolds,'' \emph{Journal of Mathematical Imaging
  and Vision}, vol.~50, no. 1-2, pp. 32--52, 2014.

\bibitem{SaxDriNg:icra09}
A.~Saxena, J.~Driemeyer, and A.~Ng, ``Learning 3-{D} object orientation from
  images,'' in \emph{ICRA}, 2009.

\bibitem{WanPulPop:atg07}
R.~Wang, K.~Pulli, and J.~Popovi\'{c}, ``Real-time enveloping with rotational
  regression,'' \emph{ACM Trans. Graph.}, vol.~26, no.~3, 2007.

\bibitem{PorikliTM06}
F.~Porikli, O.~Tuzel, and P.~Meer, ``Covariance tracking using model update
  based on lie algebra,'' in \emph{CVPR}, 2006.

\bibitem{Pennec:ijcv06}
X.~Pennec, P.~Fillard, and N.~Ayache, ``A {R}iemannian framework for tensor
  computing,'' \emph{Int. J. Comput. Vision}, vol.~66, no.~1, pp. 41--66, 2006.

\bibitem{Carreira:PAMI2014}
J.~Carreira, R.~Caseiro, J.~Batista, and C.~Sminchisescu, ``Free-form region
  description with second-order pooling,'' \emph{IEEE Trans. Pattern Anal.
  Mach. Intell.}, vol.~37, no.~6, pp. 1177--1189, 2015.

\bibitem{Cherian:TNNLS17}
A.~Cherian and S.~Sra, ``Riemannian dictionary learning and sparse coding for
  positive definite matrices,'' \emph{IEEE Transactions on Neural Networks and
  Learning Systems}, no.~99, pp. 1--13, 2017.

\bibitem{DuGKQ14}
J.~Du, A.~Goh, S.~Kushnarev, and A.~Qiu, ``Geodesic regression on orientation
  distribution functions with its application to an aging study,''
  \emph{NeuroImage}, vol.~87, pp. 416--26, 2014.

\bibitem{Fle:ijcv13}
T.~Fletcher, ``Geodesic regression and the theory of least squares on
  {R}iemannian manifolds,'' \emph{Int. J. Comput. Vision}, vol. 105, no.~2, pp.
  171--85, 2013.

\bibitem{FletcherJ07}
P.~Fletcher and S.~C. Joshi, ``Riemannian geometry for the statistical analysis
  of diffusion tensor data,'' \emph{Signal Processing}, vol.~87, no.~2, pp.
  250--262, 2007.

\bibitem{Pennec:2006}
X.~Pennec, ``Intrinsic statistics on {R}iemannian manifolds: Basic tools for
  geometric measurements,'' \emph{J. Math. Imaging Vis.}, vol.~25, no.~1, pp.
  127--154, 2006.

\bibitem{Belkin:jmlr06}
M.~Belkin, P.~Niyogi, and V.~Sindhwani, ``Manifold regularization: A geometric
  framework for learning from labeled and unlabeled examples,'' \emph{J. Mach.
  Learn. Res.}, vol.~7, pp. 2399--434, 2006.

\bibitem{Banerjee_2016_CVPR}
M.~Banerjee, R.~Chakraborty, E.~Ofori, M.~Okun, D.~Viallancourt, and B.~Vemuri,
  ``A nonlinear regression technique for manifold valued data with applications
  to medical image analysis,'' in \emph{CVPR}, 2016.

\bibitem{HongKSVN:16}
Y.~Hong, R.~Kwitt, N.~Singh, N.~Vasconcelos, and M.~Niethammer, ``Parametric
  regression on the {Grassmannian},'' \emph{{IEEE} Trans. Pattern Anal. Mach.
  Intell.}, vol.~38, no.~11, pp. 2284--2297, 2016.

\bibitem{SteinkeM:NIPS2008}
F.~S. M. and Hein, ``Non-parametric regression between manifolds,'' in
  \emph{NIPS}, 2009.

\bibitem{Hein:NIPS2009}
M.~Hein, ``Robust nonparametric regression with metric-space valued output,''
  in \emph{NIPS}, 2009.

\bibitem{LiCS2012}
X.~Li, ``Compressed sensing and matrix completion with constant proportion of
  corruptions,'' arXiv:1104.1041v2, Tech. Rep., 2012.

\bibitem{Harandi:ECCV2012}
M.~Harandi, C.~Sanderson, R.~Hartley, and B.~Lovell, ``Sparse coding and
  dictionary learning for symmetric positive definite matrices: A kernel
  approach,'' in \emph{ECCV}, 2012, pp. 216--229.

\bibitem{HarandiHSLS15}
M.~T. Harandi, R.~I. Hartley, C.~Shen, B.~C. Lovell, and C.~Sanderson,
  ``Extrinsic methods for coding and dictionary learning on {G}rassmann
  manifolds,'' \emph{International Journal of Computer Vision}, vol. 114, no.
  2-3, pp. 113--136, 2015.

\bibitem{Vemulapalli:CVPR2013}
R.~Vemulapalli, J.~Pillai, and R.~Chellappa, ``Kernel learning for extrinsic
  classification of manifold features,'' in \emph{CVPR}, 2013.

\bibitem{JayasumanaHSLH14}
S.~Jayasumana, R.~Hartley, M.~Salzmann, H.~Li, and M.~Harandi, ``Optimizing
  over radial kernels on compact manifolds,'' in \emph{CVPR}, 2014.

\bibitem{HarandiSJHL14}
M.~Harandi, M.~Salzmann, S.~Jayasumana, R.~Hartley, and H.~Li, ``Expanding the
  family of {G}rassmannian kernels: An embedding perspective,'' in \emph{ECCV},
  2014.

\bibitem{JayasumanaHSLH15}
S.~Jayasumana, R.~Hartley, M.~Salzmann, H.~Li, and M.~Harandi, ``Kernel methods
  on riemannian manifolds with gaussian {RBF} kernels,'' \emph{{IEEE} Trans.
  Pattern Anal. Mach. Intell.}, vol.~37, no.~12, pp. 2464--2477, 2015.

\bibitem{Harandi:CVPR15}
M.~Harandi and M.~Salzmann, ``Riemannian coding and dictionary learning:
  Kernels to the rescue,'' in \emph{CVPR}, 2015.

\bibitem{FeragenLH15}
A.~Feragen, F.~Lauze, and S.~Hauberg, ``Geodesic exponential kernels: When
  curvature and linearity conflict,'' in \emph{CVPR}, 2015.

\bibitem{DoCarmo:book92}
M.~P. do~Carmo, \emph{{R}iemannian Geometry}.\hskip 1em plus 0.5em minus
  0.4em\relax Birkh{\"a}user, 1992.

\bibitem{Quiroz13}
E.~A. {Papa Quiroz}, ``An extension of the proximal point algorithm with
  {B}regman distances on {Hadamard} manifolds,'' \emph{J. Glob. Optim.},
  vol.~56, no.~1, pp. 43--59, 2013.

\bibitem{NetoLimaO:98}
J.~X. da~Cruz~Neto, L.~L. de~Lima, and P.~R. Oliveira, ``Geodesic algorithms in
  {Riemannian} geometry,'' \emph{Balkan J. Geom. Appl.}, vol.~3, no.~2, pp. 89
  -- 100, 1998.

\bibitem{AttouchBRS:10}
H.~Attouch, J.~Bolte, P.~Redont, and A.~Soubeyran, ``Proximal alternating
  minimization and projection methods for nonconvex problems: An approach based
  on the {K}urdyka-{L}ojasiewicz inequality,'' \emph{Math. Oper. Res.},
  vol.~35, no.~2, pp. 438--457, 2010.

\bibitem{NetoOSS:13}
J.~X. da~Cruz~Neto, P.~R. Oliveira, P.~A.~S. Jr, and A.~Soubeyran, ``Learning
  how to play {N}ash, potential games and alternating minimization method for
  structured nonconvex problems on {R}iemannian manifolds,'' \emph{J. Convex
  Anal.}, vol.~20, no.~2, pp. 395 -- 438, 2013.

\bibitem{Absil:book07}
P.-A. Absil, R.~Mahony, and R.~Sepulchre, \emph{Optimization Algorithms on
  Matrix Manifolds}.\hskip 1em plus 0.5em minus 0.4em\relax Princeton
  University Press, 2007.

\bibitem{BoumalMAS:13}
N.~Boumal, B.~Mishra, P.-A. Absil, and R.~Sepulchre, ``{M}anopt, a {M}atlab
  toolbox for optimization on manifolds,'' \emph{J. Mach. Learn. Res.},
  vol.~15, pp. 1455--59, 2014.

\bibitem{EdelmanAS:1999}
A.~Edelman, T.~A. Arias, and S.~T. Smith, ``The geometry of algorithms with
  orthogonality constraints,'' \emph{SIAM J. Matrix Anal. Appl.}, vol.~20,
  no.~2, pp. 303--53, 1999.

\bibitem{HuaGalAbs2015}
W.~Huang, K.~A. Gallivan, and P.-A. Absil, ``A broyden class of quasi-{N}ewton
  methods for {R}iemannian optimization,'' \emph{SIAM Journal on Optimization},
  vol.~25, no.~3, pp. 1660--1685, 2015.

\bibitem{Ferreira:1998}
O.~Ferreira and P.~Oliveira, ``Subgradient algorithm on {R}iemannian
  manifolds,'' \emph{J. Optim. Theory Appl.}, vol.~97, no.~1, pp. 93--104,
  1998.

\bibitem{BBSW:SISC16}
M.~Ba\u{c}\'{a}k, R.~Bergmann, G.~Steidl, and A.~Weinmann, ``A second order
  non-smooth variational model for restoring manifold-valued images,''
  \emph{SIAM Journal on Scientific Computing}, vol.~38, no.~1, pp. A567--A597,
  2016.

\bibitem{KovnatskyGB16}
A.~Kovnatsky, K.~Glashoff, and M.~Bronstein, ``Madmm: A generic algorithm for
  non-smooth optimization on manifolds,'' in \emph{ECCV}, 2016.

\bibitem{Hosseini:SOPT17}
S.~Hosseini and A.~Uschmajew, ``A {R}iemannian gradient sampling algorithm for
  nonsmooth optimization on manifolds,'' \emph{SIAM Journal on Optimization},
  vol.~27, no.~1, pp. 173--189, 2017.

\bibitem{RockWets:98}
R.~T. Rockafellar and R.~J.-B. Wets, \emph{Variational analysis}, ser.
  Grundlehren der Mathematischen Wissenschaften.\hskip 1em plus 0.5em minus
  0.4em\relax Berlin: Springer, 1998, vol. 317.

\bibitem{Moreau:1965}
J.~J. Moreau, ``Proximit\'{e} et dualit\'{e} dans un espace hilbertien,''
  \emph{Bulletin de la Soci\'{e}t\'{e} Math\'{e}matique de France}, vol.~93,
  pp. 273--99, 1965.

\bibitem{KimEtAl:icml2015}
H.~Kim, J.~Xu, B.~Vemuri, and V.~Singh, ``Manifold-valued {D}irichlet
  processes,'' in \emph{ICML}, 2015.

\bibitem{Kheyfets2000}
A.~Kheyfets, W.~Miller, and G.~Newton, ``Schild's ladder parallel transport
  procedure for an arbitrary connection,'' \emph{International Journal of
  Theoretical Physics}, vol.~39, no.~12, pp. 2891--2898, 2000.

\bibitem{LorenziP14}
M.~Lorenzi and X.~Pennec, ``Efficient parallel transport of deformations in
  time series of images: From {S}child's to pole ladder,'' \emph{Journal of
  Mathematical Imaging and Vision}, vol.~50, no. 1-2, pp. 5--17, 2014.

\bibitem{JenkinsonBBWS:12}
M.~Jenkinson, C.~Beckmann, T.~Behrens, M.~Woolrich, and S.~Smith, ``{FSL},''
  \emph{NeuroImage}, vol.~62, no.~2, pp. 782--790, 2012.

\bibitem{Basser:95}
P.~Basser, ``Inferring microstructural features and the physiological state of
  tissues from diffusion-weighted images,'' \emph{NMR Biomed.}, vol.~8, pp.
  333--344, 1995.

\bibitem{Basser1996}
P.~Basser and C.~Pierpaoli, ``Microstructural and physiological features of
  tissues elucidated by quantitative-diffusion-tensor {MRI},'' \emph{Journal of
  Magnetic Resonance}, vol. 111, no.~3, pp. 209--219, 1996.

\end{thebibliography}
\bibliographystyle{IEEEtran}
}

\end{document}